\documentclass[10pt,journal,compsoc]{IEEEtran}
\usepackage{fancybox}
\usepackage{longtable}
\usepackage{pdflscape}
\usepackage{tikz}
\usetikzlibrary{arrows,positioning} 
\usepackage{amssymb}
\usepackage{graphicx}
\usepackage{url}
\usepackage{epstopdf}
\usepackage[verbose]{placeins}
\usepackage{rotating}
\usepackage{blindtext}
\usepackage{multirow}
\usepackage{lineno} % Start line numbering with
\usepackage[T1]{fontenc}
\usepackage{mwe}    % loads »blindtext« and »graphicx«
\usepackage{changes}
\usepackage{booktabs}
\usepackage{lipsum}% <- For dummy text
\definechangesauthor[name={Per cusse}, color=orange]{per}
\usepackage{subfig}
\usepackage{amsmath}
\usepackage{mathtools}
\usepackage{soul}
\usepackage{enumitem }
\usepackage{amsfonts}
\usepackage{epstopdf}
\usepackage{algorithm}
\usepackage{verbatim}
\usepackage{algorithmic}
\ifCLASSINFOpdf
\else
\fi

\IEEEdisplaynontitleabstractindextext
\IEEEpeerreviewmaketitle

\begin{document}
\title{Behavioral analysis of support vector machine classifier with Gaussian kernel and imbalanced data}

\author{\IEEEauthorblockN{Alaa Tharwat} \\
\IEEEauthorblockA{\textit{Faculty of Computer Science and Engineering, Frankfurt University of Applied Sciences, Frankfurt am Main, Germany}\\
engalaatharwat@hotmail.com}}

%\author{Alaa Tharwat}


%\institute{Alaa Tharwat$^*$ (Corresponding Author),Faculty of Computer Science and Engineering, Frankfurt University of Applied Sciences, Frankfurt am Main, Germany \\
%\email{\{alaa.othman\}@fh-bielefeld.de}
%}

\IEEEtitleabstractindextext{%
\begin{abstract} 
The parameters of support vector machines (SVMs) such as the penalty parameter and the kernel parameters have a great impact on the classification accuracy and the complexity of the SVM model. Therefore, the model selection in SVM involves the tuning of these parameters. However, these parameters are usually tuned and used as a black box, without understanding the mathematical background or internal details. In this paper, the behavior of the SVM classification model is analyzed when these parameters take different values with balanced and imbalanced data. This analysis including visualization, mathematical and geometrical interpretations and illustrative numerical examples with the aim of providing the basics of the Gaussian and linear kernel functions with SVM. From this analysis, we proposed a novel search algorithm. In this algorithm, we search for the optimal SVM parameters into two one-dimensional spaces instead of searching into one two-dimensional space. This reduces the computational time significantly. Moreover, in our algorithm, from the analysis of the data, the range of kernel function can be expected. This also reduces the search space and hence reduces the required computational time. Different experiments were conducted to evaluate our search algorithm using different balanced and imbalanced datasets. The results demonstrated how the proposed strategy is fast and effective than other searching strategies.
\end{abstract}
}

\maketitle

\IEEEdisplaynontitleabstractindextext
\IEEEpeerreviewmaketitle

\section{Introduction}\label{sec:introduction}
\label{Sec:Intro}
Support vector machines (SVMs) are among the well-known machine learning techniques which have been used for classification and regression problems \cite{wang2005support}. In SVM, training data are used for training a classification model. Next, this classification model is used for classifying an unknown/test sample. SVM obtains competitive results when the data is linearly separable which is always not practically possible. With overlapped classes, SVM has a penalty parameter which determines the trade-off between minimizing the training error and maximizing the SVM margin \cite{keerthi2003asymptotic}. In this case, the optimal hyperplane that separates different classes is linear. Increasing the complexity of the data to be nonlinearly separable makes using the linear hyperplane infeasible. Therefore, the kernel functions are used for mapping the data to a new feature space where the data is linearly separable. Thus, the penalty and kernel parameters have a great influence on the classification performance of SVM \cite{wang2005support}.

There are many studies optimized SVM parameters empirically by trying a finite set of values and keeping the values that achieved the best results. However, scanning the whole parameter space requires an exhaustive search and it is impractical in some applications due to the high computational time \cite{AlaaParameter}. For example, the grid search algorithm was employed to search for the optimal SVM parameters where the parameters vary with a fixed step-size through the parameter space; and this is time-consuming \cite{keerthi2003asymptotic}.

Different evolutionary optimization algorithms have been used for finding the optimal values of SVM parameters to improve the classification performance. For example, Ant Colony Optimization (ACO) was applied in \cite{zhang2010aco} for optimizing SVM parameters. In another research, Subasi employed the Particle Swarm Optimization (PSO) algorithm for finding SVM parameters \cite{subasi2013classification}. More recently, the bat algorithm, the whale optimization algorithm, and the multi-verse optimizer algorithm were used for optimizing SVM parameters \cite{tharwat2017ba,faris2018multi}. 
However, most of these studies assumed that the data is always balanced which is not frequently possible in real applications. 

The problem of imbalanced data is one of the main challenging problems to build a classification model. In the imbalanced data, the number of samples of one class (this is called the majority class) is higher than the number of samples of the other class (this is called the minority class). Hence, the classification model explores the majority class better than the minority class. As a result, the classification model tends to misclassify the minority samples either in the training or the testing phase \cite{he2008learning}. 

SVM parameters are usually tuned and used as a black box, without understanding the internal details. In this paper, in a step-by-step approach, the basics of SVM are introduced for explaining the objective function and constraints of SVM. This explanation shows numerically with illustrative examples and visualizations how SVM works with (1) linearly separable data, (2) overlapped data, and (3) nonlinearly separable data. Moreover, an overview of the kernel functions, the kernel trick, and the linear and Gaussian kernel functions are also introduced. In \cite{lin2001formulations,keerthi2003asymptotic}, mathematical analysis and proofs of SVM with the linear and Gaussian kernel functions were introduced. This analysis in \cite{lin2001formulations,keerthi2003asymptotic} missed some examples, clear explanations, and visualizations. In this paper, we present numerical examples and visualizations for explaining the behavior of SVM when the penalty and kernel parameters take different values. Moreover, our analysis takes into consideration how the balanced and imbalanced data affect the SVM model. Additionally, we propose a novel search algorithm. This algorithm has three main steps. The first two steps were introduced theoretically in \cite{lin2001formulations}. With these steps, instead of searching in a two-dimensional space, we search in two one-dimensional spaces. This decreases the required computational time significantly. The second step in \cite{lin2001formulations} assumed that there is only one line in the good region and the best solution can be found along this line. As a result, the grid search outperformed the proposed algorithm in \cite{lin2001formulations} with three datasets out of ten.  Practically, we found that there are many lines in the good region. This extends the search space slightly but increases the probability to find the optimal solution. In the previous studies, the same search space was used with different datasets. In the third step of the proposed algorithm, we propose to analyze the data to determine the search space of the kernel parameter and this reduces the search space and hence reduces the computational time. Different experiments have been carried out to evaluate the proposed search algorithm with different balanced and imbalanced data. We only focus on the binary classification problem and this can be extended to multiclass classification by using one-vs-one or one-vs-all strategies.

The rest of the paper is organized as follows: The background of the SVM classifier is introduced in details in Section \ref{Sec:SVM}. In Section \ref{Sec:ParameterOpt}, the behavior of SVM with linear and Gaussian kernel functions is introduced. This section includes visualizations, mathematical explanation, illustrative examples to show the training error, testing error, SVM margin, number of support vectors, and decision boundaries. Experimental results and discussions are introduced in Section \ref{Sec:Exp}. This section includes many experiments using real balanced and imbalanced datasets. Concluding remarks and future work are provided in Section \ref{Sec:Conc}.

\section{Theoretical background}
\label{Sec:SVM}
\subsection{Basics of learning from data}
\label{Sec:Basics}
Given, $N$ training samples/instances ($X=\{\textbf{x}_{1},\textbf{x}_{2},\dots,\textbf{x}_{N}\}$), where $\textbf{x}_{i}\in \mathcal{R}^d$ indicates the $i^{th}$ training sample and $d$ is the number of features. Each training sample has a class label, $y_i \in \{1,2,\dots,c \}$; therefore, the training set is $\{(\textbf{x}_{1},y_1),(\textbf{x}_{2},y_2),\dots,(\textbf{x}_{N},y_N)\}$, where $y_1, y_2,\dots,y_N$ indicate the class labels for $\textbf{x}_{1},\textbf{x}_{2},\dots,\textbf{x}_{N}$, respectively. The training data are randomly drawn from the input space ($X$) and system's responses ($Y$) with probability $P_X$ and $P_Y$, respectively, and hence we can say the training data consists of two random variables. Thus, the training data are drawn from an unknown probability distribution ($P(\textbf{x},y)$). As a consequence of that, different training data can be generated \cite{abu2012learning}. The testing data is also a part from the input space but it is different from the training data, i.e. the training and testing data are drawn independently according to $P(\textbf{x},y)$ \cite{AlaaParameter}.

The training data is used in the training phase for finding the relationship between $X$ and $Y$. This relationship represents a hypothesis or an approximation function ($h$). During the learning process, many hypotheses ($H=\{h_1,h_2,\dots, h_k\}$) are generated and the hypothesis that has the minimum \emph{empirical risk} ($R_{emp}$) is selected from $H$ \cite{abu2012learning}. The empirical risk is defined as the average losses of all training samples as follows:
\begin{equation*}
R_{emp}(h)=\frac{1}{N} \sum_{i=1}^{N} L(y_i, h(\textbf{x}_i))
%\label{EQN:EmpRisk}
\end{equation*}
where $h(\textbf{x}_i)$ is the expected value of $\textbf{x}_i$ and $L$ is the loss function\footnote{There are many loss functions that are used for classification problems such as 0-1 loss function which is defined as follows: $L(y_i,h(\textbf{x}_i))=\left\{\begin{matrix}
0 & \text{if } y_i= h(\textbf{x}_i)\\ 
1 & \text{if } y_i\neq h(\textbf{x}_i)\\ 
\end{matrix}\right.$} that is used for testing how well a hypothesis function ($h$) is estimating the target function ($y_i$). The estimation of the unseen/testing data is called the \emph{risk}, \emph{actual risk}, or \emph{expected risk} and it is denoted by $R$ and it is defined as follows:
\begin{equation*}
R(h)=\mathbb{E}_{(\textbf{x},y)\sim P(X,Y)}[L(y,h(\textbf{x}))] 
\end{equation*} 

However, the hypothesis that obtains the minimum empirical risk is not necessary to achieve a good risk. Therefore, the best hypothesis must be generalized to the data that we have not seen before. In other words, the best hypothesis obtains the minimum risk. Since, the joint distribution $P(X,Y)$ is unknown, the risk cannot be calculated. Hence, the goal of any learning algorithm is to reduce the gap between $R_{emp}$ and $R$ as follows:
\begin{equation*}
\mathbb{P}[\underset{h\in H}{sup} \left |   R(h)-R_{emp}(h)  \right | > \epsilon]
%\label{EQN:GenGap}
\end{equation*}
where $sup$ is called \emph{supremum} and it refers to the least upper bound of the absolute difference $\left |  R(h)-R_{emp}(h)  \right |$ and this bound is larger than a small value ($\epsilon$). The difference between the empirical risk and the risk is called the \emph{generalization gap}, and the goal of any learning algorithm is to make this gap as small as possible \cite{AlaaParameter}. 

Vapnik and Chervonenkis reported that the bound on the generalization gap is valid with probability $1-\eta$
\begin{equation}
\left |  R(h)-R_{emp}(h)  \right | \leq \Omega (N,\mathcal{V} ,\eta)
\label{EQN:LargeLaw2}
\end{equation}
where $\mathcal{V}$ is a non-negative integer and it represents the Vapnik Chervonenkis (VC) dimension and it measures the complexity of the learning model. Simply, learning models with many parameters (i.e. complex models) would have a high VC dimension, while learning models with few parameters (i.e. simple models) would have a low VC dimension. In Equation (\ref{EQN:LargeLaw2}), the empirical risk and risk depend on a particular hypothesis chosen during the training procedure \cite{burges1998tutorial}. The term $\Omega (N,\mathcal{V} ,\eta)$ in Equation (\ref{EQN:LargeLaw2}) is called the \emph{VC confidence} and it is defined as follows:
\begin{equation}
\Omega(N,\mathcal{V} ,\eta)=\sqrt{\frac{\mathcal{V}[ln(2N/\mathcal{V})+1]-ln(\eta/4)}{N}}
\label{EQN:Omega}
\end{equation}

\begin{figure}[!ht]
\centering
{\includegraphics[width=0.49\textwidth]{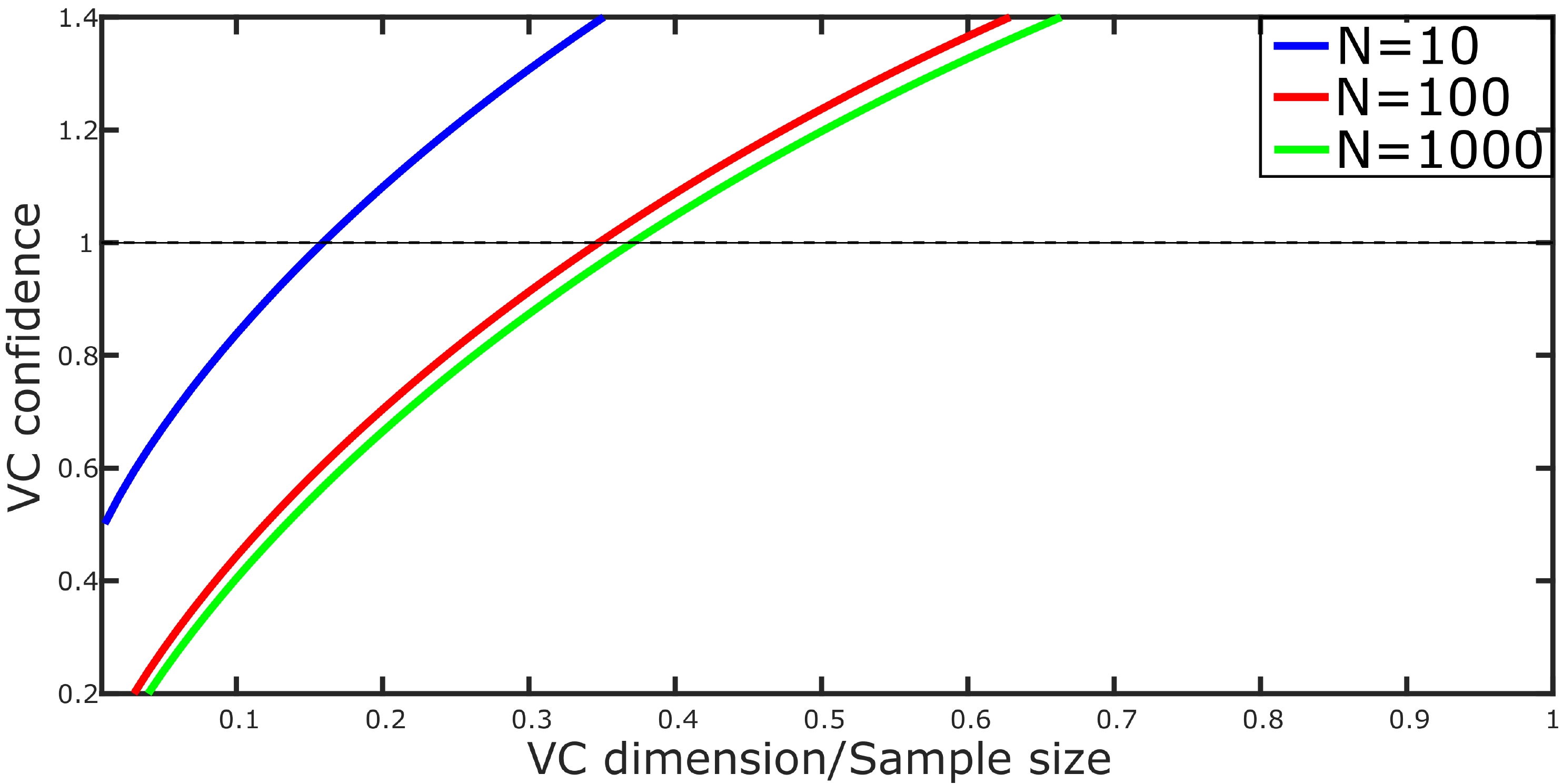}}
\caption{Visualization of the relation between the VC confidence term in Equation (\ref{EQN:Omega}) and the ratio of the VC dimension to the number of samples. }
\label{fig:VCConfidence}
\end{figure}

The VC confidence term has the following parameters:
\begin{itemize}
\item \textbf{VC dimension ($\mathcal{V}$)}: This measures the complexity of the model; thus, increasing $\mathcal{V}$ increases the number of hypotheses of the model. This increases the probability of getting a hypothesis with low or zero empirical risk. Mathematically, increasing $\mathcal{V}$ increases $\Omega$; as a result, increases the risk and this is called the \emph{overfitting} problem. On the contrary, decreasing the complexity of a model decreases $\Omega$ and this decreases the gap between the risk and the empirical risk and this means that the model generalizes well to new data. However, a very simple model increases the empirical risk and the risk; this is called the \emph{underfitting} problem \cite{abu2012learning}. Figure \ref{fig:VCConfidence} shows the influence of the model complexity on the performance of learning models. As shown, the VC confidence is a monotonic increasing function of $\mathcal{V}$ for any value of $N$. This means that the term $\Omega$ increases by increasing $\mathcal{V}$. Additionally, from the figure, it is clear that the VC confidence exceeds unity; consequently, the bound is so large.
\item \textbf{Number of training samples ($N$)}: Increasing $N$ decreases the gap between the risk and the empirical risk and this means that the empirical risk converges to the risk. This high number of training samples gives a chance for the learning model to explore the input space perfectly. This can be explained using the following example. 

\begin{figure}[!ht]
\centering
{\includegraphics[width=0.5\textwidth]{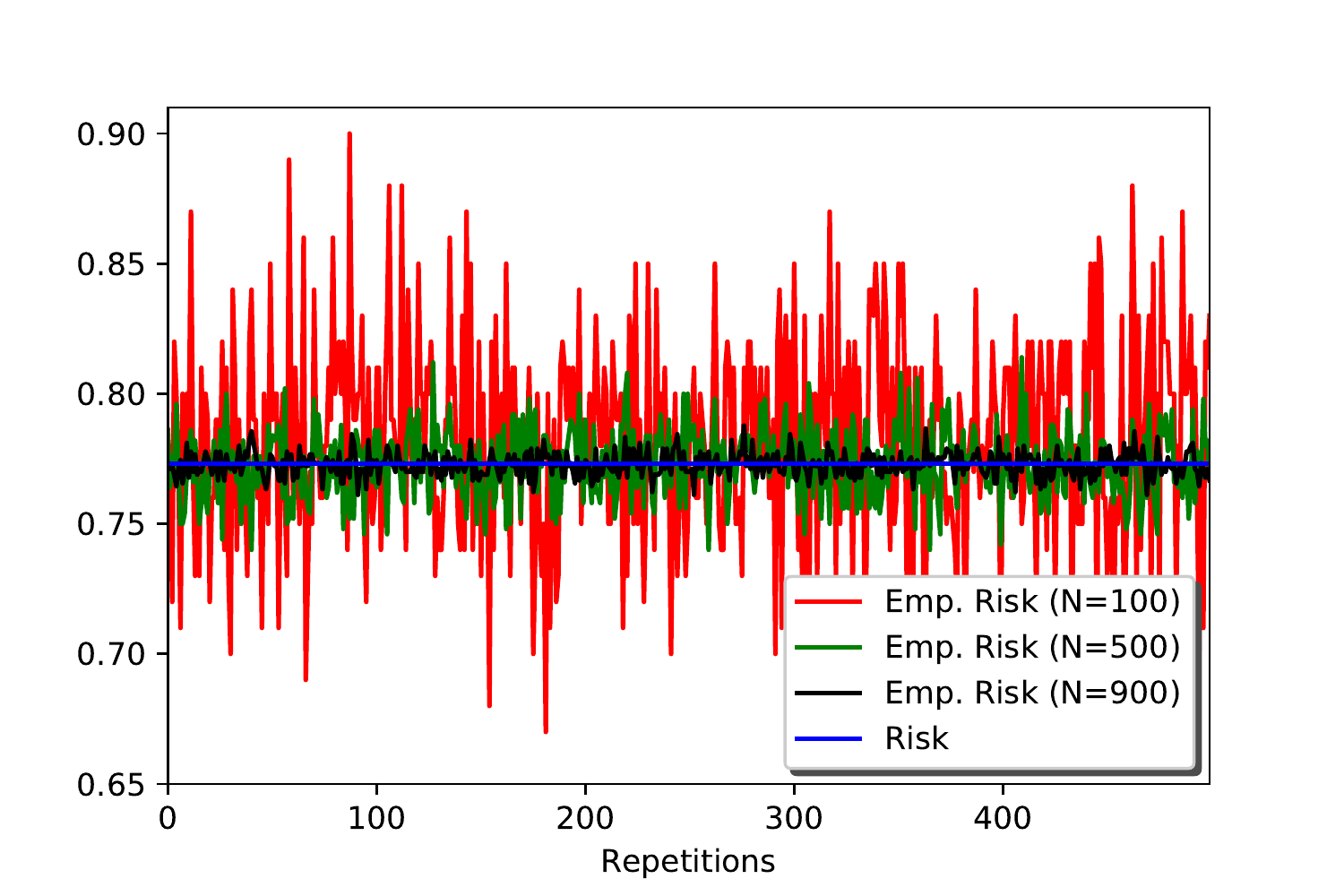}}
\caption{Visualization of the difference between the empirical risk and the risk using different numbers of training samples.}
\label{fig:Numberofsamples}
\end{figure}

Given data that has 1000 samples (this is the input space), which are classified into two classes. Given a simple classifier such as the linear discriminant classifier \cite{tharwat2016linear}. In the beginning, we train the model using all samples and test the trained model using all samples. The prediction result, in this case, represents the risk ($R$). To show the influence of the number of training samples on the generalization gap assume we have sets of training data with different sizes (100, 500, and 900). Figure \ref{fig:Numberofsamples} illustrates the results of this experiment and from the figure, all experiments were run for 500 times. Increasing the number of samples decreases the generalization gap. As shown, the risk is represented by a horizontal line (i.e. stable) because we used the same model with the whole data which represent the input space. Additionally, in Fig. \ref{fig:Numberofsamples}, with only 100 training samples that represent $\frac{100}{1000}=10\%$ from the whole data, there is a big difference between the risk and empirical risk. This big gap is decreased significantly by increasing the number of training samples to 900 samples.
\item \textbf{Confidence level ($1-\eta$)}: when the confidence level approaches one (i.e. $(1-\eta)\rightarrow 1$) this means that the confidence parameter ($\eta$) will be zero; thus, $\Omega\rightarrow \infty$. Let $\eta=0$ and then the confidence is $1-\eta=1$, the term $\mathcal{V}[ln(2N/\mathcal{V})+1]-ln(\eta/4)$ in Equation (\ref{EQN:Omega}) will be high and hence $\Omega$ increases as $\eta$ decreases \cite{abu2012learning}.
\end{itemize}

To conclude, the VC dimension has an important impact on the classification performance of learning models, where a very small VC dimension leads to the underfitting problem, while a high VC dimension leads to the overfitting problem. 

\subsection{Support Vector Machine (SVM)}
Support Vector Machines (SVM) is one of the well-known learning algorithms. The goal of SVM is not only to separate different classes by a decision boundary as in different classifiers but also to maximize the margin between different classes \cite{wang2005support}. In this section, the theoretical background of the SVM model is introduced. First, we begin with the simplest case where the data is linearly separable and there is no overlapping; thus, we will try to separate classes using linear hyperplanes. In this case, different examples and explanations are introduced for explaining how to find the optimal hyperplane. Afterward, we will allow some degree of overlapping between different classes. This will increase the complexity of the SVM model slightly, but, also we will try to separate the classes using linear hyperplanes. Increasing the complexity of the data to be nonlinearly separable makes using the linear hyperplanes inefficient. For this reason, the kernel functions are used for mapping the data from the input or feature space into a new higher-dimensional space where the data can be linearly separable. Selecting a suitable kernel function and tuning its parameters are two main challenges of using SVM \cite{wang2005support}. In this section, a brief description of the concept of SVM in the framework of classification will be introduced.

\begin{figure}[!ht]%2
\centering
\includegraphics[width=0.5\textwidth]{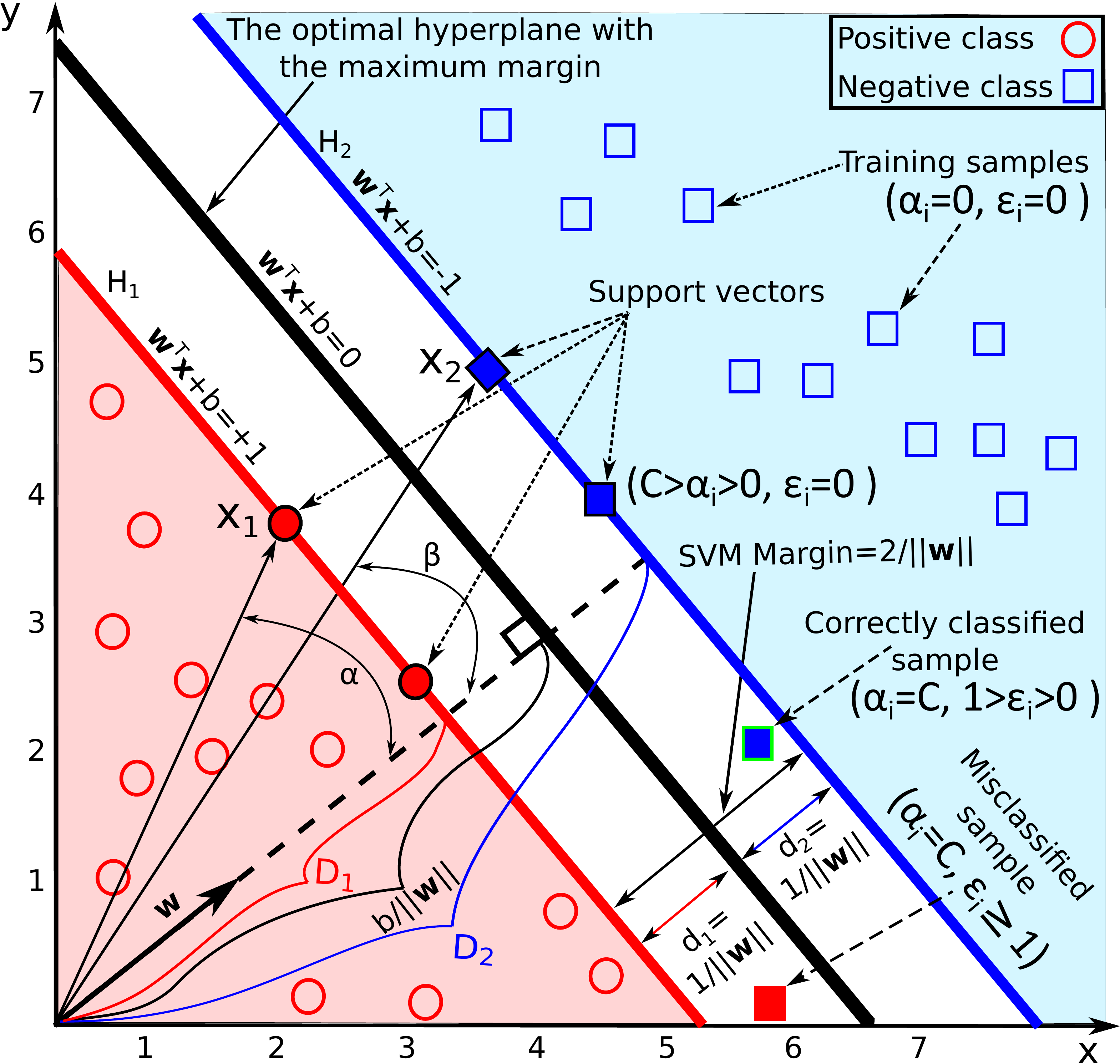} 
\caption{An example of binary classification SVM. The data consists of two classes. The optimal hyperplane is the solid thick black line, and the two planes ($H_1$ and $H_2$) are with two different colors. The two planes are parallel and the optimal hyperplane is equidistant from the two planes. }
\label{fig:SVMExampleLinear}
\end{figure}

\subsubsection{Linearly separable data}
Assume the data are linearly separable and the data consists of two classes (i.e. $y_i \in \pm 1$) as shown in Fig. \ref{fig:SVMExampleLinear}. Hence, the training data is $\{(\textbf{x}_1,y_1),(\textbf{x}_2,y_2),\dots,(\textbf{x}_N,y_N)\}$, where $N$ is the number of training data. Figure \ref{fig:SVMExampleLinear} illustrates the decision boundary which separates two classes (the positive classes ($\omega_+$) and the negative class ($\omega_-$)). The decision boundary is represented by a line but in higher dimensional spaces it will be a plane or a hyperplane \cite{wang2005support}. This hyperplane divides the feature/input space into two spaces, namely, the positive space where the samples from the positive class are located and the negative space where the samples from the negative class are located. The hyperplane is calculated as follows:
\begin{align}
\textbf{w}^{T}\textbf{x}+b=0 
\label{EQN:Hyperplane1}
\end{align}
where $\textbf{w}$ is a weight vector and it is normal to the hyperplane (see Fig. \ref{fig:SVMExampleLinear}), $b$ represents the bias or threshold, $T$ is the transpose of a matrix, and $\textbf{x}$ is the input vector or the training sample \cite{chen2005tutorial}. From Equation (\ref{EQN:Hyperplane1}), the decision or discriminant function is defined as follows:
\begin{align}
D(\textbf{x},\textbf{w},b)&= \sum_{i=1}^{d}w_ix_i+b = w_1x_1+\dots + w_dx_d+b
\label{EQN:Hyperplane2}
\end{align}
where $\textbf{w}=[w_1,w_2,\dots, w_d]$ and $\textbf{x}=[x_1,x_2,\dots,x_d]$. Given an unknown sample ($\textbf{x}_t$), using Equation (\ref{EQN:Hyperplane1}),  the decision rules will be as follows:
\begin{align*}
D(\textbf{x}_t,\textbf{w},b)= \left\{\begin{matrix}
>0 & \textbf{x}_t \in \text{ positive class}\\ 
=0 & \textbf{x}_t \text{ is on border}\\ 
<0 & \textbf{x}_t \in \text{ negative class}
\end{matrix}\right.
\end{align*}

\begin{figure}[!ht]%2
\centering
\includegraphics[width=0.5\textwidth]{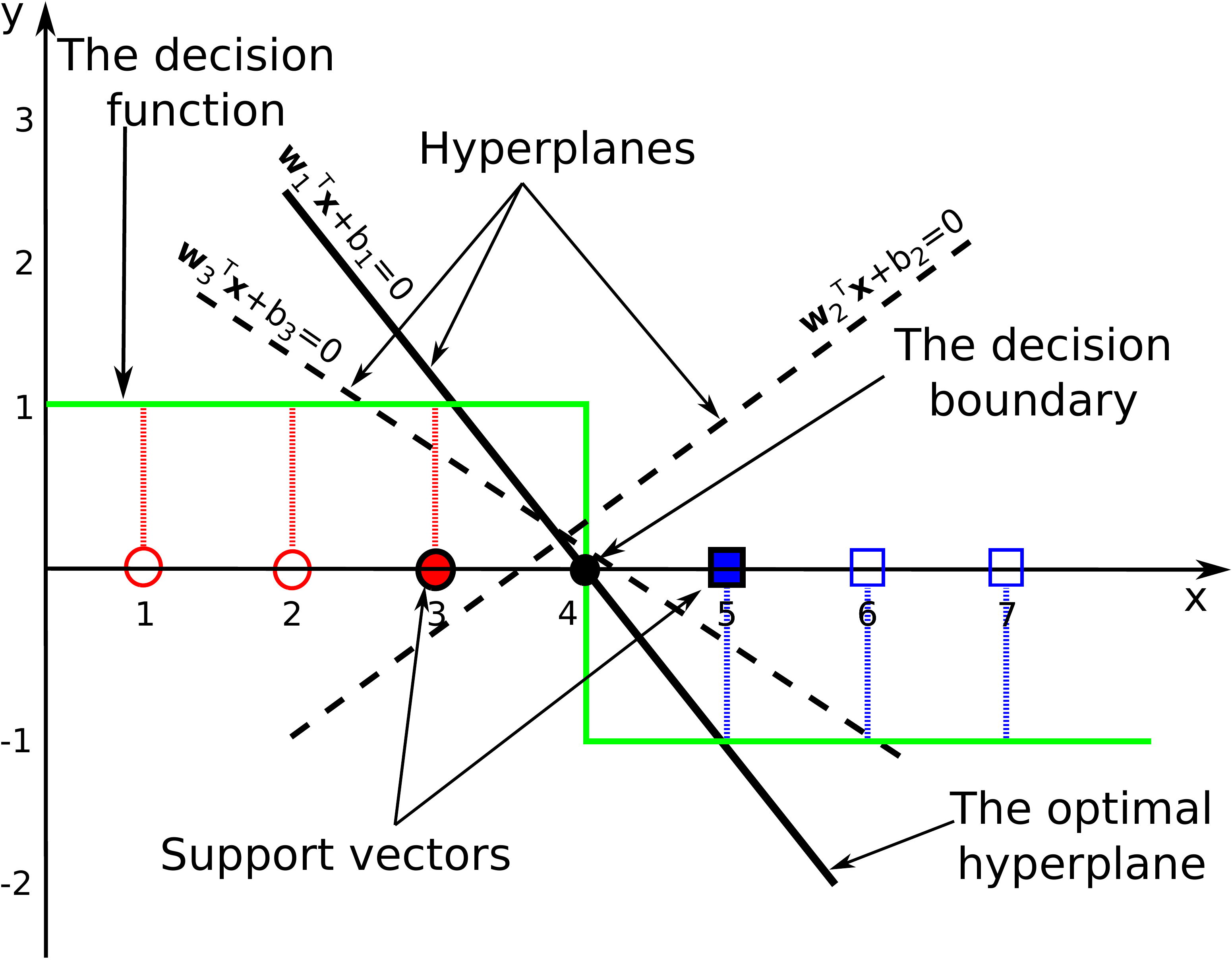} 
\caption{An example of the SVM classifier in one-dimensional space. Given two classes and each class has three samples, the figure shows three different hyperplanes, one of them is the optimal hyperplane ($\textbf{w}_1^T\textbf{x}+b_1=0$). Changing $\textbf{w}$ and $b$ generates new hyperplanes.}
\label{fig:SVMDecisionBoundary}
\end{figure}

Equations (\ref{EQN:Hyperplane1} and \ref{EQN:Hyperplane2}) indicate that the hyperplane is always \emph{over} the training/input space and hence the hyperplane lives in a $d+1$-dimensional space. Figure \ref{fig:SVMDecisionBoundary} shows how the hyperplane and the discriminant function are over the input space. The decision boundary is the intersection of the hyperplane and the input space; thus, the decision or separation boundary lives in the input space (i.e. $d$-dimensional space). This is also clear in Fig. \ref{fig:SVMDecisionBoundary}, where each sample in both classes is represented by one feature (i.e. one-dimensional space) and the decision boundary is represented by a point in the one-dimensional space. This point is the intersection between the hyperplane and the input space \cite{kecman2001learning}.

However, changing $\textbf{w}$ and $b$ generates an infinite number of hyperplanes (see Fig. \ref{fig:SVMDecisionBoundary}). In SVM, the aim is to orientate this hyperplane in such a way to be as far as possible from the closest samples of both classes; this hyperplane is the optimal hyperplane and these closest samples are called \emph{support vectors} (see Figs. \ref{fig:SVMExampleLinear} and \ref{fig:SVMDecisionBoundary}). In other words, the goal of SVM is to find the hyperplane with the largest margin. This can be achieved by determining $\textbf{w}$ and $b$ to construct the two planes ($H_1 \text{ and } H_2$) as follows:
\begin{equation}
\begin{split}
H_1 \rightarrow \textbf{w}^{T}\textbf{x}_{i}+b = +1 \;\; \text{for } y_i=+1\\
H_2 \rightarrow \textbf{w}^{T}\textbf{x}_{i}+b = -1 \;\; \text{for } y_i=-1
\end{split}
\label{EQN:Hyper}
\end{equation}
where $\textbf{w}^T\textbf{x}_i+b \geq +1$ is the plane for the positive class, i.e. $y_i=+1$, and $\textbf{w}^T\textbf{x}_i+b \leq -1$ is the plane for the negative class, i.e. $y_i=-1$. Figure \ref{fig:SVMExampleLinear} shows the optimal hyperplane and the two planes, i.e. $H_1$ and $H_2$, are parallel, have the same normal, and there are no training samples fall between them. Equation (\ref{EQN:Hyper}) can be written as follows:
\begin{equation}
y_i(\textbf{w}^T\textbf{x}_i+b)-1\geq 0 \;\; \forall i=1,2,\dots,N
\label{EQN:ST}
\end{equation}

From Equation (\ref{EQN:ST}), it is clear that $y_i(\textbf{w}^T\textbf{x}_i+b)= 1$ for all support vectors, and $y_i(\textbf{w}^T\textbf{x}_i+b)> 1$ for the other training samples. This is clear in Fig. \ref{fig:SVMExampleLinear} where the two points/samples $\textbf{x}_1$ and $\textbf{x}_2$ are on the planes $H_1$ and $H_2$, respectively (i.e. ($\textbf{x}_1$ and $\textbf{x}_2$) are support vectors). As a result, $\textbf{x}_1$ and $\textbf{x}_2$ satisfy the two planes equations in Equation (\ref{EQN:Hyper}). 

\begin{figure}[!ht]%2
\centering
\includegraphics[width=0.5\textwidth]{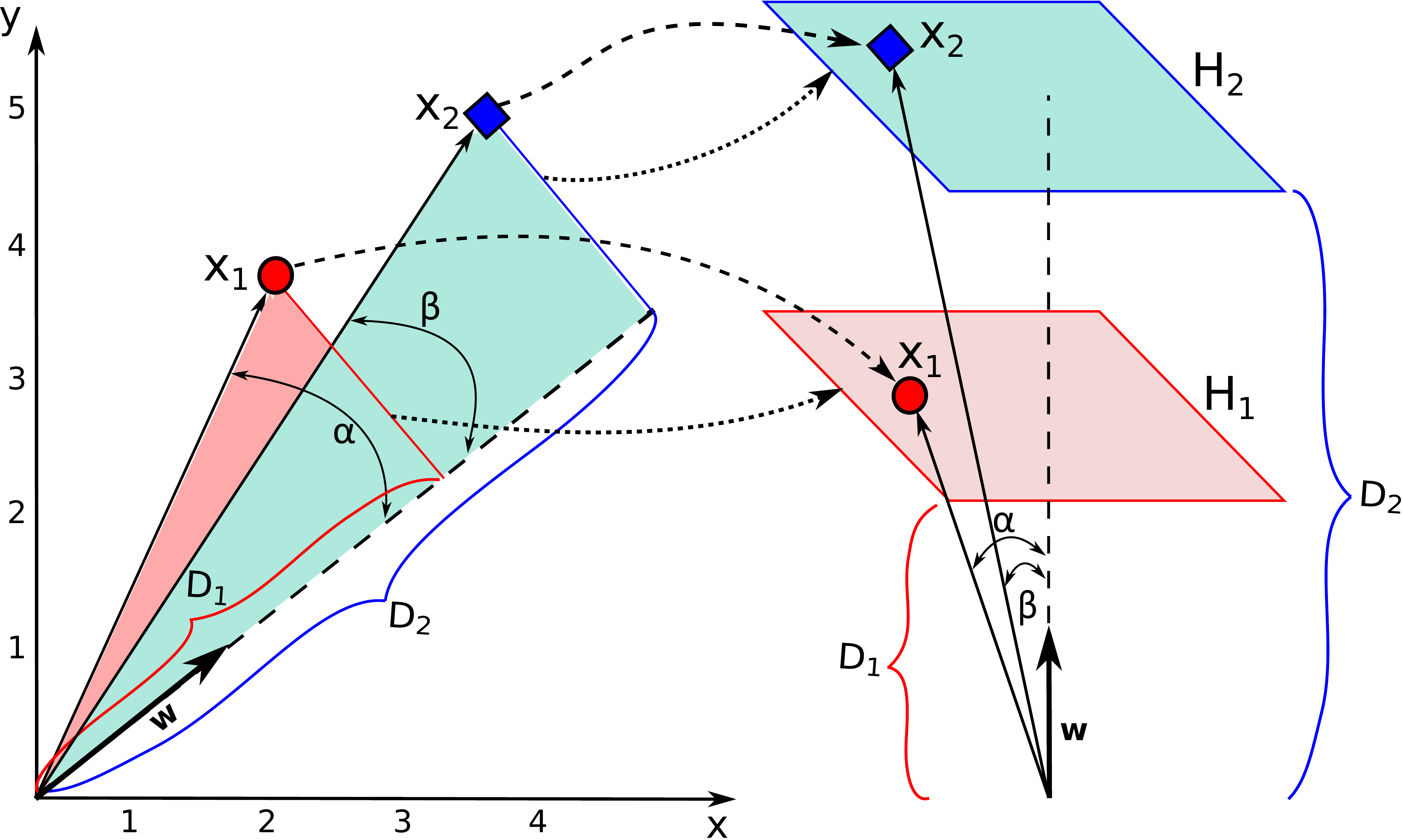} 
\caption{Visualization of the angles between $\textbf{x}_1$ and $\textbf{w}$ (i.e. $\alpha$ angle) and $\textbf{x}_2$ and $\textbf{w}$ (i.e. $\beta$ angle) (Note: this figure is a part of Fig. \ref{fig:SVMExampleLinear}).}
\label{fig:Angles}
\end{figure}

\paragraph{\textbf{Orientation of the hyperplane}}
The first step in SVM is to orientate the optimal hyperplane to maximize the margin subject to Equation (\ref{EQN:ST}). The margin of SVM is denoted by $M=d_1+d_2$, where $d_1$($d_2$) is the shortest distance from the optimal hyperplane to the closest positive (negative) sample. Thus, $d_1$ and $d_2$ represent the distance from the first and second plane, respectively, to the hyperplane, and $d_1=d_2$ (see Fig. \ref{fig:SVMExampleLinear}). This means that the hyperplane is equidistant from the two planes $H_1$ and $H_2$. In Fig. \ref{fig:SVMExampleLinear}, the SVM margin is calculated in the normal direction and it represents the distance between any support vectors from the two different classes. Hence, the margin is calculated as follows, $M=(\textbf{x}_1-\textbf{x}_2)_\textbf{w}$, where the subscript $\textbf{w}$ refers to the projection onto the weight vector $\textbf{w}$ direction, $\textbf{x}_1$ is support vector from the first class, and $\textbf{x}_2$ is a support vector from the second class. Figure \ref{fig:Angles} shows the projection of $\textbf{x}_1$ onto $\textbf{w}$ and it is denoted by $(\textbf{x}_1)_\textbf{w}=D_1$, and also the projection of $\textbf{x}_2$ onto $\textbf{w}$ and it is denoted by $(\textbf{x}_2)_\textbf{w}=D_2$; thus, $M=D_2-D_1$. Also, from Fig. \ref{fig:Angles}, the angle $\alpha$ is the angle between $\textbf{w}$ and $\textbf{x}_1$ and the angle $\beta$ is the angle between $\textbf{w}$ and $\textbf{x}_2$, and $\alpha$ and $\beta$ can be defined as follows:
\begin{equation}
cos(\alpha)=\frac{\textbf{x}_1^T\textbf{w}}{\left \| \textbf{x}_1 \right \|\left \| \textbf{w} \right \|} \;\;,\;\; cos(\beta)=\frac{\textbf{x}_2^T\textbf{w}}{\left \| \textbf{x}_2 \right \|\left \| \textbf{w} \right \|}
\label{EQN:Cos}
\end{equation}
where $\textbf{x}_i^T\textbf{w}$ is the dot product\footnote{The dot product between any two vectors $A$ and $B$ is calculated as follows: $A.B=\left \| A \right \| \left \| B \right \|cos \theta$, where $\theta$ is the angle between $A$ and $B$. It is also called the inner product or scalar product and it is also denoted by $\left \langle A, B \right \rangle$.} between $\textbf{x}_i$ and $\textbf{w}$, $\left \| \textbf{w} \right \|=\sqrt{\textbf{w}^T\textbf{w}}=\sqrt{w_1^2+w_2^2+\dots+w_d^2}$ is the Euclidean norm of $\textbf{w}$, and $\frac{\textbf{w}}{\left \| \textbf{w} \right \|}$ is the unit vector\footnote{Given a vector $A=\begin{bmatrix} 2 & 2 \end{bmatrix}$, the unit vector is a vector of length one and it is denoted by $\hat{A}$. The unit vector of the vector $A$ is calculated as follows, $\hat{A}= \frac{A}{\left \| A \right \|}=\frac{\begin{bmatrix} 2 & 2 \end{bmatrix}}{\sqrt{2^2+2^2}}=\begin{bmatrix} \frac{1}{\sqrt{2}} & \frac{1}{\sqrt{2}} \end{bmatrix}$. The norm of a unit vector is always one (i.e. $\left \| \hat{A} \right \|=\sqrt{(\frac{1}{\sqrt{2}})^2+(\frac{1}{\sqrt{2}})^2}=\sqrt{1}=1$). Multiplying a scalar value by a unit vector will be a new vector in the direction of the unit vector. For example, $3\hat{A}=\begin{bmatrix} \frac{3}{\sqrt{2}} & \frac{3}{\sqrt{2}} \end{bmatrix}$; hence, the length of the new vector will be 3.} in the direction of $\textbf{w}$. As a consequence, $D_1$ will be 
\begin{equation*}
D_1= (\textbf{x}_1)_\textbf{w}=\textbf{x}_1 \frac{\textbf{w}}{\left \| \textbf{w} \right \|}
\end{equation*}
and from Equation (\ref{EQN:Cos}),  $\textbf{x}_1^T\textbf{w}=cos(\alpha) \left \| \textbf{x}_1 \right \|\left \| \textbf{w} \right \|$ (this is simply the dot product between $\textbf{x}_1$ and $\textbf{w}$) and hence $D_1$ will be
\begin{align*}
D_1= \textbf{x}_1 \frac{\textbf{w}}{\left \| \textbf{w} \right \|} = \frac{cos(\alpha) \left \| \textbf{x}_1 \right \|\left \| \textbf{w} \right \|}{\left \| \textbf{w} \right \|}=\left \| \textbf{x}_1 \right \| cos (\alpha)
\end{align*}
similarly
\begin{equation*}
D_2=\left \| \textbf{x}_2 \right \|  cos(\beta)\
\end{equation*}

As a result, the margin is defined as follows:
\begin{align*}
M&=D_1-D_2=\left \| \textbf{x}_1 \right \|  cos(\alpha)-\left \| \textbf{x}_2 \right \|  cos(\beta)\ \nonumber \\
&=\left \| \textbf{x}_1 \right \| \frac{\textbf{x}_1^T\textbf{w}}{\left \| \textbf{x}_1 \right \|\left \| \textbf{w} \right \|}- \left \| \textbf{x}_2 \right \| \frac{\textbf{x}_2^T\textbf{w}}{\left \| \textbf{x}_2 \right \|\left \| \textbf{w} \right \|} = \frac{\textbf{x}_1^T\textbf{w}-\textbf{x}_2^T\textbf{w}}{\left \| \textbf{w} \right \|}
\end{align*}
where $\textbf{x}_1$ and $\textbf{x}_2$ as mentioned earlier are two support vectors from two different classes; hence, as denoted in Equation (\ref{EQN:ST}), $\textbf{w}^T\textbf{x}_1+b=1$ and $\textbf{w}^T\textbf{x}_2+b=-1$; thus
\begin{align*}
M=\frac{1-b-(-1-b)}{\left \| \textbf{w} \right \|}=\frac{2}{\left \| \textbf{w} \right \|}
%\label{EQN:Margin1}
\end{align*}

Figure \ref{fig:SVMExampleLinear} illustrates also the term $\frac{| b |}{\left \| \textbf{w} \right \|}$ which indicates the perpendicular distance from the origin to the optimal hyperplane. Additionally, the two planes ($H_1$ and $H_2$) have the same normal vector $\textbf{w}$, and the perpendicular distance from the origin to $H_1$ and $H_2$ is $\frac{|1-b|}{\left \| \textbf{w} \right \|}$ and $\frac{|-1-b|}{\left \| \textbf{w} \right \|}$, respectively.

\begin{figure}[!ht]%2
\centering
\includegraphics[width=0.5\textwidth]{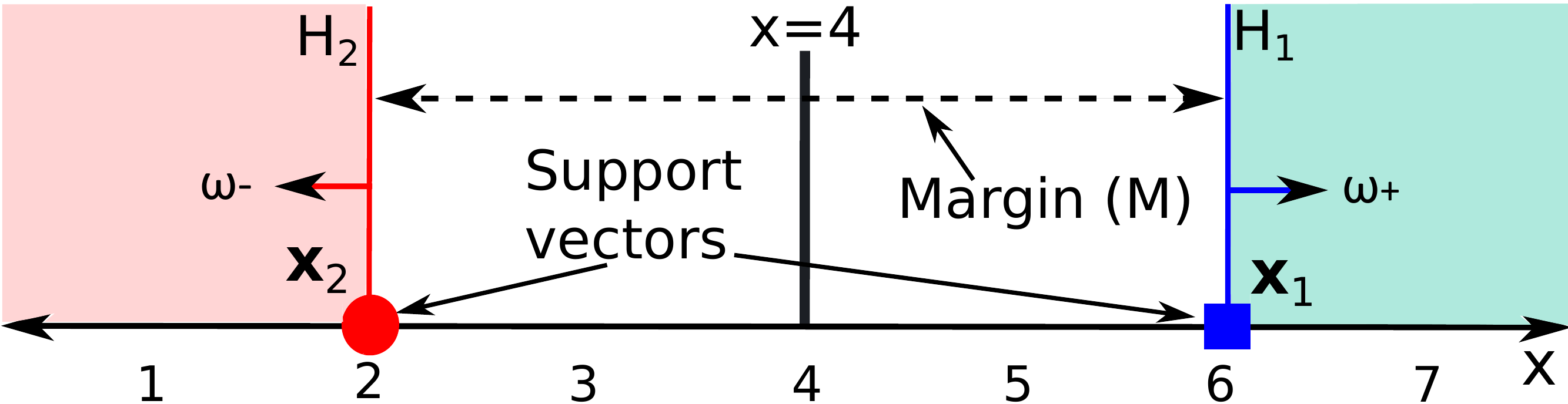} 
\caption{An example of SVM using linearly separable data.}
\label{fig:FirstExample}
\end{figure}

\paragraph{\textbf{Illustrative example}}
To explain how to orientate a hyperplane to find the optimal hyperplane, let us explain this step using a simple example. In this example, each sample is represented by only one feature (i.e. $\textbf{x}_i\in \mathcal{R}$) as shown in Fig. \ref{fig:FirstExample}. Assume we have two classes $\omega_-$ and $\omega_+$; the negative class has one sample as follows, $\omega_-=\{\textbf{x}_2\}$, $\textbf{x}_2=2$, and the positive class has also one sample as follows, $\omega_+=\{\textbf{x}_1\}$, $\textbf{x}_1=6$.

From Equation (\ref{EQN:ST})
\begin{align*}
\textbf{x}_1: \rightarrow \textbf{w}.6+b\geq 1 \;\; \text{and}\;\;  \textbf{x}_2: \rightarrow \textbf{w}.2+b\leq -1 
%\label{EQN:1DExample1}
\end{align*}

In this example, we can assume that there is a sample onto one plane, for example, $\textbf{x}_1=6$ is located onto the plane $H_1$ (i.e. the positive class), so, $\textbf{w}\textbf{x}_1+b=+1$. The plane $H_2$ is parallel to $H_1$ and the distance from $\textbf{x}_1$ to $\textbf{x}_2$ which is onto $H_2$ represents the margin ($M$). As we mentioned before, $M=\frac{2}{||\textbf{w}||}$. Hence, changing $\textbf{w}$ changes the margin. For example, the margin with $\left \| \textbf{w} \right \|=1, 2, $ and $4$ was 2, 1, and 0.5, respectively. It is worth mentioning that $\left \| \textbf{w} \right \|=0.1$ will increase the margin to 20, but it will not satisfy the constraints in Equation (\ref{EQN:ST}). This is because, with $\textbf{w}=0.1$ (this is equal to $\left \| \textbf{w} \right \|=0.1$ because $\textbf{w}$ has only one element), the margin $M$ will be 20 and hence the other sample ($\textbf{x}_2$) will be between the two planes ($H_1$ and $H_2$) and this is not matched with the constraints in Equation (\ref{EQN:ST}). This is also can be interpreted as follows, with $\textbf{w}=0.1$, $\textbf{x}_2=2$ is a sample from the negative class; thus, $0.1\times 2+b\leq -1$ and hence $b\leq -1.2$. While in the positive class, $\textbf{x}_1=6$ and $0.1\times 6+b\geq 1$ and hence $b\geq 0.4$, then, we cannot calculate $b$. Geometrically, $\textbf{w}=0.5$ obtains the maximum margin that satisfies the constraints in Equation (\ref{EQN:ST}).

After calculating $\textbf{w}$, it is easy to calculate $b$ as follows:
\begin{align*}
\text{using } \textbf{x}_1&:\; \textbf{w}\textbf{x}_1+b=1 \Rightarrow 0.5\times 6+b=1 \nonumber \\
\text{using } \textbf{x}_2&:\;\textbf{w}\textbf{x}_2+b=-1 \Rightarrow 0.5 \times 2+b=-1 
\end{align*}
hence $b$ will be -2. From Equation (\ref{EQN:Hyperplane1}), the optimal hyperplane or the discriminant function is $\textbf{w}^T\textbf{x}+b=0\Rightarrow 0.5x-2=0 \Rightarrow x=4$, and the width of the margin is equal to $\frac{2}{\left \| \textbf{w} \right \|}=\frac{2}{0.5}=4$ as shown in Fig. \ref{fig:FirstExample}. Moreover, the perpendicular distance from the decision boundary to the origin is $\frac{|b|}{\left \| \textbf{w} \right \|}=\frac{2}{\sqrt{0.5\times 0.5}}=4$ and the perpendicular distances from the two planes (i.e. $H_1$ and $H_2$) to the origin are $\frac{|1-b|}{\left \| \textbf{w} \right \|}=\frac{|1+2|}{0.5}=6$ and $\frac{|-1-b|}{\left \| \textbf{w} \right \|}=\frac{1}{0.5}=2$, respectively. 

Given an unknown or test sample $\textbf{x}_{test}$. This sample is classified by evaluating $y_{test}=\text{sign}(\textbf{w}^T\textbf{x}_{test}+b)$ and if $y_{test}$ is positive; thus, the new sample belongs to the positive class; otherwise, it belongs to the negative class. For example, given an unknown sample ($\textbf{x}_{test}=8$). To classify it, we substitute in the discriminant function equation as follows, $y_{test}=\text{sign }(\textbf{w}^T\textbf{x}_{test}+b)= \text{sign }(0.5\times 8 -2)=\text{sign }(2)$, which is positive. Thus, the unknown sample belongs to the positive class. 

\paragraph{\textbf{Finding the optimal hyperplane (primal form)}}
In our example, we tried to find the optimal hyperplane using only two samples, but, with a large set of samples, finding the optimal hyperplane will be difficult. This can be solved using one of the optimization techniques to find $\textbf{w}$ and $b$ that maximize the SVM margin subject to Equation (\ref{EQN:ST}) as follows:
\begin{align*}
&\text{min } f(\textbf{w})= \left \| \textbf{w} \right \| \nonumber \\
&\text{s.t. }  g(\textbf{w},b)=y_i(\textbf{w}^T\textbf{x}_i+b)-1\geq 0 \;\; \forall i=1,2,\dots, N
\end{align*}

Minimizing $\left \| \textbf{w} \right \|$ is equivalent to minimizing $\frac{1}{2} \left \| \textbf{w} \right \|^2$ as follows:
\begin{align}
&\text{min } f(\textbf{w})=  \frac{1}{2} \left \| \textbf{w} \right \|^2 \nonumber \\ 
&\text{s.t. } g(\textbf{w},b)=y_i(\textbf{w}^T\textbf{x}_i+b)-1\geq 0 \;\; \forall i=1,2,\dots, N
\label{EQN:SVM}
\end{align}

The margin in Equation (\ref{EQN:SVM}) is called the \emph{hard} margin, and the problem is a \emph{quadratic programming} problem with $N$ linear inequality constraints. More details about the quadratic programming problem and how it can be solved are introduced in Section \ref{Subsec:Quad}. 

The quadratic programming problem in Equation (\ref{EQN:SVM}) can be formulated into Lagrange formula by combining the objective function ($f(\textbf{w})= \frac{1}{2} \left \| \textbf{w} \right \|^2$) and the constraints ($g(\textbf{w},b)=y_i(\textbf{w}^T\textbf{x}_{i}+b)-1\geq 0$) as follows:
\begin{align}
\text{min } L(\textbf{w},b,\alpha)&=f(\textbf{w})-\sum_{i} \alpha_i g(\textbf{w},b) \nonumber \\
&=\frac{\left \|  \textbf{w}\right \|^2}{2}-\sum_{i} \alpha_i (y_i(\textbf{w}^T\textbf{x}_{i}+b)-1) \nonumber \\
&=\frac{\left \|  \textbf{w}\right \|^2}{2}-\sum_{i} \alpha_i y_i(\textbf{w}^T\textbf{x}_{i}+b)+\sum_{i=1}^{N} \alpha_i
\label{EQN:SVMOptPrimal}
\end{align}
where $\alpha_i\geq 0, i=1,2,\dots,N$ represent the Lagrange multipliers and each Lagrange multiplier ($\alpha_i$) corresponds to one training sample ($\textbf{x}_{i}$). It is worth mentioning that, $\textbf{x}_i$ in the constraint in Equation (\ref{EQN:SVM}) is not a variable because it represents the training data which is known. 

The optimal solution is a \emph{saddle} point that minimizes $L$ with respect to $\textbf{w}$ and $b$ and maximizes $L$ with respect to $\alpha_i$. To calculate the values of $\textbf{w}$, $b$, and $\alpha_i$ that minimize the objective function in Equation (\ref{EQN:SVMOptPrimal}), $L$ is differentiating with respect to $\textbf{w}$, $b$, and $\alpha_i$ and setting the derivatives to zero as follows:
\begin{align}
\frac{\partial L(\textbf{w},b,\alpha)}{\partial \textbf{w}}&=0 \nonumber \\
 &\Rightarrow \textbf{w}-\sum_{i=1}^{N}\alpha_iy_i\textbf{x}_{i}=0  \Rightarrow \textbf{w}=\sum_{i=1}^{N}\alpha_iy_i\textbf{x}_{i}
\label{EQN:SVMLPW}
\end{align} 
\begin{equation}
\frac{\partial L(\textbf{w},b,\alpha)}{\partial b}=0 \Rightarrow \sum_{i=1}^{N}\alpha_iy_i=0
\label{EQN:SVMLPb}
\end{equation} 
\begin{equation}
\frac{\partial L(\textbf{w},b,\alpha)}{\partial \alpha_i}=0 \Rightarrow y_i(\textbf{w}^T\textbf{x}_{i}+b)-1=0
\label{EQN:SVMLAlpha}
\end{equation} 

From Equation (\ref{EQN:SVMLPW}), it is clear that $\textbf{w}=\sum_{i=1}^{N}\alpha_iy_i\textbf{x}_{i}$ and hence $\textbf{w}=\sum_{\textbf{x}_i\in \omega_+} \alpha_i\textbf{x}_{i} -\sum_{\textbf{x}_i \in \omega_-}\alpha_i\textbf{x}_{i}$. In SVM, most of $\alpha_i$'s are zeros; hence, the sparseness is a common property of SVM; and the non-zero $\alpha$'s are corresponding to only the support vectors, which are the samples closest to the separating hyperplane. In other words, for each support vector ($\alpha_i>0$), the constraint in Equation (\ref{EQN:SVMOptPrimal}) is active (i.e. $y_i(\textbf{w}^T\textbf{x}_{i}+b)-1=0$); otherwise, this constraint will be inactive (this is for the other training samples). This means that there are support vectors from both classes (the positive and negative classes) and this is also clear in Equation (\ref{EQN:SVMLPb}) where $\sum_{\omega_+}\alpha_i=\sum_{\omega_-}\alpha_i$. Thus, the term $\sum \alpha_i$ (i.e. sum of Lagrange multipliers) of the negative and positive classes is equal. Finally, Equation (\ref{EQN:SVMLAlpha}) represents the constraints in Equation (\ref{EQN:ST}). 

In our example in Fig. \ref{fig:FirstExample}, we have two samples and hence the constraint $g(\textbf{w},b)$ can be expanded as follows:
\begin{align*}
g_1(\textbf{w},b)&= (\textbf{w}\textbf{x}_1+b)-1 \text{  where } y_1=+1 \nonumber \\
g_2(\textbf{w},b)&= -(\textbf{w}\textbf{x}_2+b)-1 \text{  where } y_2=-1 
\end{align*}
and the objective function will be
\begin{align}
\text{min} \;L(\textbf{w},b,\alpha)&=\frac{\left \|  \textbf{w}\right \|^2}{2}-\alpha_1g_1(\textbf{w},b)-\alpha_2g_2(\textbf{w},b) \nonumber \\
&=\frac{\left \|  \textbf{w}\right \|^2}{2} -\alpha_1(y_1(\textbf{w}\textbf{x}_1+b)-1)\nonumber \\
&-\alpha_2(y_2(\textbf{w}\textbf{x}_2+b)-1) 
\label{EQN:ExampleOptimization1}
\end{align}
where $\textbf{x}_1=6$ and it ($\textbf{x}_1$) belongs to the positive class (i.e. $y_1=+1$) and $\textbf{x}_2=2$ and it belongs to the negative class (i.e. $y_2=-1$). Equation (\ref{EQN:ExampleOptimization1}) will be
\begin{align}
\text{min} \;L(\textbf{w},b,\alpha)&=\frac{\left \|  \textbf{w}\right \|^2}{2}-\alpha_1((6\textbf{w}+b)-1)\nonumber \\
&-\alpha_2(-(2\textbf{w}+b)-1) 
\label{EQN:ExampleOptimization2}
\end{align}

The values of $\textbf{w}$, $b$, and $\alpha_i$ are calculated as follows:
\begin{equation}
\frac{\partial L(\textbf{w},b,\alpha)}{\partial \textbf{w}}=0 \Rightarrow \textbf{w}-6\alpha_1+2\alpha_2=0
\label{EQN:SVMLPWExample1}
\end{equation} 
\begin{equation}
\frac{\partial L(\textbf{w},b,\alpha)}{\partial b}=0 \Rightarrow -\alpha_1+\alpha_2=0
\label{EQN:SVMLPWExample2}
\end{equation} 
\begin{equation}
\frac{\partial L(\textbf{w},b,\alpha)}{\partial \alpha_1}=0 \Rightarrow 6\textbf{w}+b-1=0
\label{EQN:SVMLPWExample3}
\end{equation} 
\begin{equation}
\frac{\partial L(\textbf{w},b,\alpha)}{\partial \alpha_2}=0 \Rightarrow 2\textbf{w}+b+1=0
\label{EQN:SVMLPWExample4}
\end{equation} 

Equation (\ref{EQN:SVMLPWExample1}) is matched with Equation (\ref{EQN:SVMLPW}) and $\textbf{w}=6\alpha_1-2\alpha_2\Rightarrow \textbf{w}=6y_1\alpha_1+2y_2\alpha_2 \Rightarrow \textbf{w}=\textbf{x}_1y_1\alpha_1+\textbf{x}_2y_2\alpha_2=\sum_{i=1}^{N}\alpha_iy_i\textbf{x}_{i}$. Additionally, Equation (\ref{EQN:SVMLPWExample2}) agrees with Equation (\ref{EQN:SVMLPb}) and hence $\alpha_1=\alpha_2 \Rightarrow \sum_{i=1}^{N}\alpha_iy_i=0$. Equations (\ref{EQN:SVMLPWExample3} and \ref{EQN:SVMLPWExample4}) represent the constraints in Equation (\ref{EQN:ST}).

In Equations (\ref{EQN:SVMLPWExample3} and \ref{EQN:SVMLPWExample4}), the value of $\textbf{w}$ and $b$ will be 0.5 and -2, respectively, and these results are identical to the results that we obtained earlier from our geometrical and mathematical analysis. From Equations (\ref{EQN:SVMLPWExample1} and \ref{EQN:SVMLPWExample2}), $\alpha_1=\alpha_2=\frac{1}{8}$; and $\textbf{w}$ can be calculated also as follows, $\textbf{w}=\alpha_1y_1\textbf{x}_1+\alpha_2y_2\textbf{x}_2=\frac{1}{8}\times (1)\times 6+\frac{1}{8}\times (-1)\times 2=\frac{1}{2}$. Therefore, the optimal hyperplane is $\textbf{w}\textbf{x}+b=0\Rightarrow 0.5\textbf{x}-2=0$ or $\textbf{x}=4$. The optimal hyperplane or the discriminant function can also be calculated as in Equation (\ref{EQN:Hyperplane2}) and the equation of the hyperplane will be $\sum_{i=1}^{d}w_ix_i+b=\textbf{w}^T\textbf{x}+b=\frac{1}{2}\textbf{x}+b=0$. The value of $b$ is calculated by substituting any of the two samples (2,-1) or (6,+1) (i.e. $(\textbf{x}_i,y_i)$) into the hyperplane equation as follows, $\frac{1}{2}\times 2+b=-1 \Rightarrow b=-2$ and hence the final equation of the optimal hyperplane is $0.5\textbf{x}-2=0$, which is identical to the results we obtained before. 

From this example, it is interesting to know that adding/removing one or more samples outside the margin will not affect the results of our example (i.e. the optimal hyperplane, planes, decision boundary, or the margin). This is because these samples are considered as training samples and not support vectors and the Lagrange multipliers of these new samples will be zero. On the other hand, removing one of the support vectors will change the decision boundary. This is due to all training samples are not relevant to the optimization problem, i.e. their corresponding constraints do not play role in the optimization. 

\paragraph{\textbf{Finding the optimal hyperplane (dual form)}}
\label{SubSec:Dual}
In Equation (\ref{EQN:SVMOptPrimal}), $L$ represents the primal problem it is also denoted by $L_P$ in some references. The objective function is convex and a set of linear constraints defines also a convex set. Therefore, the optimization function in Equation (\ref{EQN:SVMOptPrimal}) is a convex quadratic programming problem and hence we can solve the dual problem by maximizing $L_P$ subject to (1) the constraints $\alpha_i\geq 0$, (2) the constraints that the gradient of $L_P$ with respect to $\textbf{w}$ and $b$ vanish\footnote{This optimization problem may be viewed from either of two perspectives, the primal problem or the dual problem, and this is called the \emph{duality principle}. More details about the primal and dual problems are in Section \ref{Subsec:Dual}.} \cite{chen2005tutorial}. The dual form of the SVM problem is called the \emph{Wolfe} dual \cite{burges1998tutorial}, and it can be formulated by substituting Equations (\ref{EQN:SVMLPW} and \ref{EQN:SVMLPb}) into Equation (\ref{EQN:SVMOptPrimal}) as follows:
\begin{align}
\text{max } L_D(\alpha)&= \frac{\left \|  \textbf{w}\right \|^2}{2}-\sum_{i} \alpha_i (y_i(\textbf{w}^T\textbf{x}_{i}+b)-1) \nonumber \\
&=  \underset{\frac{\left \|  \textbf{w}\right \|^2}{2}}{\underbrace{\frac{(\sum_{i=1}^{N}\alpha_iy_i\textbf{x}_{i})^2}{2}}}\nonumber \\ 
&-\sum_{i} \alpha_i (y_i( \underset{ \textbf{w}}{\underbrace{\sum_{j=1}^{N}\alpha_jy_j\textbf{x}_{j}}} \textbf{x}_{i}+b)-1)\nonumber \\
&=\frac{(\sum_{i=1}^{N}\alpha_iy_i\textbf{x}_{i})^2}{2}+\sum_{i} \alpha_i- \underset{0}{\underbrace{\sum_{i} \alpha_i y_i b}} \nonumber \\
&- \sum_{i,j} \alpha_i\alpha_j y_iy_j\textbf{x}_i\textbf{x}_j \nonumber \\
&=\sum_{i=1}^{N} \alpha_i -\frac{1}{2} \sum_{i,j} \alpha_i \alpha_j y_i y_j\textbf{x}_i^T\textbf{x}_j\nonumber  \\  
\text{ s.t.} \; &\alpha_i\geq 0 \;\; , \sum_{i=1}^{N}\alpha_i y_i=0 \; \forall i=1,2,\dots,N 
\label{EQN:SVMOptDual1}
\end{align}
where $L_D$ indicates the dual form of $L_P$ which needs to be maximized instead of minimizing $L_P$. Equation (\ref{EQN:SVMOptPrimal}) indicates that the objective function of the primal SVM problem is minimized with respect to $\textbf{w} \text{ and }b$, while with the dual SVM problem in Equation (\ref{EQN:SVMOptDual1}), the objective is to maximize $L_D$ with respect to $\alpha_i$. Moreover, the primal problem has $d+1$ primal variables ($w_1,w_2,\dots,w_d,b$), while the number of variables in the dual problem is equal to the number of training samples ($\alpha_1, \alpha_2,\dots, \alpha_{N}$), and after the learning process, the number of free parameters will be equal to the number of support vectors. Hence, with $d\gg n$, the dual problem will be faster than the primal one. 

For convex problems (e.g. SVM) the Karuch-Kuhn-Tucker (KKT) conditions are necessary and sufficient for $\textbf{w}$, $b$, and $\alpha_i$ to be a solution \cite{chen2005tutorial}. The KKT conditions are
\begin{align}
& y_i(\textbf{w}^T\textbf{x}_i+b)-1\geq 0 \nonumber \\
&\alpha_i \geq 0  \nonumber \\
&\textbf{w}=\sum_{i=1}^{N}\alpha_iy_i\textbf{x}_{i} \text{ , } \sum_{i=1}^{N}\alpha_iy_i=0 \nonumber \\
&\alpha_i \{y_i(\textbf{w}^T\textbf{x}_i+b)-1\} = 0
\label{EQN:KKTCond1}
\end{align}
where $i=1,\dots, N$, the first condition ($y_i(\textbf{w}^T\textbf{x}_i+b)-1\geq 0$) is called the primal feasibility or primal admissibility condition and it is in Equations (\ref{EQN:ST} and \ref{EQN:SVM}), the second condition ($\alpha_i \geq 0$) is called the dual feasibility condition and it is indicated in Equation (\ref{EQN:SVMOptDual1}), the two conditions in the third line are the gradient of Lagrangian and these conditions are called the zero gradient conditions as indicated in Equations (\ref{EQN:SVMLPW} and \ref{EQN:SVMLPb}); finally, the condition in the last line is called the \emph{complementary} condition. With this condition, for every sample ($\textbf{x}_i$), either the corresponding $\alpha_i$ of this sample must be zero (this is for all training samples except support vectors) or the term in squared brackets is zero (this is for all support vectors). 

The matrix notation of Equation (\ref{EQN:SVMOptDual1}) will be
\begin{align}
\text{max } L_D(\alpha)&=\sum_{i=1}^{N} \alpha_i -\frac{1}{2} \sum_{i,j} \alpha_i H_{ij} \alpha_j \nonumber  \\
&=\textbf{f}^T\alpha -\frac{1}{2} \alpha^T \textbf{H} \alpha\nonumber \\  &\text{ s.t. } \alpha_i\geq 0 \;,\; \alpha \textbf{y}^T=0
\label{EQN:SVMOptDual}
\end{align}
where $H_{ij}=y_iy_j\textbf{x}_i.\textbf{x}_j$ and $\textbf{H}$ is calculated as follows:
\begin{align*}
\textbf{H}=\begin{bmatrix}
y_1y_1\left \langle \textbf{x}_1,\textbf{x}_1 \right \rangle & \dots & y_1y_N\left \langle \textbf{x}_1,\textbf{x}_N \right \rangle\\ 
 \vdots& \ddots &\vdots \\ 
y_Ny_1\left \langle \textbf{x}_N,\textbf{x}_1 \right \rangle & \dots & y_Ny_N\left \langle \textbf{x}_N,\textbf{x}_N \right \rangle 
\end{bmatrix}
\end{align*}
and $\textbf{f}$ is $(N\times 1)$ unit vector as follows, $\textbf{f}=\vec{\textbf{1}}=\begin{bmatrix} 1 & 1 &\dots & 1 \end{bmatrix}^T$, and $\alpha=\begin{bmatrix} \alpha_1 & \alpha_2 &\dots & \alpha_N\end{bmatrix}^T$. In $\textbf{H}$, if the two samples $\textbf{x}_i$ and $\textbf{x}_j$ are completely dissimilar (e.g. perpendicular), the value of $\textbf{x}_i.\textbf{x}_j$ will be zero or very small and hence these two samples cannot contribute in $\textbf{H}$ and the objective function. On the other hand, if the two samples are similar, the value of $\textbf{x}_i.\textbf{x}_j$ will be high. If the two samples are similar and predict the same output (i.e. $y_i=y_j$); so, $y_iy_j\textbf{x}_i.\textbf{x}_j>0$ and this will decrease $L_D$. On the contrary, if the two samples are similar and predict opposite outputs (i.e. $y_i\neq y_j$); thus, $y_iy_j\textbf{x}_i.\textbf{x}_j<0$ and this will increase $L_D$. As a consequence of that, critical patterns (e.g. border line samples that belong to different classes) are good for constructing a good margin. 

Maximizing the function in Equation (\ref{EQN:SVMOptDual}) is equal to
\begin{align}
\text{min }L_D(\alpha)=\frac{1}{2} \alpha^T \textbf{H} \alpha-\textbf{f}^T\alpha
\label{EQN:SVMOptDual2}
\end{align}
subject to the same constraints. 

The solution of the dual problem ($\alpha_i^*$) determines the parameters of the optimal hyperplane ($\textbf{w}$ and b), where $\textbf{w}$ is calculated as in Equation (\ref{EQN:SVMLPW}) and hence $\textbf{w}$ is calculated using only the support vectors. The value of $b$ is calculated from the complementary condition in Equation (\ref{EQN:KKTCond1}) as follows:
\begin{align*}
\alpha_i \{y_i(\textbf{w}^T\textbf{x}_i+b)-1\}=0 
\end{align*}
for support vectors (i.e. $\alpha_i>0$)
\begin{align*}
y_i(\textbf{w}^T\textbf{x}_i+b)-1=0, \;\; i=1,2,\dots, N_{SV}
\end{align*}
where $N_{SV}$ is the number of support vectors. Therefore, $b$ is
\begin{align}
b=\frac{1}{N_{SV}}(\sum_{i=1}^{N_{SV}} (\frac{1}{y_i}-\textbf{w}^T\textbf{x}_i))
\label{EQN:bNSV}
\end{align}

Solving Equations (\ref{EQN:SVMLPW}, \ref{EQN:SVMLPb}, and \ref{EQN:bNSV}) leads to determine the values of $\textbf{w}$, $\alpha$, and $b$.

In our example in Equation (\ref{EQN:ExampleOptimization2}), from Equation (\ref{EQN:SVMLPWExample1}), $\textbf{w}=6\alpha_1-2\alpha_2$, and from Equation (\ref{EQN:SVMLPWExample2}), we found that $\alpha_1=\alpha_2$; thus, $\textbf{w}=6\alpha_1-2\alpha_1=4\alpha_1$. Also, from Equation (\ref{EQN:SVMLPWExample3}), the value of $b$ is $1-6\textbf{w}$. Therefore, the dual form of the optimization problem in our example is as follows:
\begin{align*}
\text{max} \;L_D(\alpha)&=\frac{\left \|  \textbf{w}\right \|^2}{2}-\alpha_1((6\textbf{w}+b)-1)-\alpha_2(-(2\textbf{w}+b)-1)  \nonumber \\
&= \frac{(4\alpha_1)^2}{2}-\alpha_1(24\alpha_1+b-1)-\alpha_1(-8\alpha_1-b-1) \nonumber \\
&= 8\alpha_1^2-24\alpha_1^2-b\alpha_1+\alpha_1+8\alpha_1^2+b\alpha_1+\alpha_1 \nonumber \\
&=-8\alpha_1^2+2\alpha_1
\end{align*}
therefore 
\begin{align*}
\frac{\partial L_D(\alpha)}{\partial \alpha_1}=0 \Rightarrow -16\alpha_1+2=0 
\end{align*}
hence $\alpha_1=\alpha_2=\frac{2}{16}=\frac{1}{8}$ and the value of $\textbf{w}$ is calculated as follows, $\textbf{w}=4\alpha_1=4\times \frac{1}{8}=0.5$ and $b=1-6\textbf{w}=1-6\times 0.5=-2$. These results agree with our results that were calculated using the primal problem and also the results that obtained from the geometrical and mathematical analysis.

The values of the Lagrange multipliers ($\alpha_i$) can also can be calculated from Equation (\ref{EQN:SVMOptDual2}) subject to $\sum_i \alpha_iy_i=0$, and hence $L_D$ will be
\begin{align*}
\text{min } L_D&=\frac{1}{2}\left [4\alpha_1^2+36\alpha_2^2-24\alpha_1\alpha_2  \right ]\nonumber \\
&-\left [\alpha_1+\alpha_2 \right ]-\lambda(\alpha_2-\alpha_1)
\end{align*}
where $\lambda$ is the Lagrange multiplier, the first term $\frac{1}{2}\left [4\alpha_1^2+36\alpha_2^2-24\alpha_1\alpha_2  \right ]$ represents $\frac{1}{2} \alpha^T \textbf{H} \alpha$ in Equation (\ref{EQN:SVMOptDual2}), the term $\left [\alpha_1+\alpha_2 \right ]$ is $\textbf{f}^T\alpha$ in Equation (\ref{EQN:SVMOptDual2}), and finally the term $\lambda(\alpha_2-\alpha_1)$ is the constraint $\sum_i \alpha_iy_i=0$. The solutions can be calculated by differentiating $L_D$ with respect to $\alpha_1$ and $\alpha_2$ a follows:
\begin{align*}
\frac{\partial L_D}{\partial \alpha_1}&=0\Rightarrow 4\alpha_1-12\alpha_2-1+\lambda=0 \nonumber \\
\frac{\partial L_D}{\partial \alpha_2}&=0\Rightarrow -12\alpha_1+36\alpha_2-1-\lambda=0
\end{align*}
and $\sum_i\alpha_iy_i=0$ means that $-\alpha_1+\alpha_2=0$ and hence $\alpha_1=\alpha_2$ and the above equations will be
\begin{align*}
 -8\alpha_1-1+\lambda=0 \nonumber \\
24\alpha_1-1-\lambda=0 \nonumber \\
\end{align*}
and the solution will be $\lambda=2$ and $\alpha_1=\alpha_2=\frac{1}{8}$ and these values are identical to the results that we obtained before\footnote{Using $\alpha_1$ and $\alpha_2$, we can calculate $\textbf{w}$ and $b$ as mentioned before.}. 

It is worth mentioning that if $N>d$ (i.e. the number of samples is larger than the dimension of the features space), the solution of Equation (\ref{EQN:SVMOptDual1}) will be not unique and hence there are many values of $\alpha$. More details are in the following section.

\paragraph{\textbf{Global solution and uniqueness}}
The values of $\textbf{w}$ and $b$ are not unique if the Hessian matrix is positive semidefinite and hence many points (i.e. combinations of $\textbf{w}$ and $b$) may have the same objective value. With a positive definite Hessian matrix, the solution of SVM (i.e. $\textbf{w}$ and $b$) is unique, but, the values of $\alpha$'s are different.

\begin{figure}[!ht]%2
\centering
\includegraphics[width=0.25\textwidth]{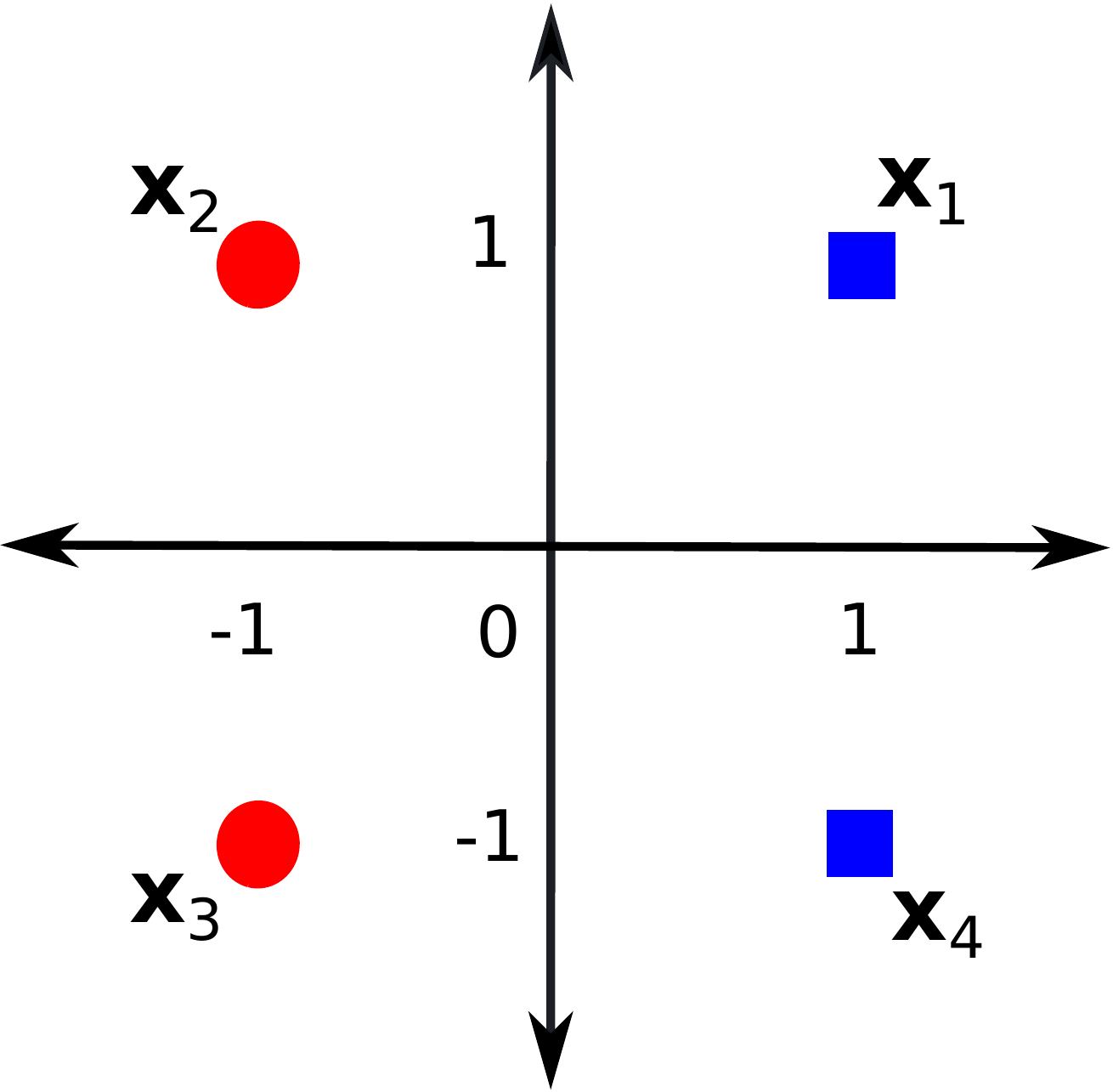} 
\caption{Visualization of the samples in our example.}
\label{fig:GlobalSoln}
\end{figure}

For example, given four samples in $R^2$: $\textbf{x}_1=[1, 1]$, $\textbf{x}_2=[-1, 1]$, $\textbf{x}_3=[-1, -1]$, and $\textbf{x}_4=[1, -1]$, and the class labels for these samples are as follows, $y=[+1, -1, -1, +1]$ (see Fig. \ref{fig:GlobalSoln})\footnote{We used the same values of the example in \cite{burges1998tutorial}, but, here we added more details, explanations, and visualizations.}. The value of $\textbf{H}$ is given by $\textbf{H} =\begin{bmatrix}
2 &0  &2 &0 \\ 
0 & 2 &0  &2 \\ 
2 &0  &2  &0 \\ 
0 & 2 &0  &2 \\ 
\end{bmatrix}$. From Equation (\ref{EQN:SVMOptDual2}), $L_D$ is given by
\begin{align*}
\text{min }L_D(\alpha)&=\frac{1}{2} \alpha^T \textbf{H} \alpha-\textbf{f}^T\alpha \nonumber \\
&=\frac{1}{2}(2\alpha_1^2+2\alpha_2^2+2\alpha_3^2+2\alpha_4^2+4\alpha_1\alpha_3+\alpha_2\alpha_4)\nonumber \\
& -(\alpha_1+\alpha_2+\alpha_3+\alpha_4)-\lambda (\alpha_1-\alpha_2-\alpha_3+\alpha_4)
\end{align*}

The solution is calculated by differentiating $L_D$ with respect to the four Lagrange multipliers ($\alpha_1$, $\alpha_2$, $\alpha_3$, and $\alpha_4$) and setting the derivatives to zero as follows:
\begin{align*}
\frac{\partial L_D}{\partial \alpha_1}=0 \rightarrow 2\alpha_1+2\alpha_3=1+\lambda \nonumber \\
\frac{\partial L_D}{\partial \alpha_2}=0 \rightarrow 2\alpha_2+2\alpha_4=1-\lambda \nonumber \\
\frac{\partial L_D}{\partial \alpha_3}=0 \rightarrow 2\alpha_1+2\alpha_3=1-\lambda \nonumber \\
\frac{\partial L_D}{\partial \alpha_4}=0 \rightarrow 2\alpha_2+2\alpha_4=1+\lambda 
\end{align*}

From the above equations, we can find different solutions/values of $\alpha$'s. For example, one solution is $\alpha_1=\alpha_2=\alpha_3=\alpha_4=\frac{1}{4}$ (with $\lambda=0$), and this means that all samples are support vectors and from Equation (\ref{EQN:SVMLPW}), the value of $\textbf{w}$ is $\textbf{w}=\sum_{i=1}^{N=4}\alpha_iy_i\textbf{x}_{i} =\frac{1}{4} \left [  (+1) \begin{bmatrix}
1\\ 
1
\end{bmatrix}+(-1) \begin{bmatrix}
-1\\ 
1
\end{bmatrix} +(-1) \begin{bmatrix}
-1\\ 
-1
\end{bmatrix}+(+1) \begin{bmatrix}
1\\ 
-1
\end{bmatrix}  \right ]
= \begin{bmatrix}
1\\ 0
\end{bmatrix}$ and $b$ is calculated according to Equation (\ref{EQN:bNSV}) as follows:
\begin{align*}
&\sum_{i=1}^{N_{SV}} (\frac{1}{y_i}-\textbf{w}^T\textbf{x}_i)= \left ( 1-\begin{bmatrix}
1 & 0
\end{bmatrix}\begin{bmatrix}
1\\ 
1
\end{bmatrix} \right ) \nonumber \\
&+\left ( -1-\begin{bmatrix}
1 & 0
\end{bmatrix}\begin{bmatrix}
-1\\ 
1
\end{bmatrix} \right ) +\left ( -1-\begin{bmatrix}
1 & 0
\end{bmatrix}\begin{bmatrix}
-1\\ 
-1
\end{bmatrix} \right )\nonumber \\
&+\left ( 1-\begin{bmatrix}
1 & 0
\end{bmatrix}\begin{bmatrix}
1\\ 
-1
\end{bmatrix} \right )=0
\end{align*}
and hence $b=\frac{1}{4}\sum_{i=1}^{N_{SV}} (\frac{1}{y_i}-\textbf{w}^T\textbf{x}_i)=0$.

Another solution that has the same $\textbf{w}$ and $b$ is that $\alpha_1=\alpha_2=\frac{1}{2}$ and $\alpha_3=\alpha_4=0$ (with $\lambda=0$). This means that the first two samples are considered as support vectors and the other two samples are training samples. Hence, the same $\textbf{w}$ and $b$ can be calculated using different values of $\alpha$'s and different numbers of support vectors. However, both solutions satisfy the constraints $\alpha_i\geq 0$ and $\sum_{i=1}^{N}\alpha_iy_i=0$.

\subsubsection{Non-separable data (Overlapping Classes)}
In the case of non-separable data or overlapped classes, more misclassified samples result. This will increase $\alpha_i$ for the misclassified training samples. As a result, the decision boundary will be affected to classify these samples correctly. In this case, most of the training samples are selected to be support vectors. Therefore, the constraints of linear SVM in Equation (\ref{EQN:ST}) must be relaxed by adding a non-negative slack\footnote{The slack variable is a variable that is added to the inequality constraint where a linear combination of variables is less than or equal to a given constant to convert it to an equality and non-negative constraint.} variable ($\epsilon_i$) as follows:
\begin{equation}
y_i(\textbf{w}^T\textbf{x}_i+b)-1+\epsilon_i \geq 0 \text{ , } \epsilon_i\geq 0\; 
\label{EQN:SVMEpsilon1}
\end{equation}
where $\epsilon_i$ is the distance between the $i^{th}$ training sample (see Fig. \ref{fig:SVMExampleLinear}) and the corresponding margin plane. Also, $\epsilon_i$ represent the marginal error of the $i^{th}$ training sample that permits a margin relaxation (soft margin) and it should be minimized. The data inside the soft margin or soft SVM are neglected \cite{kecman2001learning}. 

The objective function of SVM after adding $\epsilon_i$ will be as follows:
\begin{align}
\text{min } L(\textbf{w},b,\epsilon)&=\frac{1}{2} \left \| \textbf{w} \right \|^2 \nonumber \\
&+ C(\sum\text{ distances of misclassified samples}) \nonumber \\ 
&=\frac{1}{2} \left \| \textbf{w} \right \|^2+ \frac{C}{k}\sum_{i=1}^{N} \epsilon_i^k \nonumber \\ 
&\text{ s.t. } y_i(\textbf{w}^T\textbf{x}_i+b) -1+\epsilon_i \geq 0,\;\;\epsilon_i\geq 0 
\label{EQN:SVMERROR}
\end{align}
where $\forall i=1,2,\dots,N$, $k$ is a positive integer, $C$ represents the regularization or penalty parameter and it controls the trade-off between the size of the margin ($\frac{1}{2} \left \| \textbf{w} \right \|^2$) and the slack variable penalty or the training error ($\frac{C}{k}\sum_{i=1}^{N} \epsilon_i^k$). The term $\frac{C}{k}\sum_{i=1}^{N} \epsilon_i^k$ represents also the distance of error samples to their correct place. As a consequence, a small value of $C$ allows the constraints to be easily ignored, i.e. large or hard margin, while a large $C$ makes the constraints difficult to ignore, i.e. soft margin. In Equation (\ref{EQN:SVMERROR}), all constraints can be satisfied if $\epsilon_i$ is sufficiently large and the optimization problem still quadratic and there is a unique solution but with $2N$ linear inequality constraints \cite{kecman2001learning}. Equation (\ref{EQN:SVMERROR}) is formalized into Lagrange formula as follows:
\begin{align}
\text{min } L_P(\textbf{w},&b,\epsilon,\alpha,\mu)= \frac{1}{2} \left \| \textbf{w} \right \|^2+ \frac{C}{k}\sum_{i=1}^{N} \epsilon_i^k \nonumber \\
&-\sum_{i=1}^{N}\alpha_i[y_i(\textbf{w}^T\textbf{x}_i+b)-1+\epsilon_i]-\mu_i \epsilon_i
\label{EQN:SVMERROR2}
\end{align}
where $\alpha_i$ and $\mu_i$ are non-negative Lagrange multipliers. Hence, $L_P$ has to be minimized with respect to $\textbf{w}$, $b$, and $\epsilon_i$ and maximized with respect to $\alpha_i$ and $\mu_i$. With $k=1$ (this is called \emph{L1-SVM}), and by differentiating $L_P$ with respect to $\epsilon_i$ and setting the derivatives to zero (as in Equations (\ref{EQN:SVMLPW}, \ref{EQN:SVMLPb})) we found that
\begin{equation}
\frac{\partial L_P}{\partial \epsilon_i}=0 \Rightarrow C=\alpha_i+\mu_i
\label{EQN:SVMLEp}
\end{equation} 
and the KKT complementary condition will be
\begin{align}
\alpha_i\{y_i(\textbf{w}^T\textbf{x}_i+b) -1+\epsilon_i \} =0, \; i=1,\dots,N \nonumber \\
\mu_i\epsilon_i=(C-\alpha_i)\epsilon_i=0 , \; i=1,\dots,N
\label{EQN:SVMPrimalEpsilon}
\end{align}

The dual problem can be formulated by substituting Equations (\ref{EQN:SVMLPW}, \ref{EQN:SVMLPb}, and \ref{EQN:SVMLEp}) into Equation (\ref{EQN:SVMERROR2}), the dual problem can be written as follows:
\begin{align*}
\text{max } L_D(\alpha)&= \frac{\left \|  \textbf{w}\right \|^2}{2}+C\sum_i \epsilon_i  \nonumber \\
&-\sum_{i} \alpha_i (y_i(\textbf{w}^T\textbf{x}_{i}+b)-1+\epsilon_i)-\mu_i \epsilon_i \nonumber \\
&=  \underset{\frac{\left \|  \textbf{w}\right \|^2}{2}}{\underbrace{\frac{(\sum_{i=1}^{N}\alpha_iy_i\textbf{x}_{i})^2}{2}}}+C \sum_i\epsilon_i \nonumber \\ 
&-\sum_{i} \alpha_i (y_i( \underset{ \textbf{w}}{\underbrace{\sum_{j=1}^{N}\alpha_jy_j\textbf{x}_{j}}} \textbf{x}_{i}+b)-1+\epsilon_i)-\mu_i \epsilon_i\nonumber \\
&=\frac{(\sum_{i=1}^{N}\alpha_iy_i\textbf{x}_{i})^2}{2}+C\sum_i\epsilon_i+ \sum_{i} \alpha_i- \underset{0}{\underbrace{\sum_{i} \alpha_i y_i b}} \nonumber \\
&- \sum_{i,j} \alpha_i\alpha_j y_iy_j\textbf{x}_i\textbf{x}_j\underset{-C\sum_i \epsilon_i}{\underbrace{-\sum_i\alpha_i\epsilon_i-\mu_i\epsilon_i}} \nonumber \\
&=\sum_{i=1}^{N} \alpha_i -\frac{1}{2} \sum_{i,j} \alpha_i \alpha_j y_i y_j\textbf{x}_i^T\textbf{x}_j\nonumber  \\  
\text{ s.t.} \; &\alpha_i\geq 0 \;\; , \sum_{i=1}^{N}\alpha_i y_i=0 \; \forall i=1,2,\dots,N 
%\label{EQN:SVMOErrorL1}
\end{align*} 

Hence, neither the $\epsilon_i$ nor their Lagrange multipliers appear in the dual form and hence the dual problem for the overlapped data will be identical to the linear separable data in Equation (\ref{EQN:SVMOptDual1}). The only difference between the optimization problem of the separable and non-separable cases is that $\alpha_i$ and $\mu_i$ are upper-bounded by $C$ as indicated in Equation (\ref{EQN:SVMLEp}). From Equation (\ref{EQN:SVMLEp}), it can be remarked that SVs with $\alpha_i=C$ lie outside the margin or on the margin boundary. The value of $C$ is determined by the user. There are three possible solutions for $\alpha_i$:
\begin{itemize}
\item With $\alpha_i=0$ and $\epsilon_i=0$, this means that the $i^{th}$ sample is not a support vector and it is correctly classified.
\item $C>\alpha_i>0$; thus, the corresponding sample is a support vector this is because the complementary condition in Equation (\ref{EQN:SVMPrimalEpsilon}) leads to $y_i(\textbf{w}^T\textbf{x}_i+b) -1+\epsilon_i=0$. With $\epsilon_i=0$; thus, $y_i(\textbf{w}^T\textbf{x}_i+b) =1$ and hence $\textbf{x}_i$ is called \emph{unbounded} or \emph{free} support vector and this sample lies on the margin, i.e. correctly classified.
\item $\alpha_i=C$, this means that $\alpha_i$ reached to the upper-bound ($C$), and according to the complementary condition in Equation (\ref{EQN:SVMPrimalEpsilon}), $y_i(\textbf{w}^T\textbf{x}_i+b) -1+\epsilon_i=0$, and if
\begin{itemize}
\item $\epsilon_i\geq 1$, according to Equation (\ref{EQN:SVMEpsilon1}) the value of $ -1+\epsilon_i$ will be positive; thus, $y_i(\textbf{w}^T\textbf{x}_i+b)$ is negative, then the decision function ($\textbf{w}^T\textbf{x}_i + b$) and the class label ($y_i$) have different signs, indicating that $\textbf{x}_i$ is misclassified. 
\item $1>\epsilon_i> 0$, the sample is in between the margin and the correct side of hyperplane, i.e. correctly classified.
\item $\epsilon_i=0$ means that $\textbf{x}_i$ is a support vector and this is called \emph{bounded} support vector and it lies on the wrong side of the margin.
\end{itemize}
\end{itemize}

With $k=2$ (\emph{L2-SVM}), the term $\frac{C}{k}\sum_{i=1}^{N} \epsilon_i^k$ in Equation (\ref{EQN:SVMERROR}) will be $\frac{C}{2}\sum_{i=1}^{N} \epsilon_i^2$ and hence the last term in Equation (\ref{EQN:SVMERROR2}) (i.e. $\mu_i\epsilon_i$) is not necessary; thus, there are no longer complementarity constraints ($\mu_i \epsilon_i=(C-\alpha_i)\epsilon_i=0$) in Equation (\ref{EQN:SVMPrimalEpsilon}) \cite{lin2001formulations}. Therefore, Equation (\ref{EQN:SVMERROR2}) will be
\begin{align}
\text{min } L_P(\textbf{w},b,\epsilon,\alpha,\mu)&= \frac{1}{2} \left \| \textbf{w} \right \|^2+ \frac{C}{2}\sum_{i=1}^{N} \epsilon_i^2 \nonumber \\
&-\sum_{i=1}^{N}\alpha_i[y_i(\textbf{w}^T\textbf{x}_i+b)-1+\epsilon_i]
\label{EQN:SVMERROR2L2}
\end{align} 
and hence 
\begin{equation*}
\frac{\partial L_P}{\partial \epsilon_i}=0 \Rightarrow C\epsilon_i-\alpha_i=0 \Rightarrow C\epsilon_i=\alpha_i
%\label{EQN:SVMLEpL2}
\end{equation*} 

From the above equation and Equation (\ref{EQN:SVMERROR2L2}), the terms $\frac{C}{2}\sum_{i=1}^{N} \epsilon_i^2-\sum_{i=1}^{N}\alpha_i \epsilon_i$ will be $\frac{C}{2}\sum_{i=1}^{N}(\frac{\alpha_i}{C})^2-\sum_{i=1}^{N}\frac{\alpha_i^2}{C}=-\frac{\alpha_i^2}{C}$ and the objective function of the dual problem will be
\begin{align*}
\text{max } L_D(\alpha)&=\sum_{i=1}^{N} \alpha_i -\frac{1}{2} \sum_{i,j} \alpha_i \left ( H_{ij}+\frac{1}{C} \right) \alpha_j \nonumber  \\
&=\textbf{f}^T\alpha -\frac{1}{2} \alpha^T \left (\textbf{H}+\frac{1}{C}\textbf{I}\right) \alpha\nonumber \\  &\text{ s.t. } \alpha_i\geq 0 \;,\; \alpha \textbf{y}^T=0
%\label{EQN:SVMOptDualL2}
\end{align*} 
where $\textbf{I}$ is the identity matrix. The only difference between the above equation and Equation (\ref{EQN:SVMOptDual}) is the term $\frac{1}{C}\textbf{I}$ which is added to $\textbf{H}$ and this means that $\frac{1}{C}$ is added to the diagonal entries of $\textbf{H}$ and this ensuring its positive definiteness and stabilizing the solution than the L1-SVM. Moreover, the number of support vectors of L1-SVM is less than L2-SVM; in other words, L1-SVM produces more sparse solutions. In L2-SVM, there is no upper bound for $\alpha_i$ and the only requirement is $\alpha_i$ to be nonnegative \cite{shigeo2005support}. Maximizing the above equation is equal to 
\begin{align}
\text{min } L_D(\alpha)&= \frac{1}{2} \alpha^T\left ( \textbf{H}+\frac{1}{C}\textbf{I} \right )\alpha-\textbf{f}^T \alpha
\label{EQN:L2normDual}
\end{align}
subject to the same constraints.

\subsubsection{Nonlinear separable data}
If the data are non-linearly separable, the kernel functions can be used for transforming the data from the current/input space to a higher-dimensional space using a nonlinear function ($\phi$), where the data can be linearly separable. The kernel function is defined as the dot product of nonlinear functions as follows, $K(\textbf{x}_i,\textbf{x}_j)=\phi(\textbf{x}_i)^T\phi(\textbf{x}_j)$ \cite{AlaaParameter}. The objective function of the SVM classifier will be
\begin{align}
\text{min } &L(\textbf{w},b)=\frac{1}{2} \left \| \textbf{w} \right \|^2+ \frac{C}{k}\sum_{i=1}^{N} \epsilon_i^k \nonumber \\
\text{ s.t. }  &y_i(\textbf{w}^T\phi(\textbf{x}_i)+b) -1+\epsilon_i \geq 0 \;\; \forall i =1,2,\dots,N 
\label{EQN:SVMERROR3}
\end{align}
and the dual form will be identical to Equation (\ref{EQN:SVMOptDual2}) subject to the same constraints, and the only difference is that $H_{ij}=\sum_{i,j}\alpha_i \alpha_j y_iy_i K(\textbf{x}_i,\textbf{x}_j)$ \cite{scholkopf2001learning}.

In SVM, the most well-known kernel functions are:
\begin{itemize}
\item \emph{Linear} kernel, $K(\textbf{x}_i,\textbf{x}_j)=\left \langle \textbf{x}_i,\textbf{x}_j \right \rangle$, this kernel is the same as the original input space,
\item \emph{Radial basis function} (RBF) or \emph{Gaussian} kernel, $K(\textbf{x}_i,\textbf{x}_j)=\text{exp}({-||\textbf{x}_i-\textbf{x}_j||^2}/{2\sigma^2})$, and
\item \emph{Polynomial} kernel of degree $d$, $K(\textbf{x}_i,\textbf{x}_j)=(\left \langle \textbf{x}_i,\textbf{x}_j \right \rangle )^d$. 
\end{itemize}

\begin{figure*}[!ht]%2
\centering
\includegraphics[width=0.85\textwidth]{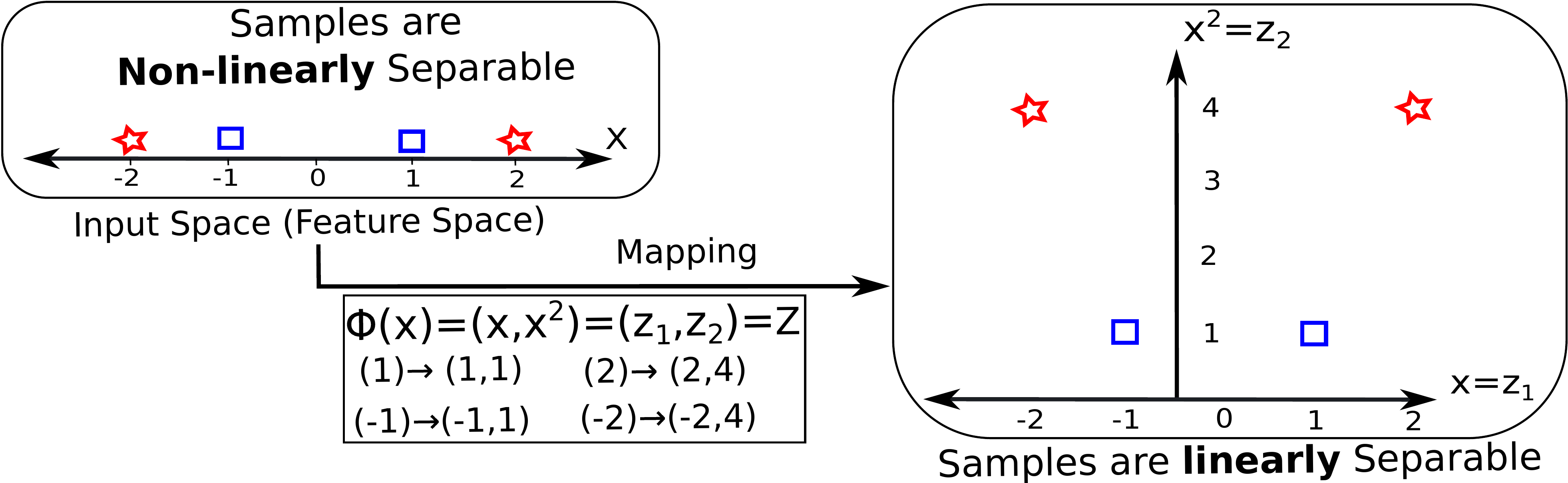} 
\caption{An example of the kernel function. }
\label{fig:Kernel}
\end{figure*}

Figure \ref{fig:Kernel} shows an example to explain how the kernel function is used to map the data ($X\in \mathcal{R}^n$) into a higher dimensional space ($F$). The figure shows the input data ($X\in \mathcal{R}^1$) which consists of two nonlinearly separable classes. Each class has two samples. The kernel function is used for mapping the data from the input space where the data cannot be linearly separable to a new feature space $F \in \mathcal{R}^d$, $d>n$ (in the figure $d=2$), as follows, $\phi: {X} \rightarrow {F}$. The data in the new feature space are linearly separable and then we can apply the standard SVM \cite{scholkopf1999advances}. 

Calculating the scalar product ($\phi ^T(\textbf{x}_i) \phi (\textbf{x}_j)$) may lead to the \emph{curse of dimensionality} problem and need a high computational efforts. This is because the scalar product $\textbf{x}^T_i \textbf{x}_j$ in Equation (\ref{EQN:SVMOptDual}) is replaced by the scalar product $\phi (\textbf{x}_i)^T \phi (\textbf{x}_j)$ in the feature space. However, this problem can be handled because the kernel function is a function in the feature space and hence the mapping step can be avoided because the mapping will be calculated directly by calculating the kernel $K(\textbf{x}_i,\textbf{x}_j)$ for the training data in the input space, this is the so-called the \emph{kernel trick} \cite{ben2010user,kecman2001learning}. In other words, the kernel function can calculate the transformation of the data into the new space without explicitly visiting it (the new space) and this reduces the required computational efforts. The kernel trick can be explained using the following example. Given $\textbf{x}, \textbf{y} \in \mathcal{R}^2$, where $\textbf{x}=[x_1\; x_2]^T$ and $\textbf{y}=[y_1\; y_2]^T$, and let the transformation function is $\phi (\textbf{x})=[x_1^2\; \sqrt{2} x_1x_2 \;x_2^2]$, $\phi: \mathcal{R}^2 \rightarrow \mathcal{R}^3$. The dot product can be calculated as follows:
\begin{align}
\phi^T (\textbf{x})\phi (\textbf{y}) &=[x_{1}^2\; \sqrt{2}x_{1}x_{2}\; x_{2}^2][y_{1}^2 \;\sqrt{2}y_{1}y_{2}\; y_{2}^2]^T \nonumber \\
&=x_{1}^2y_{1}^2+ 2x_{1}x_{2}y_{1}y_{2}+ x_{2}^2y_{2}^2 \nonumber \\
&=(x_{1}y_{1}+ x_{2}y_{2})^2 =(\textbf{x}.\textbf{y})^2=K(\textbf{x},\textbf{y})
\label{EQN:Kernel}
\end{align}
From the above equation, there is no need to perform mapping for each sample; instead, $(\textbf{x}_i^T\textbf{x}_j)^2$ is calculated in the input feature space. It is worth mentioning that two different representations may correspond to the same kernel. For example, the transformation function $\phi(\textbf{x})=[x_{1}^2\; x_{1}x_{2}\; x_{2}x_{1}\; x_{2}^2]$ has the same kernel function of the transformation function in Equation (\ref{EQN:Kernel}).

\begin{figure}[!ht]%2
\centering
\includegraphics[width=0.5\textwidth]{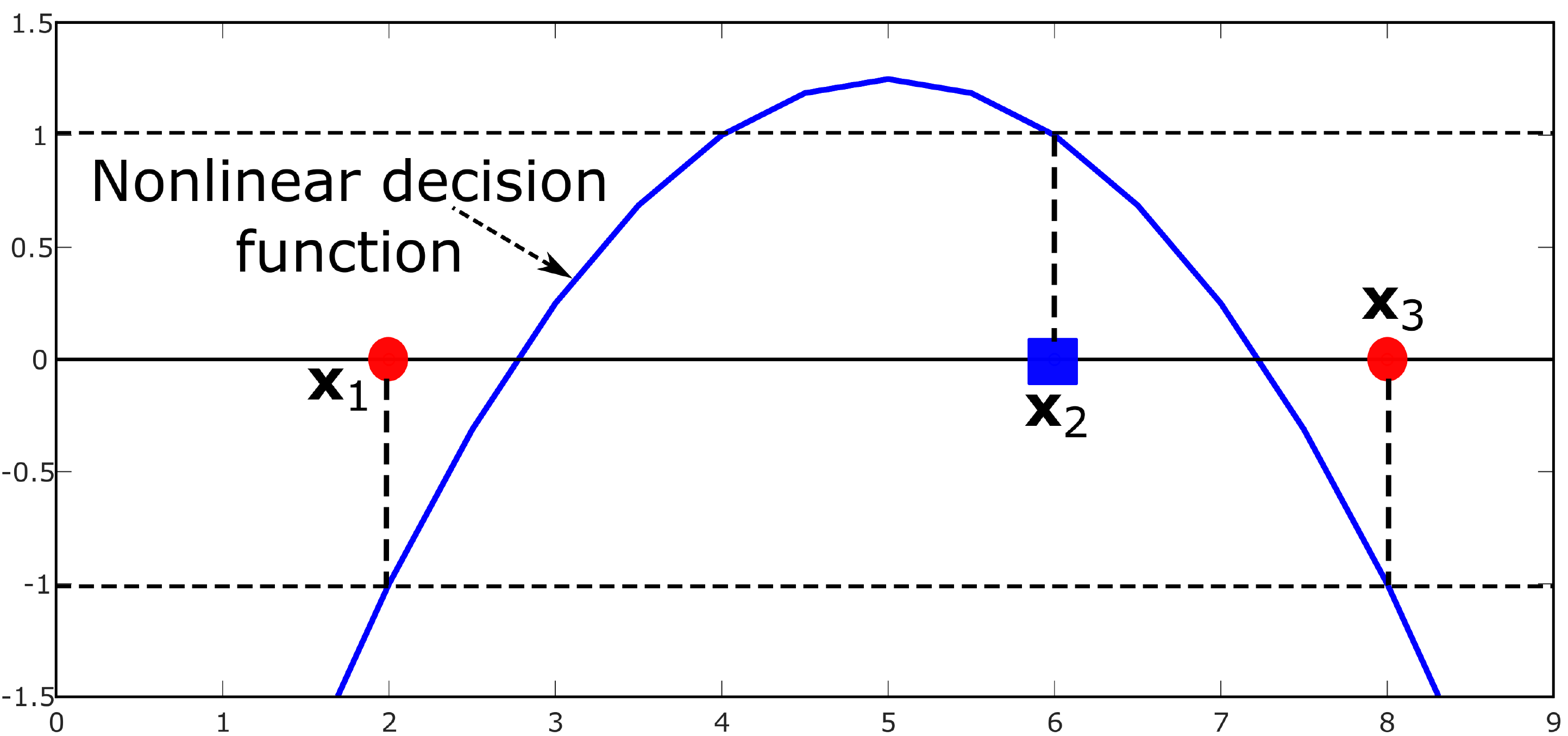} 
\caption{The nonlinear decision function (i.e. hyperplane) for our example. The negative class has two samples with the red color and the positive class has only one sample with the blue color.}
\label{fig:KernelExample}
\end{figure}

\paragraph{\textbf{Kernel example}}
The goal of this simple example is to explain how the kernel function is used for mapping two nonlinearly separable data into higher dimensional space, where the data can be linearly separable. Given three samples as shown in Fig. \ref{fig:KernelExample}, where the positive class has one sample ($\textbf{x}_2=6$) and the negative class has two samples ($\textbf{x}_1=2$ and $\textbf{x}_3=8$). In this example, we used a simple polynomial kernel with degree $d=2$ and the kernel function is ($K(x,y)=\phi(\textbf{x}).\phi(\textbf{y})=(\textbf{x}.\textbf{y}+1)^2$, where $\phi(x)=\begin{bmatrix} x^2 &\sqrt{2}x&1
\end{bmatrix}$). The matrix $\textbf{H}$ of this data is as follows:
\begin{align*}
\textbf{H} =\begin{bmatrix}
25 &-169  &289 \\ 
-169 & 1369 &-2401 \\ 
289 &-2401 &4225 \\ 
\end{bmatrix}
\end{align*}

In the above equation, $\textbf{H}=y_iy_jK(\textbf{x}_i,\textbf{x}_j)$. For example, $H(1,1)=y_1y_1K(\textbf{x}_1,\textbf{x}_1)=-1\times -1\times K(2,2)=(2\times 2 +1)^2=25$ and similarly all the other elements of $\textbf{H}$ can be calculated. From Equation (\ref{EQN:SVMOptDual2}), $L_D$ is given by
\begin{align*}
\text{min }L_D(\alpha)&=\frac{1}{2}(25\alpha_1^2+1369\alpha_2^2+4225\alpha_3^2\nonumber \\
& -338 \alpha_1\alpha_2+578\alpha_1\alpha_3-4802\alpha_2\alpha_3)\nonumber \\
&-(\alpha_1+\alpha_2+\alpha_3 )-\lambda(\alpha_2-\alpha_1-\alpha_3)
\end{align*}
subject to $\sum_i\alpha_iy_i=0$ (i.e. $-\alpha_1+\alpha_2-\alpha_3=0$), where $\lambda$ is a Lagrange multiplier. The solution is calculated by differentiating $L_D$ with respect to the three Lagrange multipliers ($\alpha_1$, $\alpha_2$, and $\alpha_3$) and setting the derivatives to zero as follows:
\begin{align*}
\frac{\partial L_D}{\partial \alpha_1}&=0 \rightarrow 25\alpha_1-169\alpha_2+289\alpha_3=1+\lambda \nonumber \\
\frac{\partial L_D}{\partial \alpha_2}&=0 \rightarrow -169\alpha_1+1369\alpha_2-2401\alpha_3=1-\lambda \nonumber \\
\frac{\partial L_D}{\partial \alpha_3}&=0 \rightarrow 289\alpha_1-2401\alpha_2+4225\alpha_3=1+\lambda
\end{align*}
hence the solution will be $\alpha_1=0.7396$, $\alpha_2=1.5938$, $\alpha_3=0.8542$, and $\lambda=-5$. As a result, the three samples in this example are support vectors. The hyperplane equation is calculated as follows: $\sum_{i=1}^3 y_i\alpha_iK(\textbf{x}_i,x)+b=0.7396(-1)(2x+1)^2-1.5938(+1)(6x+1)^2-0.8542(-1)(8x+1)^2+b=-0.25x^2+2.5x+b$. The value of $b$ is calculated using one of the points (($2,-1$), ($6,+1$), or ($8,-1$)), and the value of $b$ is -5; thus, the hyperplane equation is $-0.25x^2+2.5-5$. Figure \ref{fig:KernelExample} shows the nonlinear (quadratic) hyperplane. As reported in \cite{kecman2005support}, the decision function is also can be calculated as follows:
\begin{align*}
\sum_{i=1}^3 & y_i\alpha_i \phi(x_i)\phi(x)+b =\phi(\textbf{x})( 0.7396(-1)\begin{bmatrix} 4 &4&1 \end{bmatrix}\nonumber \\
&+1.5938(+1)\begin{bmatrix} 36 &12&1 \end{bmatrix} +0.8542(-1)\begin{bmatrix} 64 &16&1 \end{bmatrix})+b\nonumber \\
&=\begin{bmatrix} x^2 &\sqrt{2} x&1 \end{bmatrix}\left [-\frac{1}{4}+\frac{2.5}{\sqrt{2}}+0  \right ]+b
\end{align*}

As a consequence, the hyperplane equation is $-\frac{1}{4}x^2+2.5x+b$ and substitute in one of the points ($(2,-1)$, $(6,+1)$, or $8,-1$) to calculate the value of $b$. The value of $b$ is -5 and hence the hyperplane equation is $-0.25x^2+2.5-5$ and this is the same hyperplane we obtained in Fig. \ref{fig:KernelExample}.

\section{SVM parameter optimization}
\label{Sec:ParameterOpt}
This section introduces different experiments and mathematical explanation to show the effect of the SVM parameters on the classification performance with the balanced and imbalanced data. In all experiments, we used a simple two-dimensional binary classification data to visualize the correctly classified samples, misclassified samples, SVs, decision boundary, and margin borders. 

%http://www.svms.org/parameters/
\begin{figure*}[!ht]
\centering
%\subfloat[$C=0.0001$ \label{LDB:p0001}]{\includegraphics[width=0.330\textwidth]{Linearp0001.eps}}\hfill
\subfloat[$C=0.01$ \label{LDB:p01}]{\includegraphics[width=0.330\textwidth]{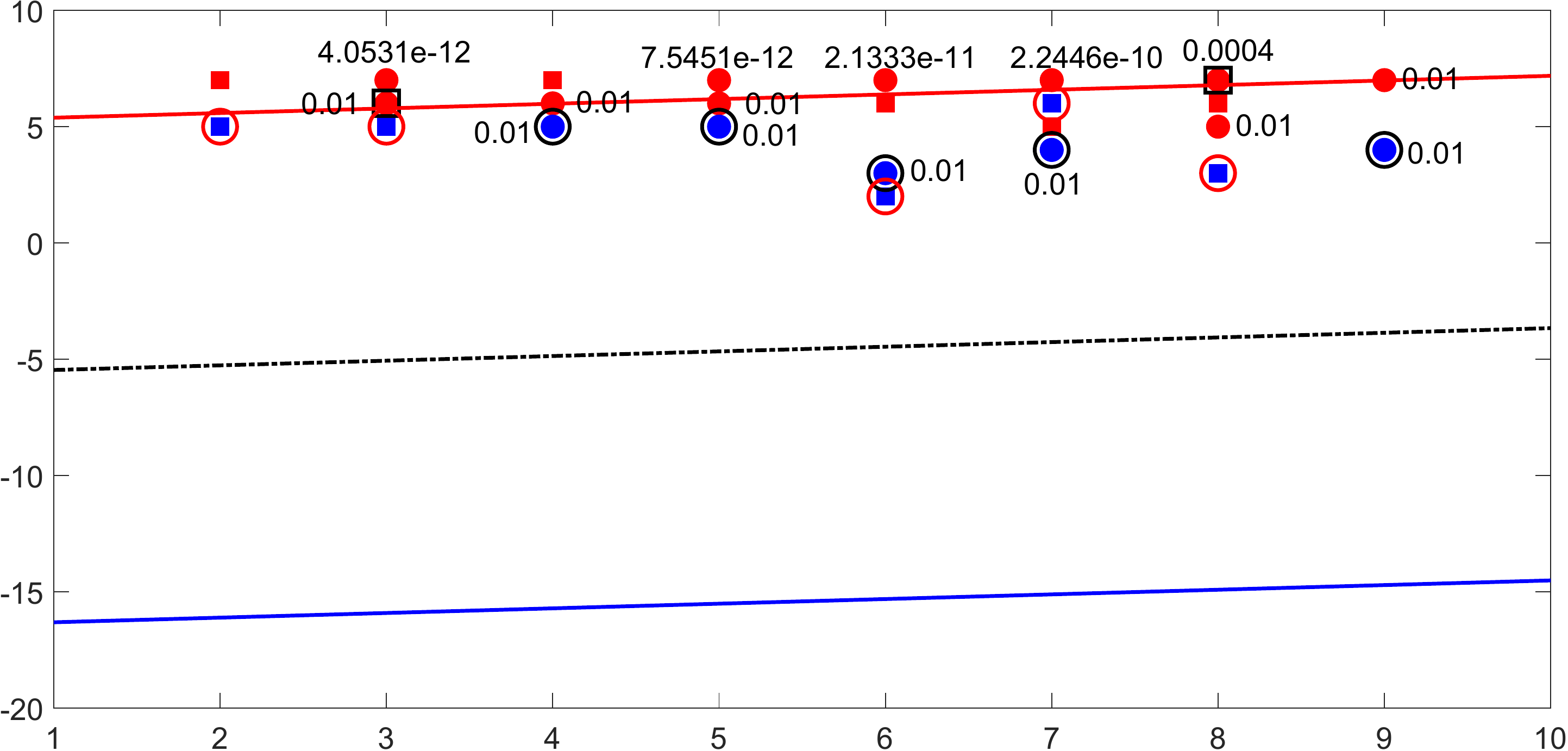}}\hfill
\subfloat[$C=0.05$ \label{LDB:p05}]{\includegraphics[width=0.330\textwidth]{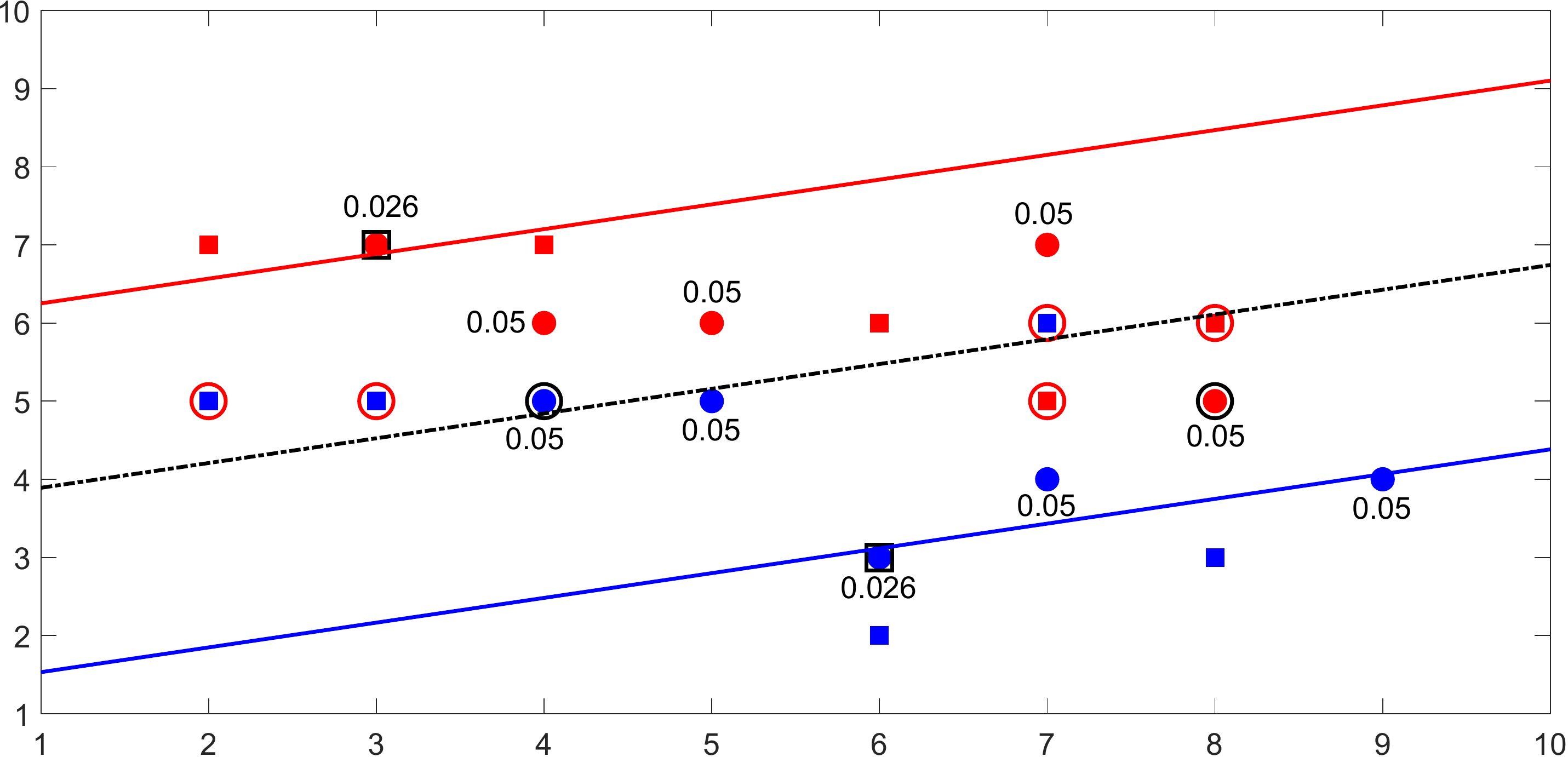}}\hfill
\subfloat[$C=1$ \label{LDB:p1}]{\includegraphics[width=0.330\textwidth]{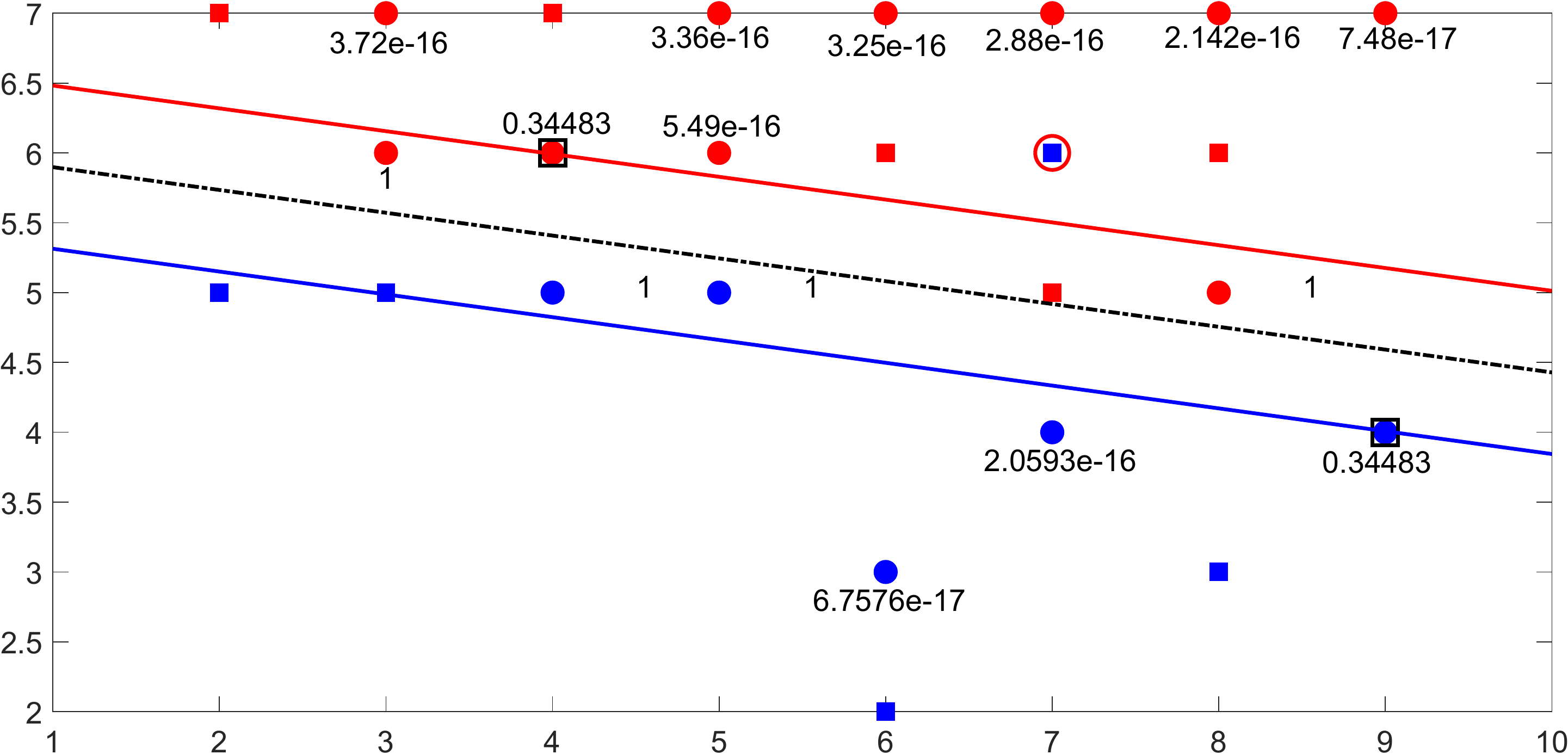}}\hfill
\subfloat[$C=5.78$ \label{LDB:5p78}]{\includegraphics[width=0.330\textwidth]{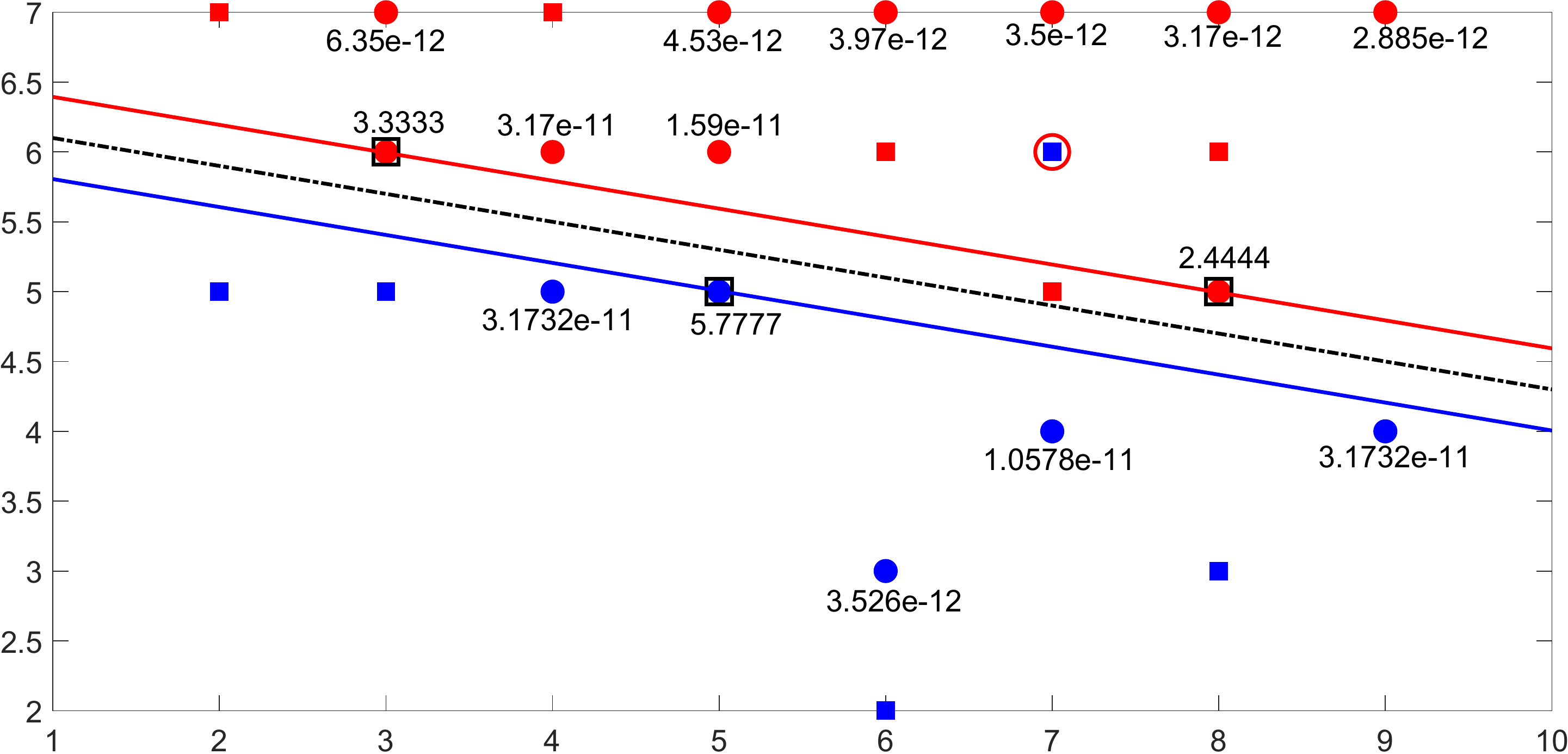}}\hfill
\subfloat[$C=10$ \label{LDB:10}]{\includegraphics[width=0.330\textwidth]{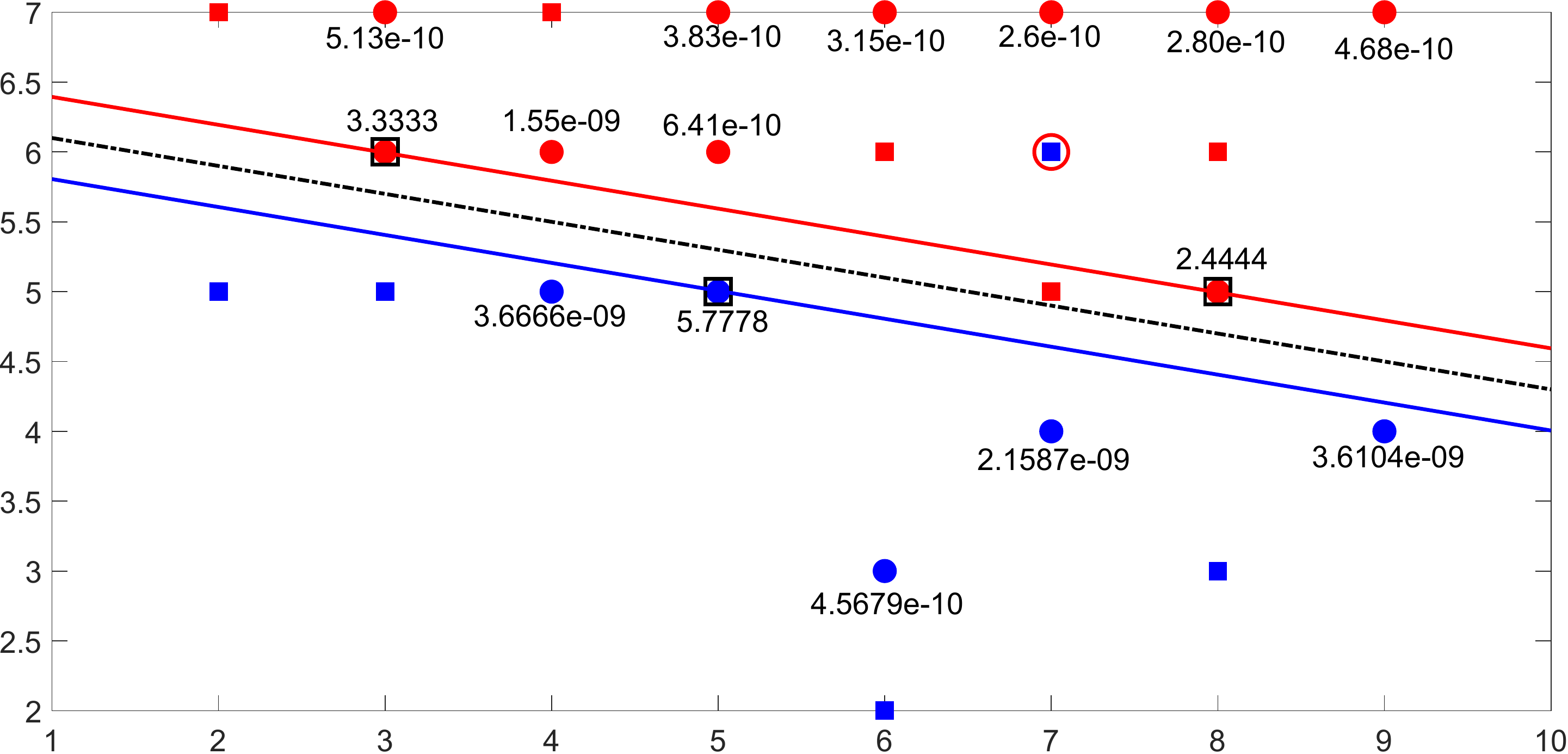}}\hfill
\subfloat[$C=100$ \label{LDB:100}]{\includegraphics[width=0.330\textwidth]{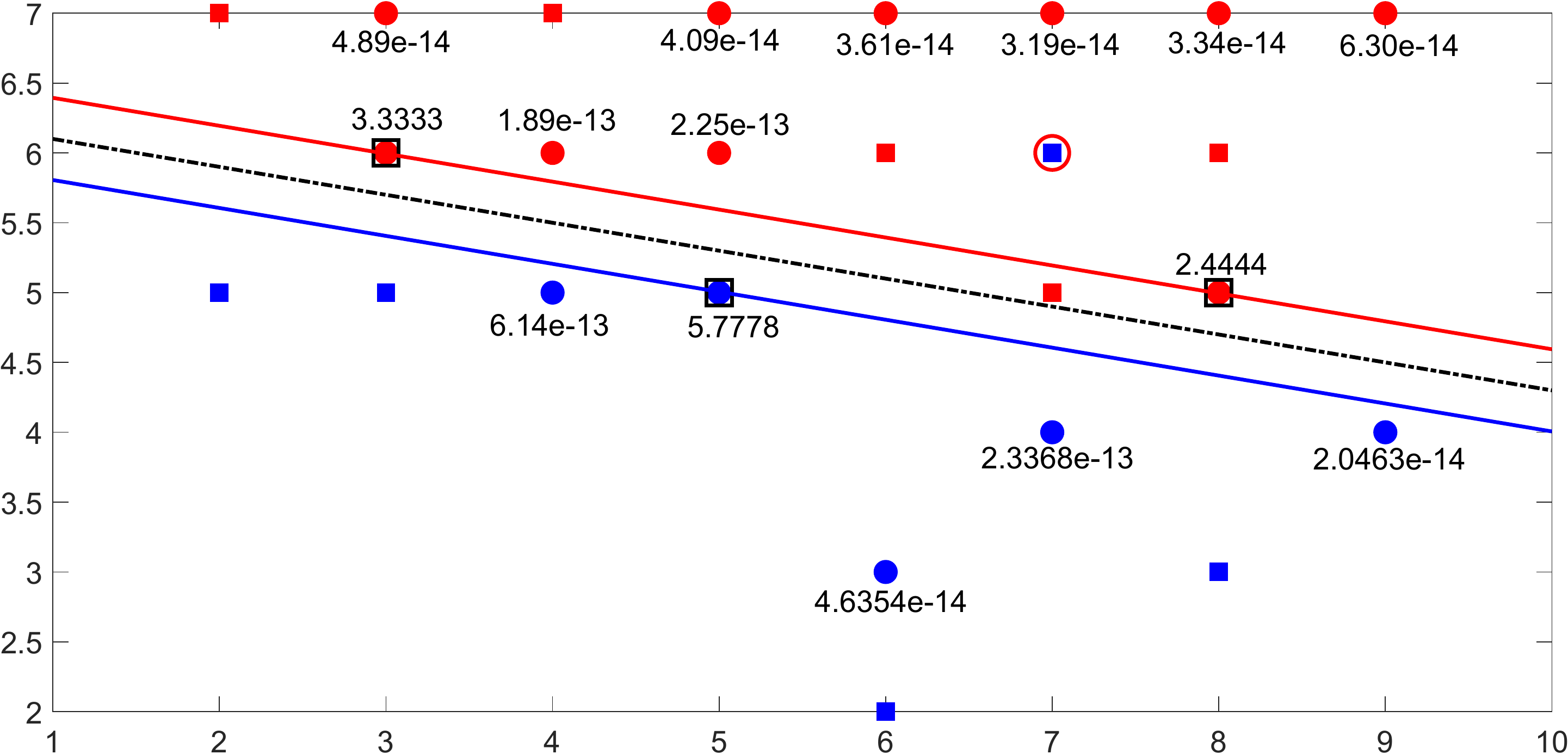}}\hfill
\caption{The influence of the penalty parameter ($C$) on the performance of linear SVM. The filled circles represent the training samples, filled squares are the testing samples, decision boundaries are the dotted black lines, the red and blue lines indicate the two planes, support vectors samples from both classes are marked by surrounding it by squares, the misclassified training and testing samples are marked by surrounding it by black and red circles, respectively, and the value of $\alpha$ for each training sample is reported.}
\label{fig:LinearSVM}
\end{figure*}

\subsection{Linear kernel function}
\label{SubSEc:Linear}
The linear kernel function or simply linear SVM is the simplest kernel function and it has no parameters as follows, $K(\textbf{x}_i,\textbf{x}_j)=\left \langle \textbf{x}_i,\textbf{x}_j \right \rangle$. Thus, we can say the linear kernel adds nothing to the SVM classifier. The SVM classifier with the linear kernel has only the penalty or regularization parameter ($C$) \cite{lin2001formulations}. As mentioned before, the parameter $C$ has a great influence on the classification performance of SVM, where it controls the trade-off between maximizing the size of the margin and minimizing the training error (see Equation (\ref{EQN:SVMERROR})) \cite{wang2005support}. As a consequence, changing the value of $C$ changes the optimal hyperplane; as a consequence, changes the training error, the number of support vectors, the testing error, and the margin of SVM. 

\begin{table*}[!ht]
\centering
\caption{The number of misclassified training samples (\# TrE.), number of misclassified testing samples (\# TsE.), SVM margin, and number of support vectors (\# SVs) of the linear kernel SVM using different values of $C$.}
\label{Table:LinearSVM}
\resizebox{0.8\textwidth}{!}{%
\begin{tabular}{|l|l|l|l|l|l|l|}
\hline
Results & $C=0.01$ &$C=0.05$ & $C=1$ & $C=5.78$ & $C=10$  & $C=100$  \\ \hline
\# TrE.  & 5 & 2      & 0     & 0     & 0    & 0    \\ \hline
\# SVs   & 2 & 2      & 2     & 3     & 3    & 3    \\ \hline
Margin  &21.70& 4.34   & 1.17  & 0.59  &0.59 & 0.59 \\ \hline
\# TsE.  &5  & 3      & 1     & 1    & 1    & 1       \\ \hline
\end{tabular}}
\end{table*}

\subsubsection{Illustrative example}
\begin{figure}[!t]
\centering
{\includegraphics[width=0.5\textwidth]{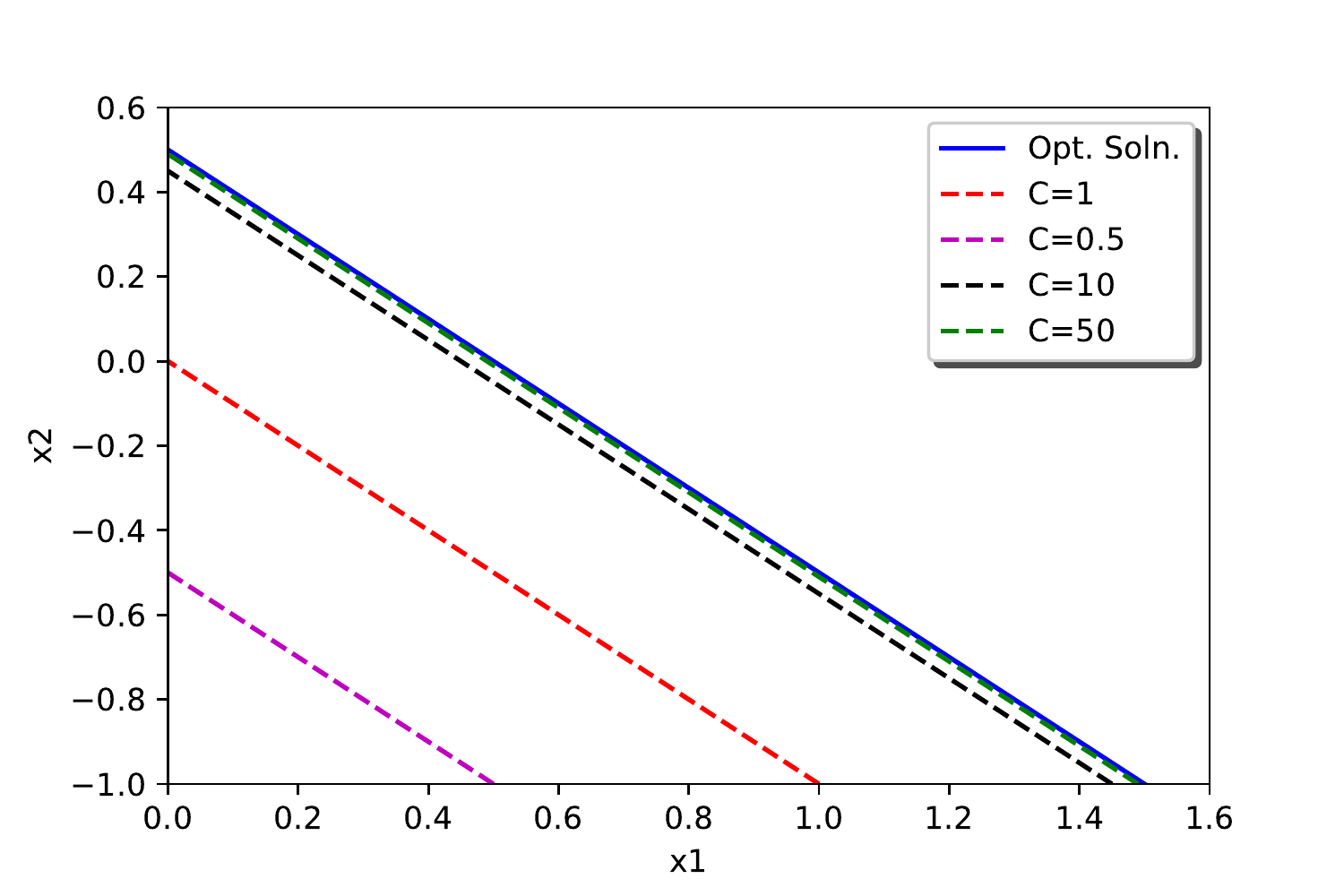}}\hfill
\caption{A visualization example to show how $C\rightarrow \infty$ converges to the optimal solution.}
\label{fig:CSimpleExample}
\end{figure}

Given a data with two classes and it consists of 25 samples; 15 samples for training ($N_1=10$ and $N_2=5$) and ten samples for testing; consequently, the data is imbalanced and the \emph{imbalance ratio} is $\frac{N_1}{N_2}=\frac{10}{5}=2$, where $N_1$ and $N_2$ represent the number of samples of the majority and minority classes, respectively. As shown in Fig. \ref{fig:LinearSVM}, the data is linearly separable and hence we will use the linear kernel. In this example, the classification performance was evaluated using different values of $C$. The results of this example are summarized in Table \ref{Table:LinearSVM} and Fig. \ref{fig:LinearSVM}. From these results, it can be remarked that:
\begin{itemize}
\item If $C$ is smaller than a certain limit then the minority class samples are misclassified; so, $\alpha_i=C \text{ with } y_i=-1$. However, as indicated in Equation (\ref{EQN:SVMLPb}), $\sum_{i=1}^{N}\alpha_iy_i=0$; Accordingly, there is at least one sample satisfies $\alpha_i<C$ (i.e. correctly classified), where $y_i=+1$. This means that this positive sample lies above or on the plane $H_1$. Decreasing $C$ increases the margin which means that the plane $H_2$ and the decision boundary ($\textbf{w}^{T}\textbf{x}_{i}+b = 0$) are forced to move away from $H_1$ where the training samples are located. As a consequence of that, the half space $\textbf{w}^{T}\textbf{x}_{i}+b \geq 0$ covers the whole training data. This is clear in Fig. \ref{LDB:p05} where the value of $C$ is 0.05 and as shown, there are two misclassified training samples and both samples are from the minority class. Decreasing the value of $C$ to be 0.01 increases the margin significantly from 4.34 (with $C=0.05$) to 21.7 (with $C=0.01$). Additionally, as in Fig. \ref{LDB:p01}, the positive samples lie on or above the positive plane ($H_1$) and with a small $C$ the other plane ($H_2$) and the optimal hyperplane are forced to move away from $H_1$. Decreasing $C$ to 0.01 increases the number of misclassified training samples to five samples which are the minority class samples. Hence, $C\rightarrow 0$, the number of misclassified training samples is $N_2$, which indicates the severe underfitting. 
\item If $C\rightarrow \infty$ this means that the margin will be reduced and $H_1$ and $H_2$ touch the decision boundary. From Fig. \ref{fig:LinearSVM} (d, e, and f) and Table \ref{Table:LinearSVM}, we found that the results when $C=$5.78, 10 and 100 were identical. As shown, in the three cases (i.e. $C=5.78$, $C=10$, and $C=100$), the same training error, testing error, margin width, number of support vectors were obtained. Hence, as reported in \cite{lin2001formulations}, increasing $C$ to be more than a specific limit ($C\geq C^*$) may have the same results. In other words, SVM with $C \rightarrow \infty$ approaches the optimal hyperplane. This can be explained using the following simple example. In this example, we used only three samples ($\textbf{x}_1=\begin{bmatrix} 0 &0 \end{bmatrix}$, $\textbf{x}_2=\begin{bmatrix}
1 & 0\end{bmatrix}$, and $\textbf{x}_3=\begin{bmatrix}
0 & 1\end{bmatrix}$), where the first sample belongs the negative class and the other two sample belong to the positive class\footnote{This example is introduced in \cite{lin2001formulations}; but, more details are added to make it clearer.}. Before training SVM, we can expect that the optimal hyperplane is $2x_1+ 2x_2-1=0$. In this example, we will use L2-SVM. As in Equation (\ref{EQN:L2normDual}), the optimal solution can be calculated as follows:
\begin{align*}
\begin{bmatrix}
\frac{1}{C} & 0 & 0\\ 
0 & 1+\frac{1}{C} &0 \\ 
0 & 0 & 1+\frac{1}{C}
\end{bmatrix} \begin{bmatrix}
\alpha_1\\ 
\alpha_2\\ 
\alpha_3
\end{bmatrix}+\lambda \begin{bmatrix}
-1\\ 
-1\\ 
1
\end{bmatrix}-\begin{bmatrix}
1\\ 
1\\ 
1
\end{bmatrix}=\begin{bmatrix}
\mu_1\\ 
\mu_2\\ 
\mu_3
\end{bmatrix}
\end{align*}
where $\lambda$ is the Lagrange multiplier for the constraint ($\textbf{Y}^T \alpha=0$) and $\mu_i$ represent the Lagrange multipliers of the constraints $\alpha_i\geq 0$. The three samples are support vectors so $\mu_1=\mu_2=\mu_3=0$; thus, the above equations can be written as follows:
\begin{align*}
\begin{bmatrix}
\textbf{0} & \textbf{Y}^T\\ 
\textbf{Y}& \frac{\textbf{I}}{C}+\textbf{H}
\end{bmatrix}\begin{bmatrix}
\lambda \\ 
\alpha
\end{bmatrix}=\begin{bmatrix}
0 \\
\vec{\textbf{1}}\end{bmatrix}
\end{align*}

By solving the above equations, the solutions are as follows: $\lambda=\frac{C-1}{-(C+3)}$, $\alpha_1=(1+\lambda)C$, $\alpha_2=\alpha_3=\frac{C(1-\lambda)}{C+1}=\frac{2C}{C+3}$, and the separating hyperplane is defined as follows:
\begin{align*}
\begin{bmatrix}
\frac{C(1-b)}{C+1} & \frac{C(1-b)}{C+1}
\end{bmatrix} \textbf{x}+\lambda= \begin{bmatrix}
2C & 2C\end{bmatrix}\textbf{x}+(1-C)
\end{align*}
and changing the value of $C$ changes the obtained hyperplane as in Fig. \ref{fig:CSimpleExample}. For example, with $C=1$, the obtained hyperplane is $\begin{bmatrix}
2C & 2C\end{bmatrix}\textbf{x}+(1-C)\Rightarrow \begin{bmatrix}
2 & 2\end{bmatrix}\begin{bmatrix}
x_1 & x_2\end{bmatrix}$. However, it can be seen from the figure that as $C\rightarrow \infty$ approaches the optimal hyperplane.

However, as reported in \cite{lin2001formulations}, with $C\geq C^*$, the solution approaches the solution of the hard margin problem in Equation (\ref{EQN:SVM}). The value of $C^*$ can be calculated easily by setting $C=\infty$ and then $C^*=max_i \alpha_i$. This is clear in our example when we set $C=100$ (i.e. C is very large) and we found that $max_i \alpha_i=C^*\approx 5.78$. As a consequence of that, with $C\geq 5.78$, the obtained results will be the same. This is clear in Fig. \ref{LDB:5p78}, \ref{LDB:10}, and \ref{LDB:100}, where the value of $C\geq 5.78$. However, the same training and testing results obtained with $C=1$. To conclude, with large values of $C$, the model converges to the optimal hyperplane, but large $C$ increases the weight of the non-separable samples and hence one outlier or critical sample can determine the decision boundary, which makes the classifier more sensitive to the noise in the data. This leads to severe overfitting, i.e. increases the complexity of SVM and makes the decision boundary sharp \cite{keerthi2003asymptotic,wang2005support,ben2008support}.
\end{itemize}

According to Equation (\ref{EQN:SVMERROR}), decreasing or neglecting $C$ minimizes the norm of $\textbf{w}$ and this maximizes the margin \cite{keerthi2003asymptotic,lin2001formulations}. In our example, the margin was $\approx 21.7$  when $C=0.01$ and the margin significantly decreased to $\approx 0.59$ with $C=5.78$. The large margin may have all training samples as in Fig. \ref{LDB:p01} and this increases the number of misclassified training and testing samples. By contrast, with a large $C$, the margin is small as in Fig. \ref{LDB:5p78} and hence there are no training samples between the two planes. 

\begin{figure}[!t]
\centering
\subfloat[\label{fig:1}]{\includegraphics[width=0.35\textwidth]{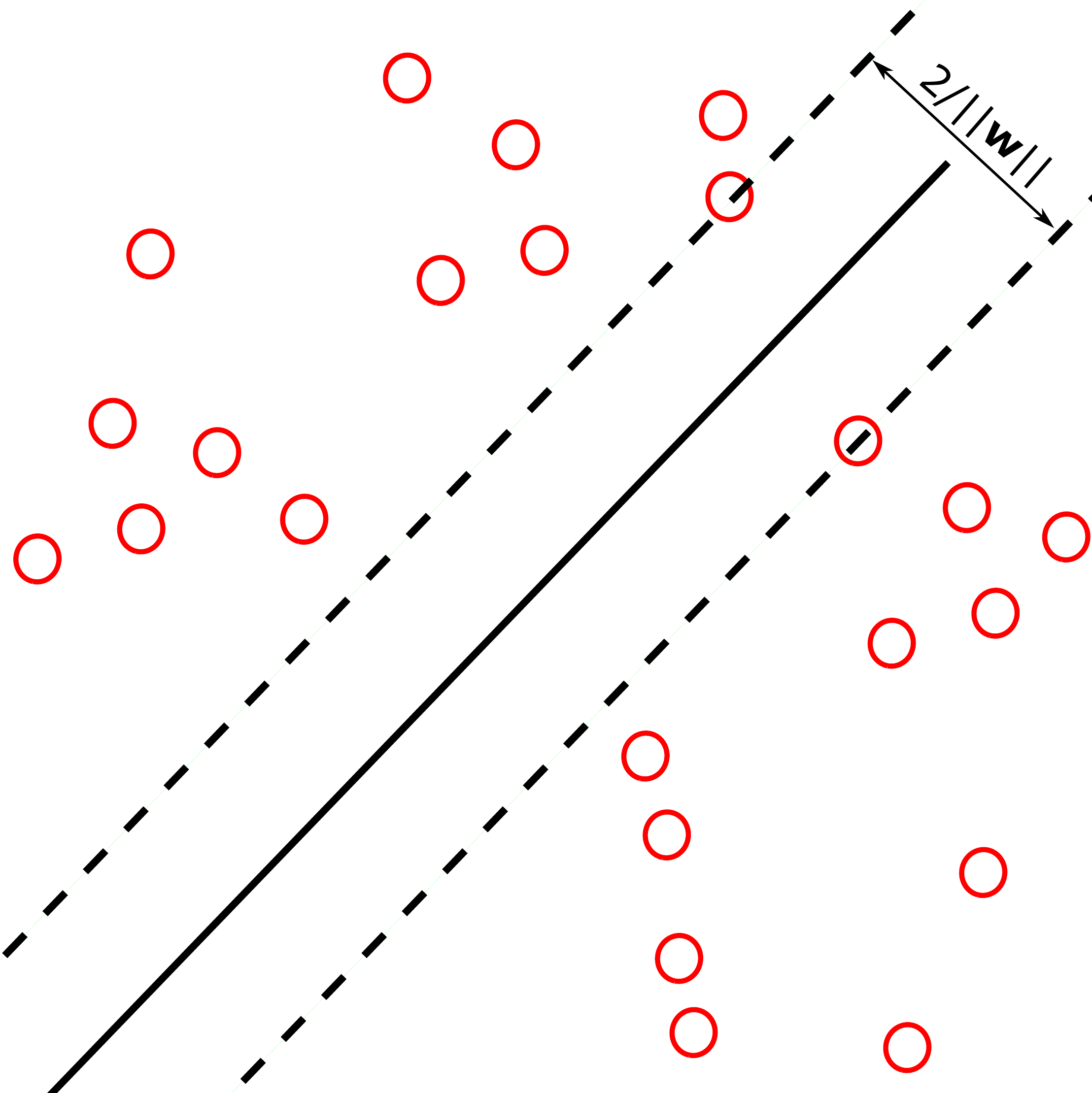}}\hfill
\subfloat[\label{fig:1}]{\includegraphics[width=0.35\textwidth]{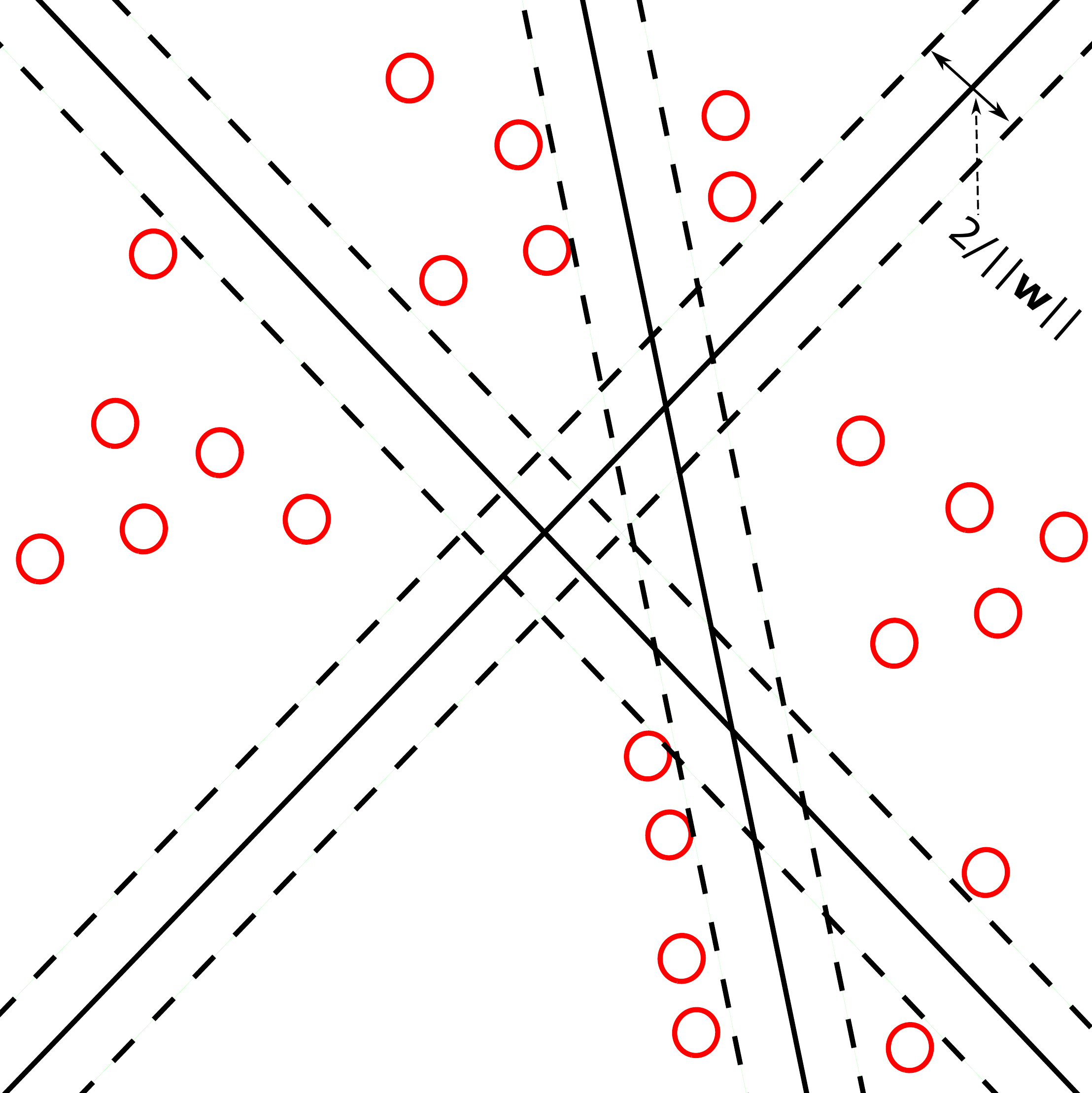}}\hfill
\caption{An illustrative example to show the influence of the margin width on the VC dimension. (a) Hyperplanes with a large margin create only a small number of possibilities to separate the data (e.g. with this large margin there is only one hyperplane that separates the data). (b) Hyperplanes with a small margin increase the chance of creating more separating hyperplanes.}
\label{fig:MarginvsVCdimension}
\end{figure}

Figure \ref{fig:MarginvsVCdimension} shows also that increasing the SVM margin decreases the VC dimension. As shown, with a high margin, there is a small number of possibilities to separate the two classes and this reduces the VC dimension. As shown in Fig. \ref{fig:MarginvsVCdimension}(a), with a large margin, there is only one possible hyperplane. On the other hand, with a small margin as shown in Fig. \ref{fig:MarginvsVCdimension}(b), the number of possible separating hyperplanes is high and this increases the VC dimension. As a consequence, a small $C$ increases the margin and this reduces the VC dimension and may lead to the underfitting problem. While a large $C$ reduces the margin; as a consequence, increases the VC dimension and this may lead the overfitting problem.

In terms of support vectors, increasing $C$ increases the number of support vectors. As in Table \ref{Table:LinearSVM}, the number of support vectors was small with a small $C$ and increasing $C$ increases the number of support vectors. The number of support vector reflects the complexity of the classification model \cite{keerthi2003asymptotic,wang2005support}. Vapnik in \cite{vapnik2013nature} introduced an alternative bound on the risk of SVM as follows:
\begin{align}
E[P(error)]\leq \frac{E[\text{number of support vectors}]}{N}
\label{EQN:Bound2}
\end{align}
where $E[\text{number of support vectors}]$ is the expectation of the number of support vectors over all choices of the training samples with size $N$, $P(error)$ is the risk for a model that was trained using $N-1$ samples, and $E[P(error)]$ is the expectation of the risk over all choices of the training data with size $N-1$. Practically, this bound can be estimated by removing one of the training samples and then retrain the model and test the removed sample. This process can be repeated for all training samples. We can easily find that removing training samples that are not support vectors would not affect the optimal hyperplane while removing one support vector generates a new hyperplane and the worst case is that every support vector will become an error. As indicated in Equation (\ref{EQN:Bound2}), the upper bound of the risk represents the expectation over all training sets with size $N-1$. As a consequence, the error bound is independent of the dimensionality of the input space and a model with fewer support vectors gives better performance \cite{burges1998tutorial}. This agrees with our example, where with a small $C$, the model is simple and has a small number of support vectors while a large $C$ increases the number of support vectors and this is could be an indicator for the overfitting problem.

We have conducted the same experiments with overlapped classes and we obtained the same findings. In comparison with the linearly separable data, with the overlapped classes, the values of $\alpha_i$ increase and some of them reach the upper bound $C$. Hence, we cannot determine the value of $C^*$ as in the linearly separable data. However, also, there is a specific limit ($C^*$) and we can estimate it by trying different values of $C$, and with $C\geq C^*$, the same results are obtained. 

To conclude, there is neither default value for $C$ nor even a theory that can help to determine the value of it. Therefore, the optimal value of $C$ can be obtained by trying a finite number of values to find the value that achieves the minimum classification error.  

\subsection{RBF kernel function}
The Gaussian or RBF kernel has only one parameter ($\sigma$) as indicated in Equation (\ref{EQN:RBFKernel}), and some references use $\gamma=\frac{1}{2\sigma^2}$ instead of $\sigma$ \cite{keerthi2003asymptotic}. The RBF kernel has many variants such as the exponential kernel which is defined as follows, $K(\textbf{x}_i,\textbf{x}_j)=\text{exp}(-\frac{\left \| \textbf{x}_i-\textbf{x}_j \right \|}{2\sigma^2})$. Hence, the RBF kernel sometimes called the \emph{squared exponential} kernel. The Laplacian kernel is also another variant of the RBF kernel and it is defined as follows, $K(\textbf{x}_i,\textbf{x}_j)=\text{exp}(-\frac{\left \| \textbf{x}_i-\textbf{x}_j \right \|}{2\sigma})$. 

\begin{equation}
K(x_i,x_j)=\text{exp}(-\frac{\left \| \textbf{x}_i-\textbf{x}_j \right \|^2}{2\sigma^2})
\label{EQN:RBFKernel}
\end{equation}

\begin{figure}[!t]
\centering
{\includegraphics[width=0.5\textwidth]{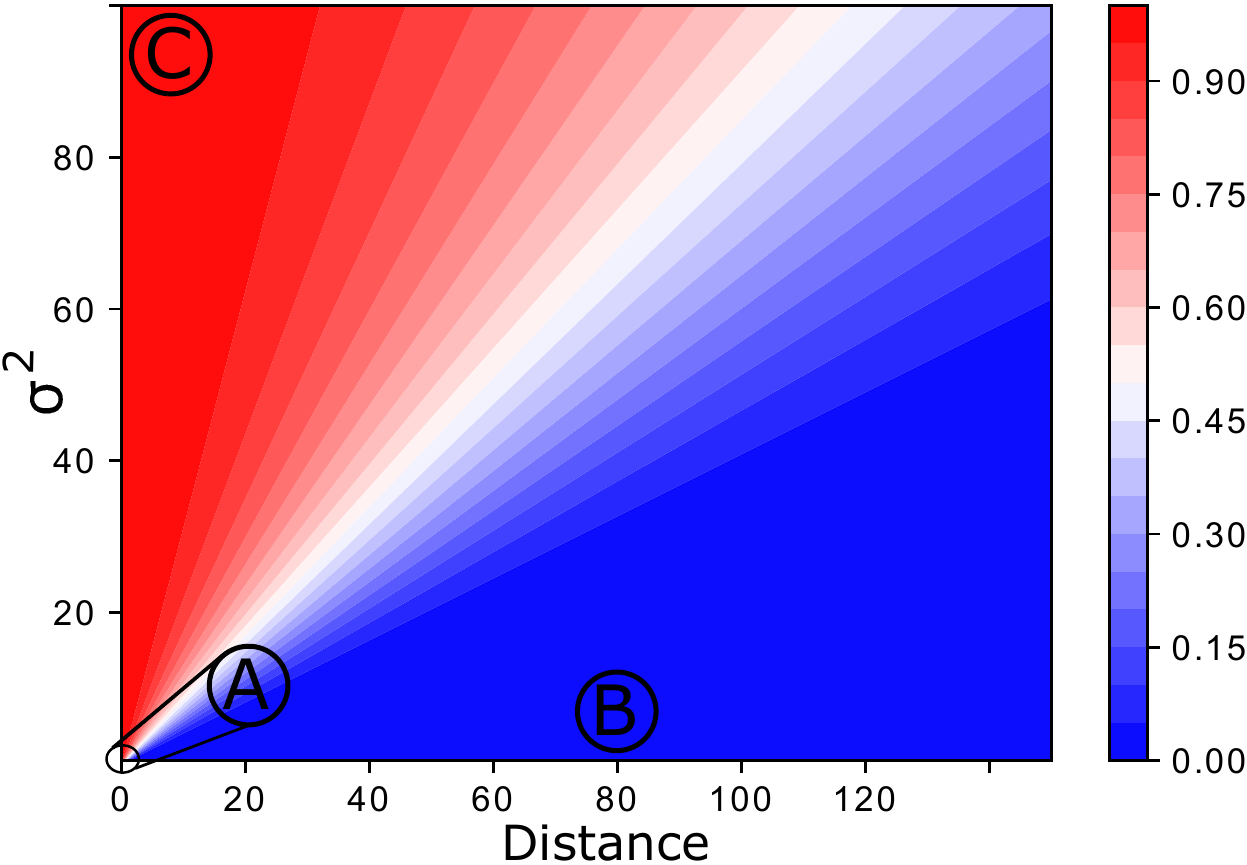}}\hfill
\caption{Visualization of the kernel value with different values of $\sigma^2$ and the distance ($\left \| \textbf{x}_i-\textbf{x}_j \right \|$). }
\label{fig:RBFKernel}
\end{figure}

In Equation (\ref{EQN:RBFKernel}), the kernel value is highly affected by the ratio of the distance between samples ($\left \| \textbf{x}_i-\textbf{x}_j \right \|$) and the value of $\sigma$ and this is clear in Fig. \ref{fig:RBFKernel}. As shown, the kernel function for any two samples tends to zero (or a tiny value approximate to zero) if $\frac{\left \| \textbf{x}_i-\textbf{x}_j \right \|^2}{2\sigma^2}\rightarrow \infty $ (see Fig. \ref{fig:RBFKernel}(B)). For example, let $\sigma=0.01$ and $\left \| x_i-x_j \right \|=10$. The kernel values is calculated as follows, $K(\textbf{x}_{i},\textbf{x}_j)=\text{exp}(-\frac{10^2}{2\times 0.01^2}) =\text{exp}(-500000)\approx 0$. As a consequence of that, a very small value of $\sigma$ transforms the pairwise distance between samples to be approximately zero in the new feature space. If the distance between samples is less than $\sigma$ (i.e. $\left \| \textbf{x}_i-\textbf{x}_j \right \|\leq \sigma$), the value of $K(\textbf{x}_i,\textbf{x}_j)$ will be high. This is clear in Fig. \ref{fig:RBFKernel}(A and C), when $\left \| \textbf{x}_i-\textbf{x}_j \right \|\leq \sigma$, the kernel value will be at least $\text{exp}(-\frac{1}{2})=0.61$ (see Fig. \ref{fig:RBFKernel}(A)). Hence, for any two identical samples (i.e. $\left\| \textbf{x}_i-\textbf{x}_j \right \|=0$), the kernel function is one, and similarly the kernel value for all samples which are within the range of $\sigma$ will be equal to one which is the maximum kernel value (see Fig. \ref{fig:RBFKernel}(C)).

To conclude, $\sigma$ affects only the distances within its range and hence $\sigma$ is called the \emph{kernel width}. 

\subsubsection{Illustrative example}
\label{SubSec:IllustrativeExampleRBF}
\begin{figure*}[!ht]
\centering
\subfloat[$\sigma=0.1$ and $C=0.1$ \label{RBF:p1}]{\includegraphics[width=0.250\textwidth]{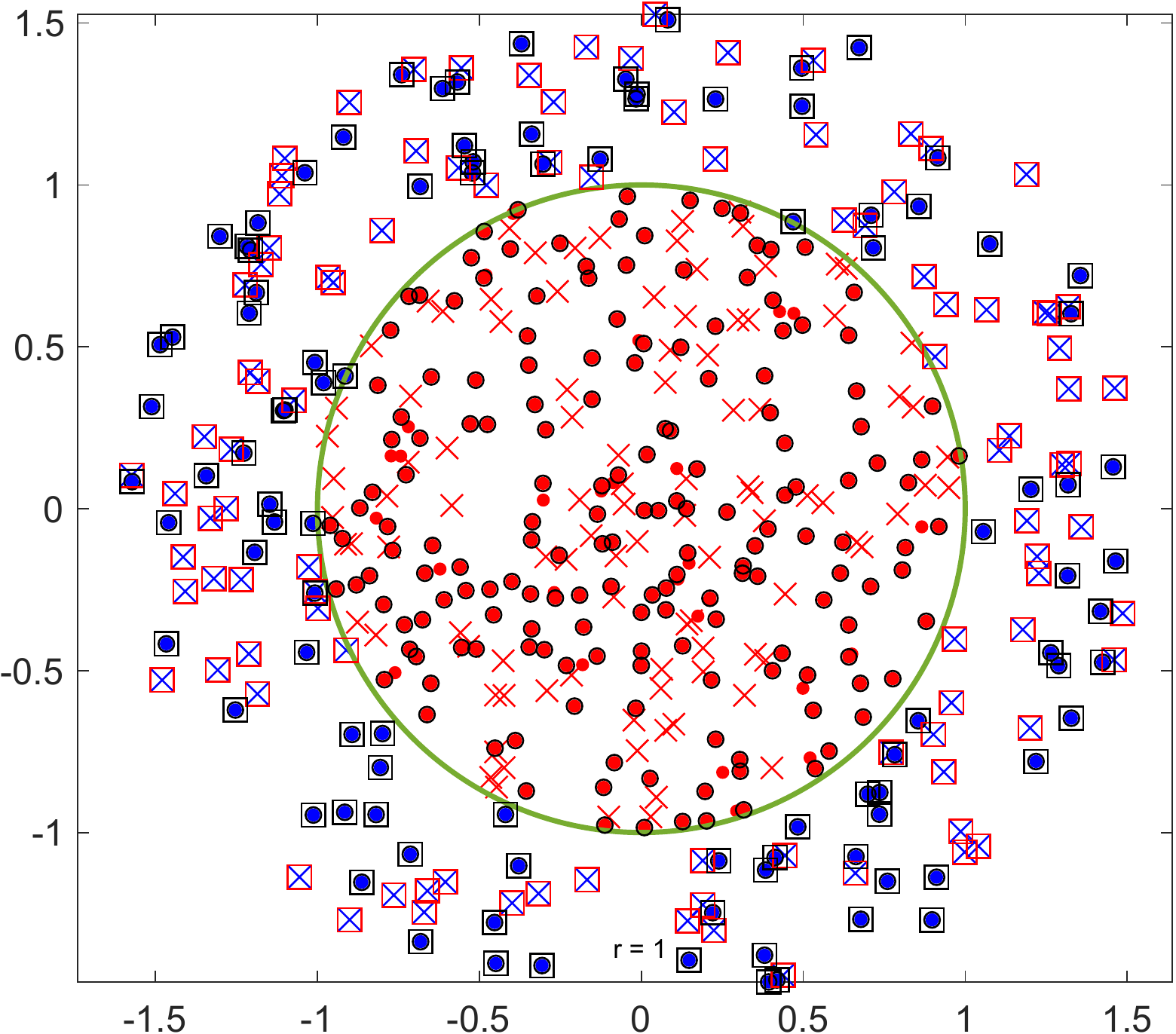}}\hfill
\subfloat[$\sigma=0.1$ and $C=0.5$ \label{RBF:p5}]{\includegraphics[width=0.250\textwidth]{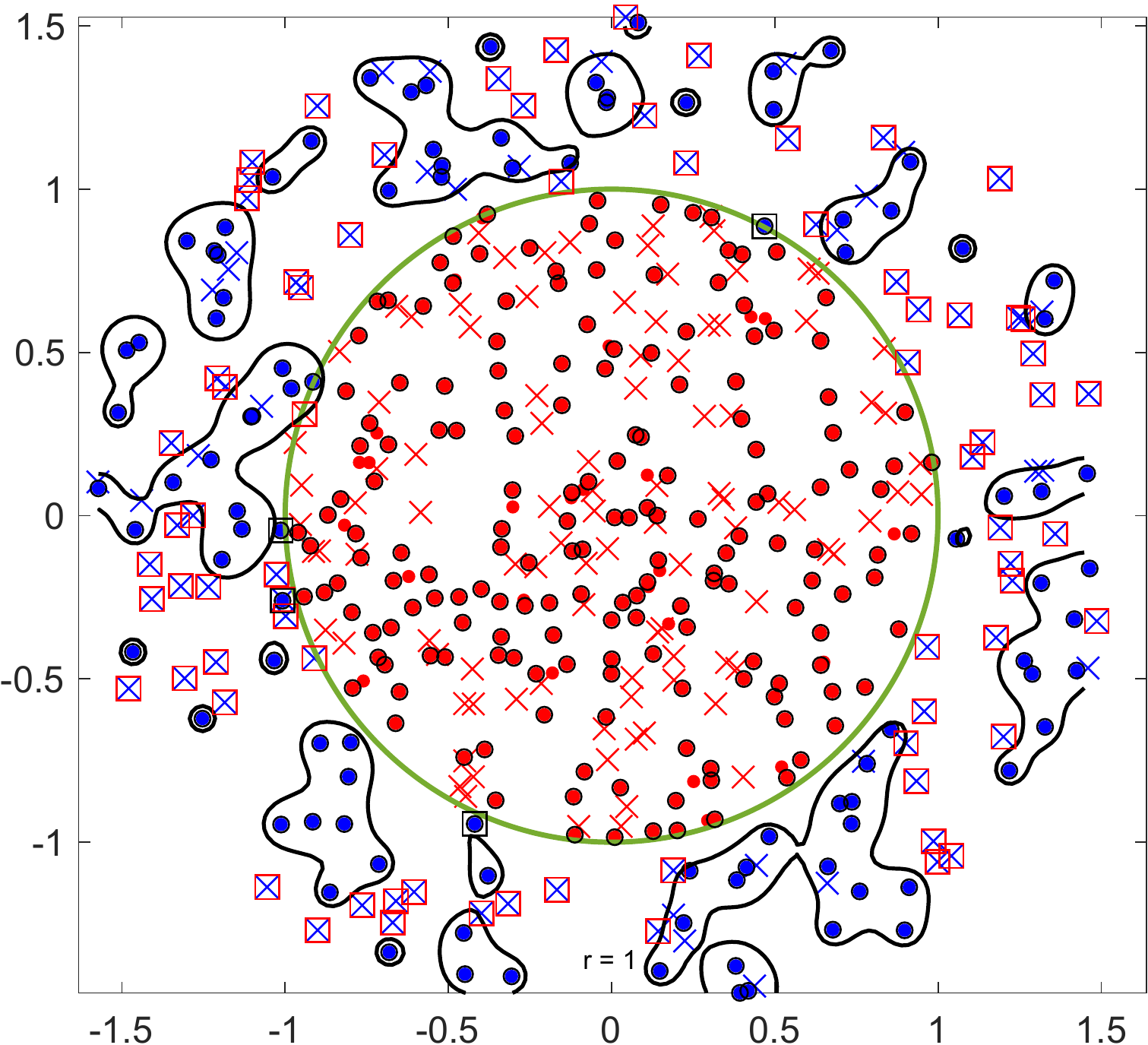}}\hfill
\subfloat[$\sigma=0.1$ and $C=\frac{C_{lim}}{2}=\frac{2}{3}$ \label{RBF:lim}]{\includegraphics[width=0.250\textwidth]{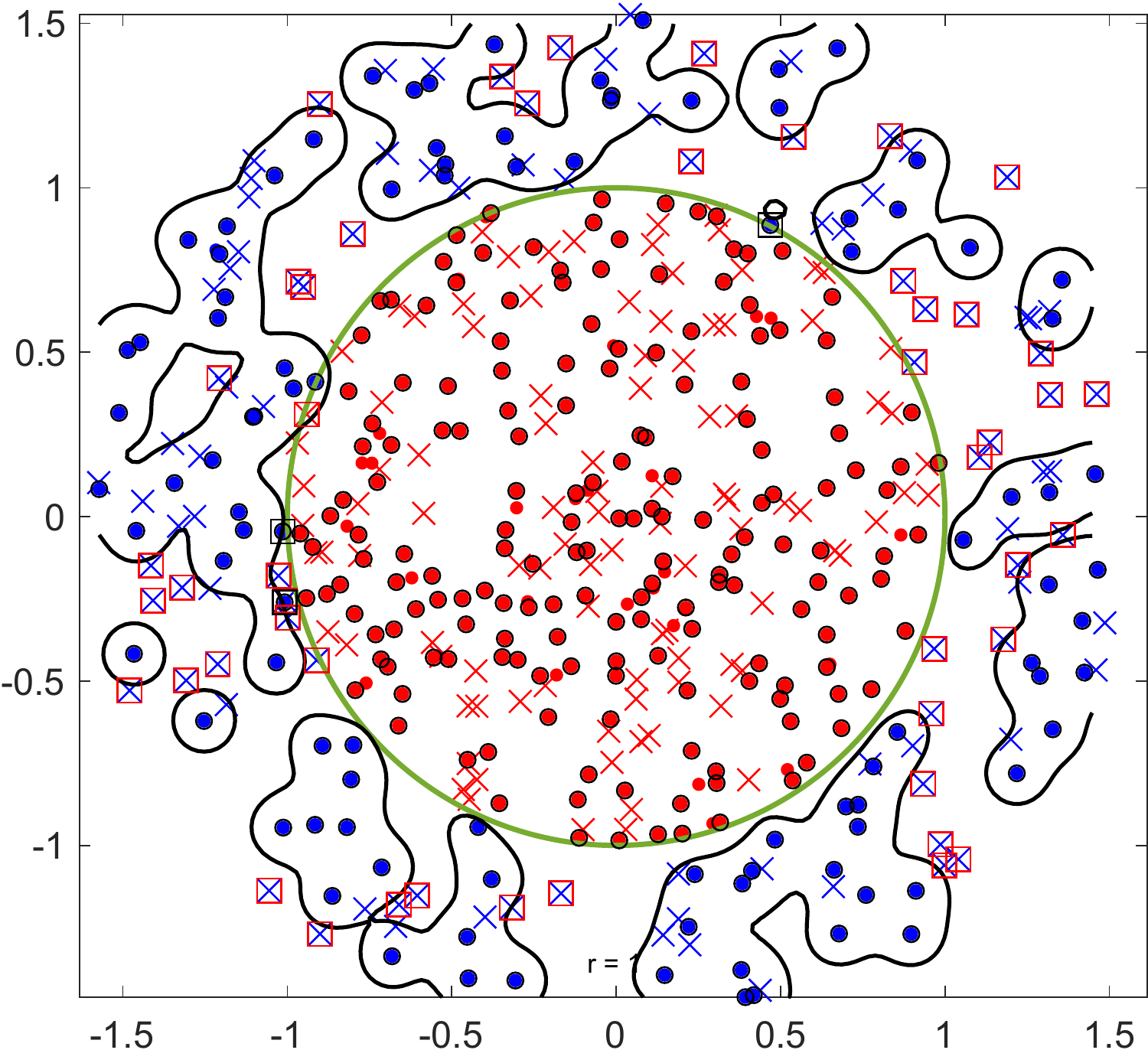}}\hfill
\subfloat[$\sigma=0.1$ and $C=1$ \label{RBF:1}]{\includegraphics[width=0.250\textwidth]{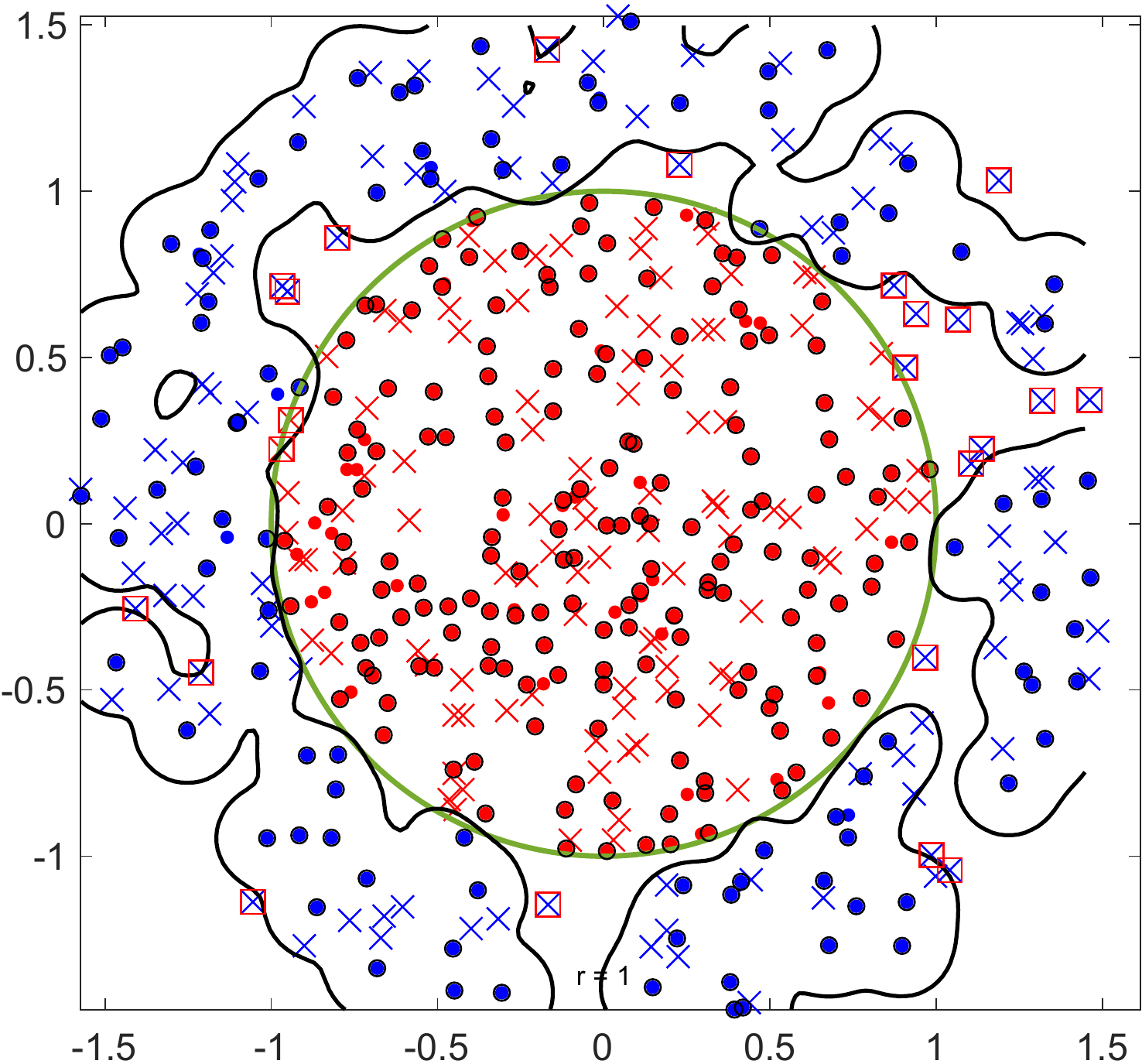}}\hfill
\caption{The classification performance of SVM with $\sigma=0.1$ (i.e. small $\sigma$) and different values of $C$. Decision boundaries (block solid line), support vectors samples are marked by surrounding it by square shapes, (\protect\tikz \protect\draw[thick, color=black, fill=blue] plot[mark=*, mark options={scale=1.1}] (0,0); and \protect\tikz \protect\draw[thick, color=black, fill=red] plot[mark=*, mark options={scale=1.1}] (0,0);) represent the minority and majority classes in the training data, respectively, and (\textcolor{blue}{x} and \textcolor{red}{x}) represent the testing samples. The green circle represent the target function or the optimal decision boundary between the two classes.}
\label{fig:RBF2}
\end{figure*}

In this example, the training data consists of 300 samples, where the first and second classes have 200 and 100 samples, respectively, (i.e. $N_1=200$ and $N_2=100$). The testing data is balanced and it consists of 200 samples, and each class has 100 samples. As shown in Fig. \ref{fig:RBF2}, the data is nonlinearly separable and hence we will use the RBF kernel. In this experiment, we have evaluated the SVM model using different combinations of $C$ and $\sigma$.

\begin{itemize}
\item With a small $\sigma^2$ (i.e. $\sigma^2 \rightarrow 0$), $\text{exp}(-\frac{\left \| \textbf{x}_i-\textbf{x}_j \right \|^2}{2\sigma^2})\rightarrow \delta_{ij} $, where $\delta_{ij}=1$ if $i=j$; otherwise, $\delta_{ij}=0$. Hence, the objective function in Equation (\ref{EQN:SVMOptDual2}) will be as follows, $\text{min} \frac{1}{2} \alpha^T \alpha -\textbf{f}^T \alpha$. Hence, as proved in \cite{lin2001formulations}, there is a certain limit of $C$ and this is denoted by $C_{lim}=\frac{2N_1}{N}$, and the SVM model is very different around this limit. In our example, $C_{lim}=\frac{2\times 200}{300}=\frac{4}{3}$.

With a large $C$ (i.e. $C\geq \frac{C_{lim}}{2}$), there is a small region around each sample of the minority class and the rest of the whole space belongs to the majority class. Hence, large $C$ overfits the training data. This is clear in Fig. \ref{RBF:1} where $C=1$ (i.e. $C\geq \frac{C_{lim}}{2}$) and as shown, the minority samples are surrounded by a small region and the rest of the space belongs to the majority class. Therefore, all the misclassified samples belong to the minority class and the training error is zero. This means that the SVM model does not generalize well to the test data and hence the model is overfitted. While for a small $C$ (i.e. $C<\frac{C_{lim}}{2}$), the SVM model tends to be underfitted and the whole space belongs to the majority class. This is clear in Fig. \ref{RBF:p1} and \ref{RBF:p5}. In Fig. \ref{RBF:p5}, with $C=0.5$, the majority of the space belongs to the majority class and hence most of the minority samples are misclassified. Decreasing $C$ to 0.1 makes the whole space belongs to the majority class and hence all minority samples are misclassified as in Fig. \ref{RBF:p1}.

In both cases (i.e. large and small $C$), the minority samples of the testing data are misclassified. Figure \ref{fig:RBFKernel2} shows the contour plot of the testing error to test the behavior of SVM with different combinations of $C$ and $\sigma^2$. Each point in the contour plot is a combination of $C$ and $\sigma^2$, where the values of $C$ are ranged from 0.1 to $100$ and the values of $\sigma^2$ are ranged from $0.01$ to $10$. As shown in Fig. \ref{fig:RBFKernel2} (region A), with a small $\sigma^2$, the testing error is high when the value of $C$ is less than $\frac{C_{lim}}{2}$ and this means that the model is underfitted. With a small $\sigma^2$, increasing $C$ makes the model moves from the underfitting region to the overfitting region (see Fig. \ref{fig:RBFKernel2}(B)).

\begin{figure}[!t]
\centering
{\includegraphics[width=0.5\textwidth]{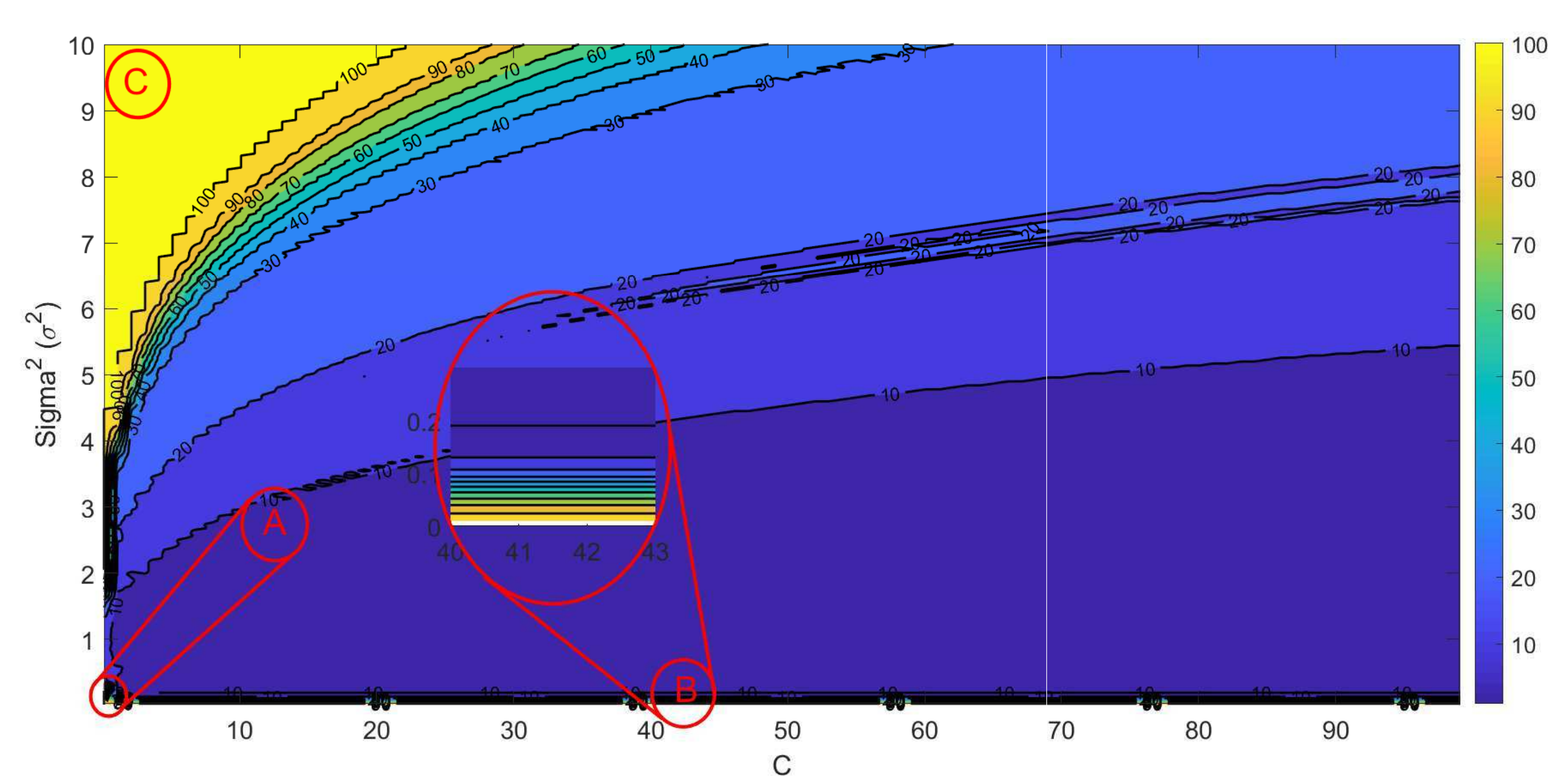}}\hfill
\caption{A contour plot to show the behavior the testing error of SVM with different combinations of $C$ and $\sigma^2$ (we used the same nonlinearly separable data in the example in Section \ref{SubSec:IllustrativeExampleRBF}).}
\label{fig:RBFKernel2}
\end{figure}

\item With a large $\sigma$ (i.e. $\sigma^2 \rightarrow \infty$), the kernel function can be written as follows:
\begin{align*}
K(\textbf{x}_i,\textbf{x}_j)&=\text{exp}(-\frac{\left \| \textbf{x}_i-\textbf{x}_j \right \|^2}{2\sigma^2}) \nonumber \\
&=1-\frac{\left \| \textbf{x}_i-\textbf{x}_j \right \|^2}{2\sigma^2}+O(\frac{\left \| \textbf{x}_i-\textbf{x}_j \right \|^2}{\sigma^2}) \nonumber \\
&=1-\frac{\left \| \textbf{x}_i\right \|^2}{2\sigma^2}- \frac{\left \| \textbf{x}_j \right \|^2}{2\sigma^2}+\frac{ \textbf{x}_i^T\textbf{x}_j }{\sigma^2}+O(\frac{\left \| \textbf{x}_i-\textbf{x}_j \right \|^2}{\sigma^2}) 
%\label{EQN:RBFKernelhighsigma}
\end{align*}
hence, $\textbf{H}$ will be
\begin{align*}
\textbf{H}=\sum_{i,j}\alpha_i\alpha_jy_iy_jK(\textbf{x}_i,\textbf{x}_j)&=T_1+\frac{T_2+T_3+T_4}{2\sigma^2} \nonumber \\
&+\frac{1}{2}\sum_{i,j} \alpha_i\alpha_j y_iy_j \frac{\Delta_{ij}}{\sigma^2}
\end{align*}
where
\begin{align*}
&T_1=\sum_{i,j}\alpha_i\alpha_jy_iy_j, \;\;\;  T_2=-\sum_{i,j}\alpha_i\alpha_jy_iy_j\left \| \textbf{x}_i \right \|^2 \nonumber \\
&T_3=-\sum_{i,j}\alpha_i\alpha_jy_iy_j\left \| \textbf{x}_j \right \|^2, \;\;\; T_4=2\sum_{i,j} \alpha_i\alpha_jy_iy_j\textbf{x}_i^T\textbf{x}_j \nonumber \\
&\underset{\sigma^2\rightarrow \infty}{\text{lim}} \Delta_{ij}=0
\end{align*}

By the equality constraint ($\sum_{i}\alpha_iy_i=0$), $T_1=0$. Also, $T_2$ can be written as follows, $-(\sum_{i}\alpha_iy_i\left \| \textbf{x}_i \right \|^2)(\sum_{j}\alpha_jy_j)$ and hence $T_2=0$ and similarly $T_3=0$ \cite{lin2001formulations}. By defining $\tilde{\alpha}_i=\frac{\alpha_i}{\sigma^2}$, the objective function can be written as follows:
\begin{align*}
\underset{\tilde{\alpha}}{\text{min}} \frac{F}{\sigma^2}=\frac{1}{2}\sum_{i,j}\tilde{\alpha}_i\tilde{\alpha}_jy_iy_j\tilde{K}_{ij}-\sum_i \tilde{\alpha}_i \nonumber \\
\text{subject to: } 0\leq \tilde{\alpha}_i \leq \tilde{C}, i=1,\dots,N, \;\; \textbf{Y}^T\tilde{\alpha}=0
\end{align*}
where $\tilde{K}_{ij}=\textbf{x}_i^T\textbf{x}_j+\Delta_{ij}$ and $\tilde{C}=\frac{C}{\sigma^2}$. Hence, with $\sigma^2\rightarrow \infty$, $\tilde{K}_{ij}\rightarrow \textbf{x}_i^T\textbf{x}_j$. As a consequence of that, for a fixed $C$ and $\sigma^2\rightarrow \infty$, SVM with RBF kernel behaves like linear SVM with $\tilde{C}$ \cite{lin2001formulations}. This means that two combinations $\frac{C_1}{\sigma^2_1}=\frac{C_2}{\sigma^2_2}=\tilde{C}$ have the same generalization error. Hence, the contour lines of the hyperparameter space in this case will be straight lines with slope 1: $\text{log} \sigma^2=\text{log} C-\text{log} \tilde{C}$ \cite{keerthi2003asymptotic}. Thus, for large $\sigma^2$, all SVM models which are defined by points on that straight line are nearly the same as the linear SVM with $\tilde{C}$. Figure \ref{fig:RBFKernel2} shows that with a large $\sigma^2$, the contour lines of the hyperparameter space are represented by straight lines. Figure \ref{fig:Ctilde}(a and b) illustrates that the SVM performance of two different combinations are nearly the same. These two points lie on the straight line where $\tilde{C}=\frac{C_1}{\sigma^2_1}=\frac{C_2}{\sigma^2_2}$. In Fig. \ref{RBF:Ctild1}, with $(C_1=20, \sigma^2_1=8)$, the training and testing errors were 7 and 6, respectively, while with $(C_2=30, \sigma^2_2=12)$, the training and testing errors were 6 and 6. Hence, the SVM classifiers in both cases are nearly the same and this small difference may be due to the fact that our training and testing data are generated randomly in each case. In Fig. \ref{RBF:CTild3}, the point $(C_3=0.025, \sigma^2_3=0.01)$ lies on the same line; but, it has different performance than the other two points. This is because, this relation is only applicable when $\sigma^2 \rightarrow \infty$.

In Fig. \ref{fig:RBFKernel2}(C), the model with a large $\sigma^2$ and small $C$ obtained high training and testing errors (i.e. the model is underfitted), and the testing error is $N_2$. This is clear in Fig. \ref{fig:LargeSigma}, where $C=1$ and $\sigma^2=100$. As shown, the whole space belongs to the majority class and the training and testing minority samples were misclassified and hence the model is underfitted. This is can be explained simply as follows: with a large $\sigma^2$ and small $C$, $\tilde{C}=\frac{1}{100}=0.01$. Hence, the model behaves like linear SVM with $\tilde{C}=0.01$ and as we mentioned before, linear SVM with a small $C$ increases the margin and all samples will be in the half space of the majority class. Hence, the minority samples will be misclassified (see Fig. \ref{LDB:p01}).
\end{itemize}

\begin{figure}[!t]
\centering
{\includegraphics[width=0.45\textwidth]{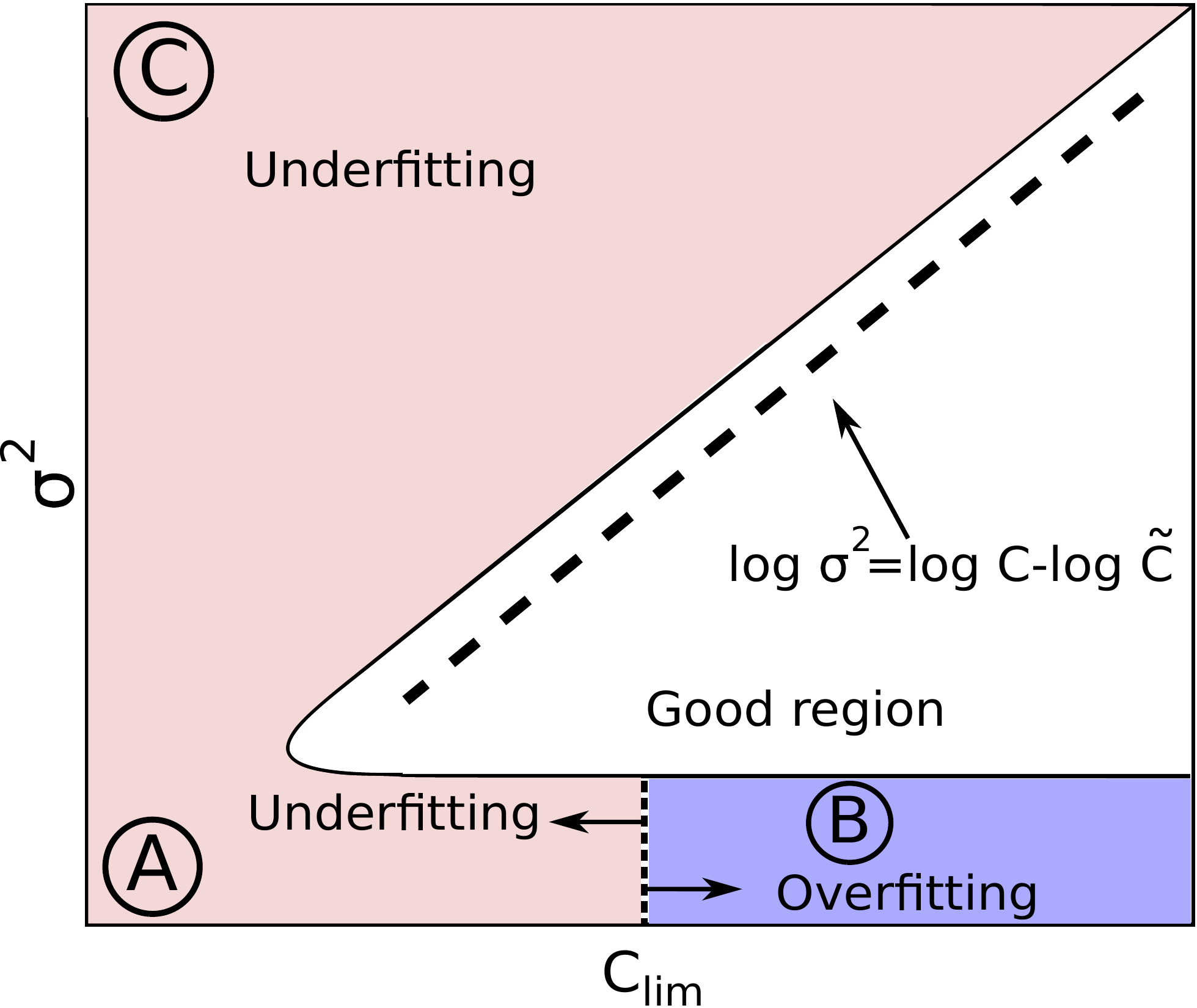}}\hfill
\caption{A boundary curve separating the good region and the underfitting/overfitting region in SVM with the RBF kernel. In (A and C) the model suffers from the underfitting, while in (B) the model is overfitted. Inside the good region, for each fixed $\tilde{C}$, there is a straight line with a unit slope is defined as follows, $\text{log} \sigma^2=\text{log} C-\text{log} \tilde{C}$. With $\sigma^2 \rightarrow \infty$, the model behaves as a linear SVM with $\tilde{C}$. The dotted line corresponds to $\tilde{C}$ that gives the optimal generalization error for the linear SVM.}
\label{fig:RBFKernel3}
\end{figure}

From the above findings, we can conclude that the parameter space of $C$ and $\sigma^2$ has different regions: good region and underfitting/overfitting region. This good region has the values of $C$ and $\sigma^2$ that obtain the best generalization error. As shown in Fig. \ref{fig:RBFKernel3}, inside the good region, with a large $\sigma^2$, there is a straight line(s) with a unit slope. Therefore, instead of searching in the whole space of $C$ and $\sigma^2$, we can search inside the good region.

\begin{figure*}[!ht]
\centering
\subfloat[$\sigma^2=8$ and $C=20$ \label{RBF:Ctild1}]{\includegraphics[width=0.330\textwidth]{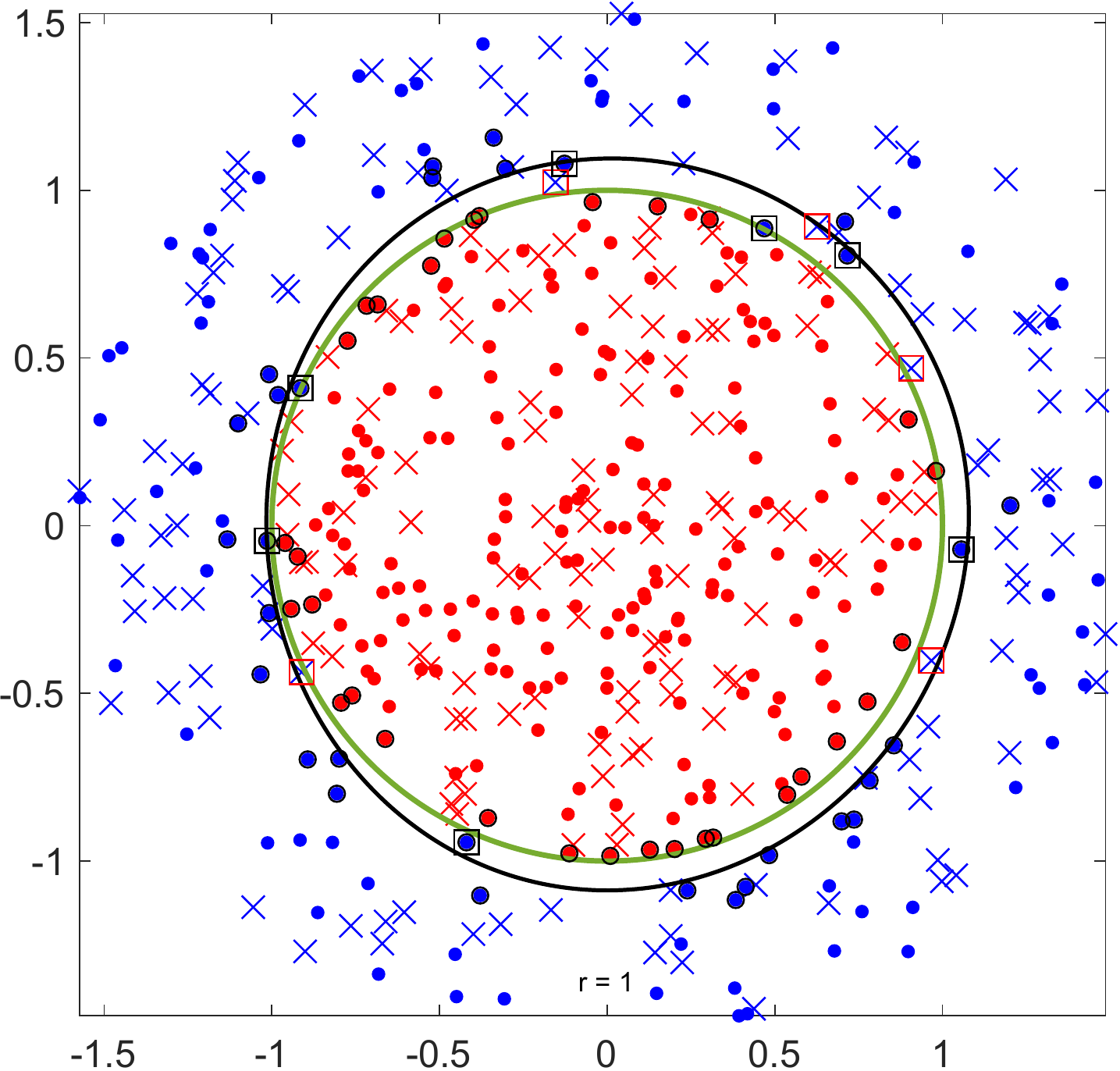}}\hfill
\subfloat[$\sigma^2=12$ and $C=30$ \label{RBF:CTild2}]{\includegraphics[width=0.330\textwidth]{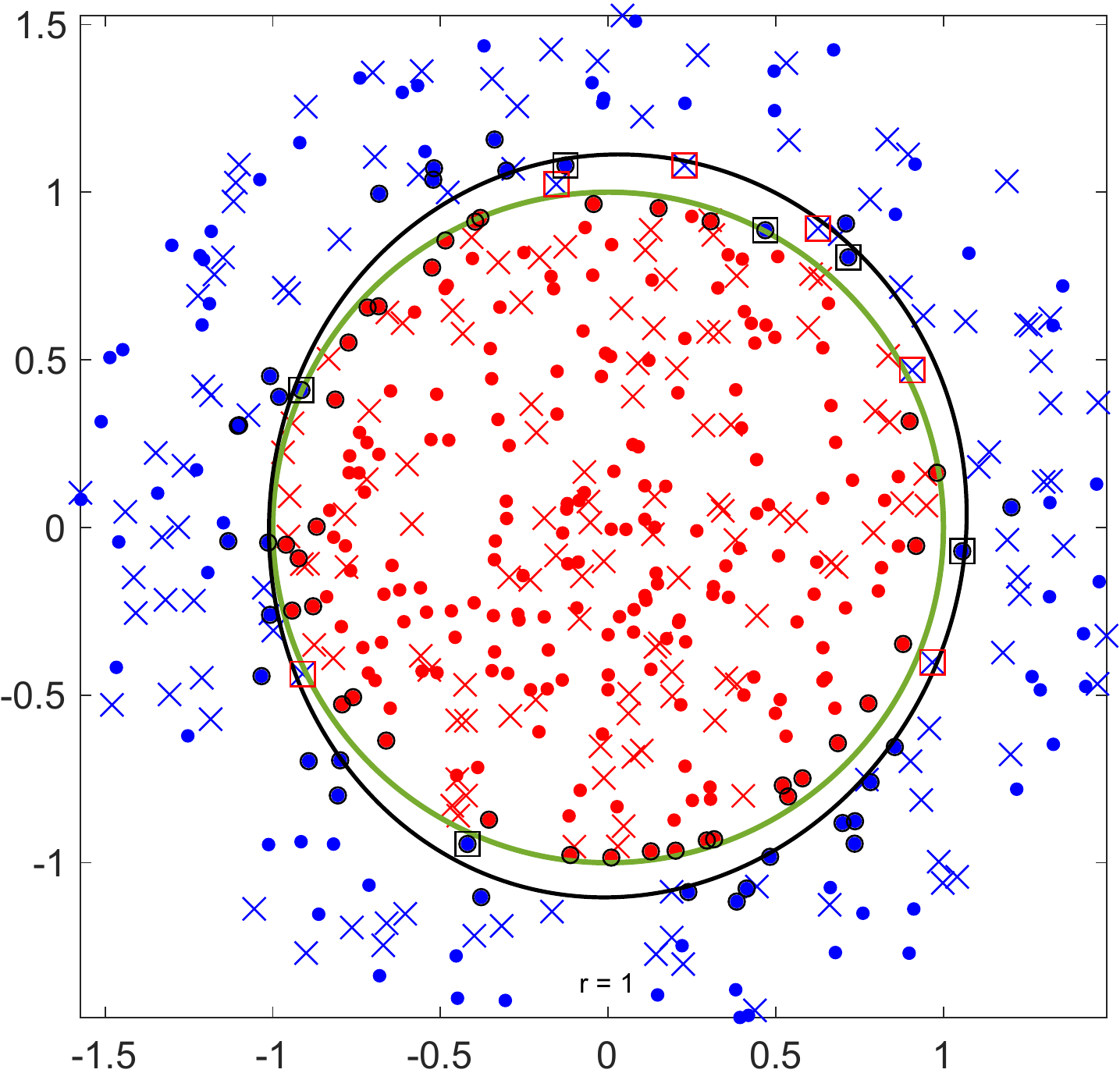}}\hfill
\subfloat[$\sigma^2=0.01$ and $C=0.025$ \label{RBF:CTild3}]{\includegraphics[width=0.330\textwidth]{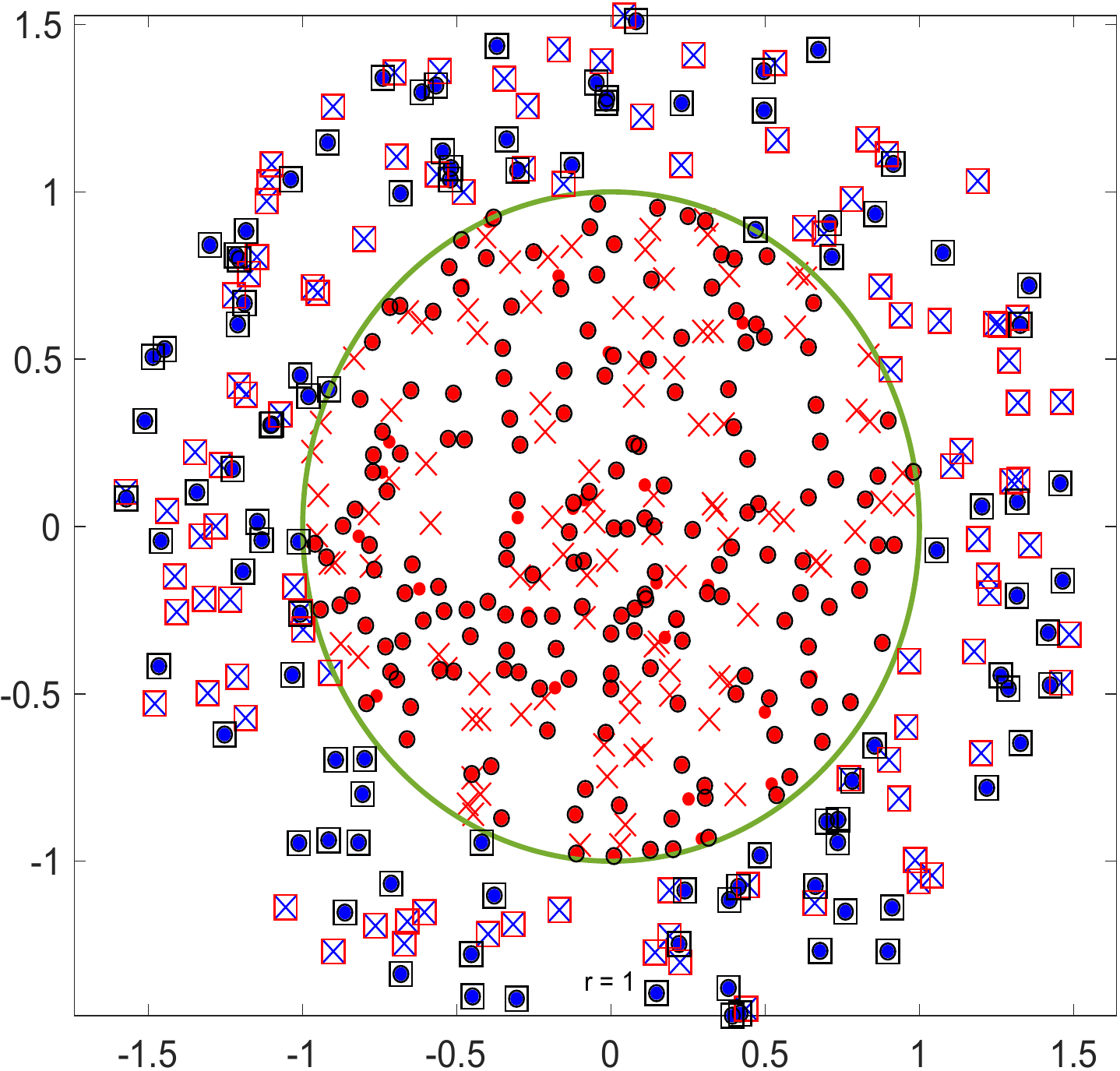}}\hfill
\caption{Visualization of the performance of SVM classifier using large $\sigma^2$. (a, b, and c) are different points/combinations along a line $\text{log} \sigma^2=\text{log} C-\text{log} \tilde{C}$ which has a unit slope. The two combinations $(C_1=20, \sigma^2_1=8)$ and $(C_2=30, \sigma^2_2=12)$ in (a) and (b), respectively, have approximately the same classification performance. (c) This point ($C_3=0.025, \sigma^2_3=0.01$) along the same line but it has difference classification performance.}
\label{fig:Ctilde}
\end{figure*}

\begin{figure}[!t]
\centering
{\includegraphics[width=0.49\textwidth]{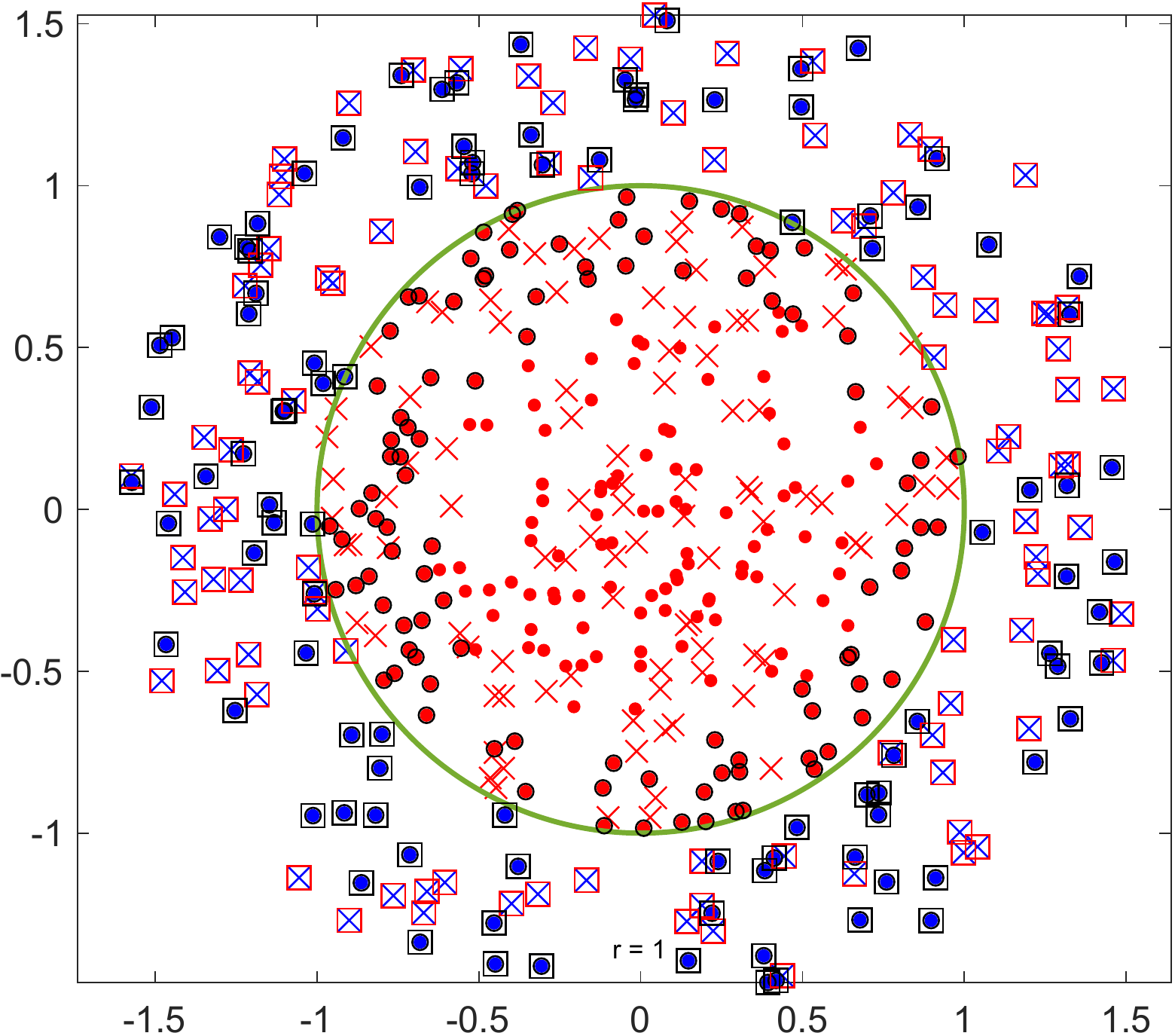}}\hfill
\caption{The performance of SVM with a large $\sigma^2=100$ ($C=1$).}
\label{fig:LargeSigma}
\end{figure}

\subsection{The proposed algorithm for model selection}
In many studies, it is usual to form a two-dimensional uniform grid, for example with $r\times r$ dimensions, and then search inside this space for the combination that gives the best generalization error. However, trying $r^2$ combinations is expensive and in many cases is not practically possible. Increasing the dimension of the search space increases the required computational time. Instead, based on our analysis, the proposed algorithm reduces the computational time significantly. More details are as follows:
\begin{enumerate}
\item Let we use the linear SVM and then search for the best $C$ that achieves the best generalization errors and call it $\tilde{C}$. This step sometimes gives a reasonable classification performance. However, practically, adding some nonlinearities by using a kernel function help to improve the classification performance.
\item According to our analysis and as reported in \cite{AlaaParameter}, with a large $\sigma^2$, the samples are mapped to be close to each other and the new hypersurface is almost flat, and the model tends to be underfitted. While a small $\sigma^2$ makes the model focus on a small set and a very small $\sigma^2$ transforms samples into different subspaces and hence the transformed samples are far from each other and the new hyperspace will be spiky, and the model tends to be overfitted. Therefore, instead of searching in the whole space of $\sigma$, with some data analysis, we can determine the range of $\sigma$. This analysis including calculating the maximum and minimum distances between the samples within the same class. For example, if the maximum distance between samples within the same class is one, setting $\sigma^2$ to 10 means that all samples are within the range of $\sigma^2$ and hence the model tends to be underfitted. Finding the range of $\sigma$ reduces the search space significantly and hence reduces the required computational time.
\item From the first step, fix $\tilde{C}$ and search for the best $(C, \sigma^2)$ along the line $\text{log} \sigma^2=\text{log} C-\text{log} \tilde{C}$. Therefore, instead of searching in a two-dimensional space, our proposed algorithm searches only in two one-dimensional spaces. Hence, our algorithm requires only $2r$ combinations of $(C,\sigma^2)$ to be tried instead of $r^2$ with the grid search. This decreases the required computational time dramatically. The algorithm in \cite{keerthi2003asymptotic} assumed that there is one line in the good region. Practically, we found that trying different values for $\tilde{C}$ obtained good generalization errors. Thus, within the good region, we can search for the best $(C, \sigma^2)$ along different lines with different values of $\tilde{C}$ and this increases the chance of finding the optimal solution. This extends the search space slightly; therefore, it requires additional computational time; but, it is still very small compared with searching in the whole search space. 
\end{enumerate}

In the next section, different experiments are conducted to compare our proposed algorithm with the grid search algorithm.

\section{Experimental Results and Discussion}
\label{Sec:Exp}
In this section, different experiments were carried out to evaluate our proposed search algorithm. We used ten standard classification datasets and these datasets were obtained from the University of California at Irvin (UCI) Machine Learning Repository \cite{blake1998uci} and KEEL\footnote{Available at http://sci2s.ugr.es/keel/imbalanced.php.}. The descriptions of all datasets are summarized in Table \ref{Table:Datasets}. All datasets are binary\footnote{Iris dataset has three classes; but, we used only two classes.}. The chosen datasets have a reasonably wide range: number of features (4 to 60), number of samples (100 to 768), and imbalance ratio (1 to 15.5) and so the empirical evaluation demonstrates the applicability of the proposed algorithm with different types of datasets. The first six datasets in the table are approximately balanced (IR$\approx 1$) while the last four datasets are imbalanced, Due to the presence of imbalanced data, we used the accuracy, sensitivity, and specificity metrics \cite{tharwat2018classification}.

In all experiments, the first step of our algorithm uses the linear SVM and search for $\tilde{C}$ that obtains the best testing error. Second, we analyze the data to find the feasible range of $\sigma$ parameter. Finally, we search for the optimal $(C, \sigma^2)$ that obtains the best testing error. In all experiments, $k$-fold cross-validation tests have been used and we used five-fold cross-validation to estimate the results. For further evaluation for the proposed algorithm, different comparisons with the grid search algorithm and also the results of some relevant studies were conducted. In the grid search algorithm, the search space for $\text{log}C$ was $[-7-7]$ and for $\text{log}\sigma^2$ was $[-8,8]$, and the space is uniformly spaced.

\begin{table}[!ht]
\centering
\caption{Datasets description.}
\resizebox{0.5\textwidth}{!}{%
\begin{tabular}{cccc}
\hline
{ Dataset} & { Dim.} & { \# Samples} & ($N_1, N_2$), IR \\ \hline 
Iris           & 4               & 100           & (50, 50), 1             \\ 
Sonar         & 60              & 208           & (111, 97), 1.1            \\ 
Liver-disorders         & 6               & 345           & (200, 145), 1.4             \\ 
Diabetes           & 8              & 768           & (500, 268), 1.9             \\ 
Breast cancer           & 9              & 683           & (444, 239), 1.9             \\ 
Iono          & 33              & 225           & (225,126), 1.8             \\ \hline \hline
Ecoli1           & 7              & 336           & (259, 77), 3.36             \\ 
Ecoli2           & 7              & 336           & (284, 52), 5.5             \\ 
Glass2           & 9              & 214           & (197, 17), 11.6             \\ 
Glass4           & 9              & 214           & (201, 13), 15.5             \\ \hline
\end{tabular}}
\label{Table:Datasets}
\end{table}

\subsection{Iris dataset}
\begin{figure}[!t]
\centering
\subfloat{\includegraphics[width=0.490\textwidth]{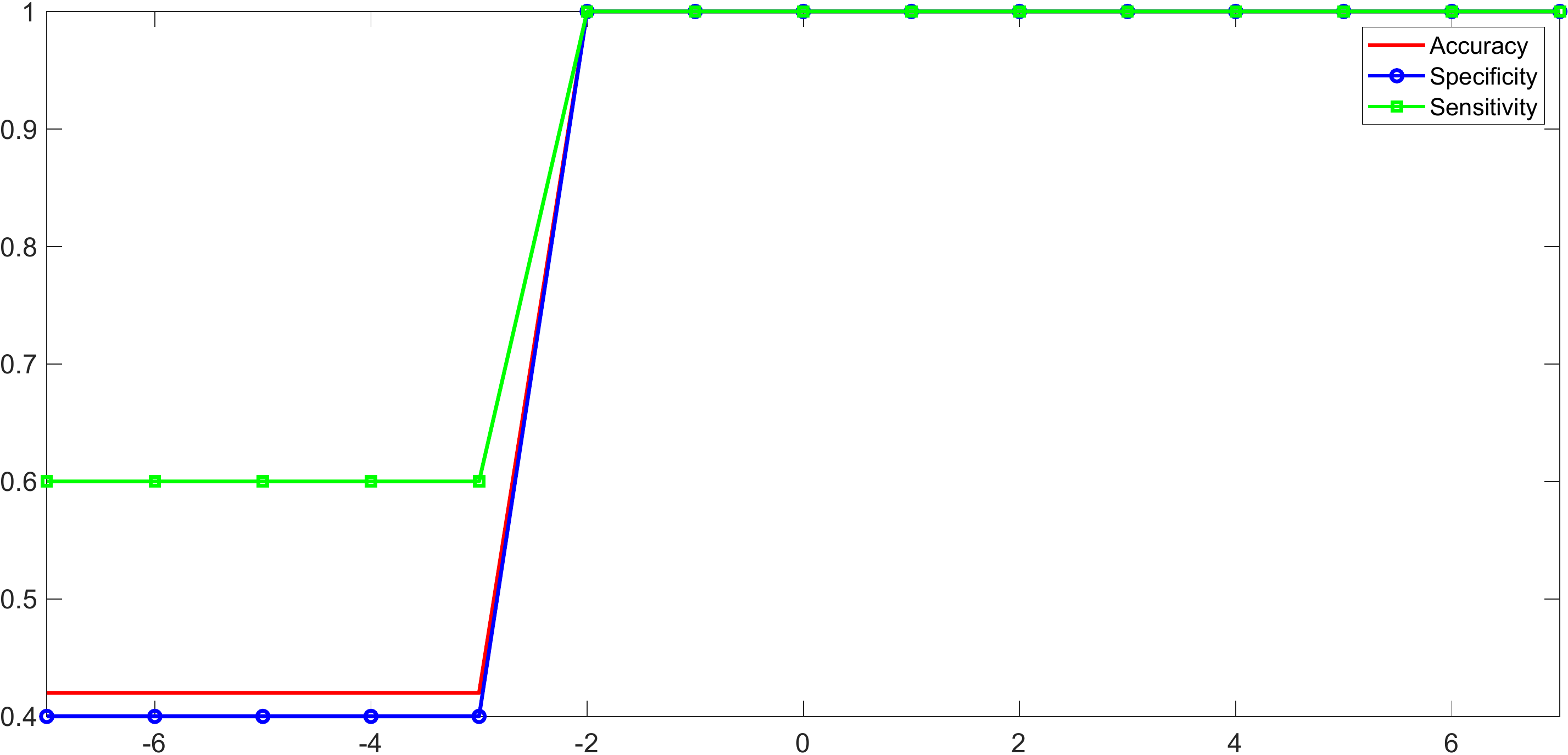}}\hfill
\subfloat{\includegraphics[width=0.490\textwidth]{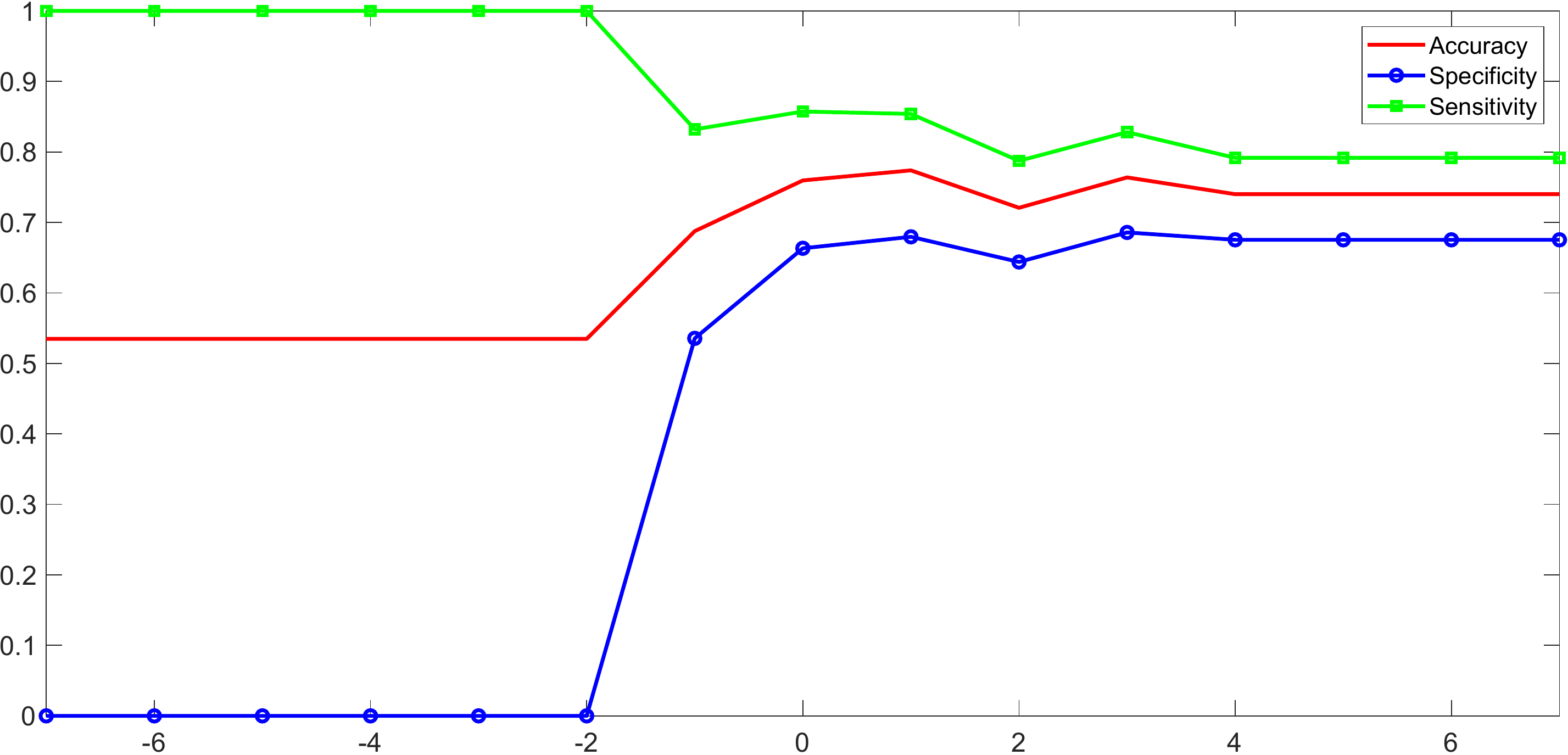}}\hfill
\caption{Variation of the accuracy, sensitivity, and specificity of the linear SVM with $C$. (a) Iris dataset. (b) Sonar dataset. The horizontal line is the values of $\text{log}C$.}
\label{fig:Iris1}
\end{figure}

The iris dataset as shown in Table \ref{Table:Datasets} has three balanced classes, each class 50 samples and each sample is represented by four features. We used only two classes (the first two classes). The first step in the searching strategy is to search for the best $C$ that gives the minimum generalization error. Figure \ref{fig:Iris1}(a) shows the accuracy, the sensitivity, and the specificity of the linear SVM with different values of $C$. As shown, the range of $\text{log}C$ was $[-7,7]$ and the best performance obtained when $C\geq 0.01$ (i.e. $\text{log}C\geq -2$). Hence, $\tilde{C}=0.01, 0.1, 1, \dots$. To reduce the search space of $\sigma^2$, in the second step, we found that the maximum distances between the samples in the first and second classes were 2.43 and 2.65, respectively, and the minimum distances were 0.1 and 0.2, respectively. Therefore, the search space of $\sigma^2$ was small and it was ranged from $0.1^2$ to $2.65^2\approx 7$. Next, we searched for the optimal $(C, \sigma^2)$ along the line $\text{log} \sigma^2=\text{log} C-\text{log} \tilde{C}$. As a consequence, a small number of points have to be tried and this is the reason why a small grid spacing was used for getting better results. In this dataset, the grid space was 0.01. This is for two reasons, (1) because the number of points that have to be tried is small (e.g. the number of points is four when $\tilde{C}=0.1$), and (2) to scan the search space carefully. However, changing the grid spacing changes also the number of points that have to be tried.

We have tried different values for $\tilde{C}$ and we found that: (1) with $\tilde{C}=0.01$, the best results obtained when $(C=0.1, \sigma^2=10)$, (2) with $\tilde{C}=0.1$, the best results obtained when $(C=0.1, \sigma^2=1)$ and $(C=1, \sigma^2=10)$, (3) with $\tilde{C}=1$, the best results obtained when $(C=1, \sigma^2=1)$ and $(C=10, \sigma^2=10)$, (4) with $\tilde{C}=10$, the best results obtained when $(C=10, \sigma^2=1)$ and $(C=10, \sigma^2=100)$. The best result obtained was: accuracy=100\%, sensitivity=100\%, and specificity=100\%. It is worth mentioning that as shown in Fig. \ref{fig:Iris1}(a), linear SVM obtained also competitive results. 

\subsection{Sonar dataset}
Figure \ref{fig:Iris1}(b) shows the obtained results with different values of $C$. The best results obtained when $C\geq 1$ and hence we can set $\tilde{C}$ to different values to find the optimal $(C, \sigma^2)$. After doing the analysis for the dataset we found that the maximum distances between samples of the first and second classes were 3.28 and 3.32; respectively, while the minimum distances were 0.23 and 0.18, respectively. Therefore, the range of $\sigma^2$ is approximately $[0.2^2,3.3^2]\approx [0.04,10.89]$. The grid spacing was 0.01 and the best results obtained with (1) $\tilde{C}=1$ were: accuracy=83.6\%, sensitivity=79.5\%, and specificity=88.1\% when $(C=1, \sigma^2=1)$ and (2) $\tilde{C}=10$ were: accuracy=84.6\%, sensitivity=78.0\%, and specificity=90.3\% when $(C=100, \sigma^2=10)$. The obtained results are much better than the results of linear SVM (see Fig. \ref{fig:Iris1}(b)).

\subsection{Liver dataset}
In the Liver-disorders or simply liver dataset, $\tilde{C}\geq 0.01$. The minimum distance between samples of both classes was zero and the maximum distances between samples of the first (second) class was 208.62 (213.73). Hence, the range of $\sigma^2$ is approximately $[0.001, 100000]$. After conducting different values of $\tilde{C}$, the optimal results (accuracy=73.91\%, sensitivity=63.48\%, and specificity=81.56\%) were obtained with $\tilde{C}=0.1$ and $(C=10000, \sigma^2=100000)$.

\subsection{Pima Indians diabetes}
In the first step in our search algorithm, we found that $\tilde{C}\geq 0.01$. After analyzing the data we found that the minimum distance between samples of the same class was 2.87 and the maximum distance between samples of the same class was 867.5. Hence, the range of $\sigma^2$ is approximately $[10, 2\times10^6]$. By trying different values of $\tilde{C}$, we found that the optimal solution was found when $\tilde{C}=1$, $C=10^6$, and $\sigma^2=10^6$, and the optimal result was: accuracy 77.6\%, sensitivity 57.0\%, and specificity 77.6\%.

\subsection{Breast cancer}
The best results using linear SVM were obtained when $\tilde{C}\geq 0.001$. The minimum distance between samples of the same class was zero and the maximum distance was 23.3. Hence, the range of $\sigma^2$ is $[10^{-3}, 1000]$. The best results in the second step of our search strategy were accuracy=96.93\%, sensitivity=96.96\%, and specificity=97.13\% and these results were obtained when $\tilde{C}=0.001$, $C=0.1$, and $\sigma^2=100$. 

\subsection{Iono dataset}
In the first step of our search algorithm (i.e. linear SVM), $\tilde{C}$ was $\geq 0.01$. The maximum distance between the samples of the same class was 9.75 and the minimum distance was zero. Hence, the range of the search space of $\sigma^2$ was $[0.001,100]$. The optimal solution was: accuracy=94.86\%, sensitivity=87.2\%, and specificity=98.7\% and this optimal solution was found when $\tilde{C}=1$ and $(C=10, \sigma^2=1)$. 

\subsection{Imbalanced datasets}
In all the previous experiments, the datasets have a small IR. Practically, there are many problems suffer from the imbalanced data problem with high IR. The goal of our experiments in this section is to test our search strategy with imbalanced data with high IR. These datasets are divided into two divisions: (1) datasets with IR lower than nine, and (2) datasets with IR higher than nine.

\subsubsection{Imbalanced datasets with IR lower than nine}
In this section, we used only two imbalanced datasets with IR lower than nine. In both datasets, there are two classes, and each sample is represented by seven features. The details of each experiment are as follows:

\begin{itemize}
\item \textbf{Ecoli1 dataset}: With this dataset, in the first step, the best results obtained when $\tilde{C}\geq 1$ and the maximum distance between samples from the same class was 1.28 and the minimum distance between samples from the same class was 0.04. Therefore, the range of $\sigma^2$ is $[0.01, 10]$. Searching along the line which is defined in the second step of our searching algorithm obtains competitive results. The best accuracy was 90.72\% with sensitivity=79\% and specificity=94.3\% and this solution obtained when $\tilde{C}=100$, $C=10$, and $\sigma^2=0.1$.
\item \textbf{Ecoli2 dataset}: With linear SVM, the best results achieved with $\tilde{C}\geq 1$. We used the same range of $\sigma^2$ that we used in the Ecoli1 dataset, and we found that the best accuracy was 95.9\% with sensitivity 85.7\% and specificity 97.9\% and this solution was found when $\tilde{C}=10$, $C=1$, and $\sigma^2=0.1$. However, the best sensitivity obtained when $\tilde{C}=1000$ and $(C=10000, \sigma^2=10)$, and the sensitivity was 91\% (accuracy=95.6\% and specificity=96.5\%).
\end{itemize}

To conclude, these findings indicate that our searching strategy obtains competitive results even with the imbalanced data with IR$\leq 9$. 

\subsubsection{Imbalanced datasets with IR higher than nine}
In this section, we used only two imbalanced datasets with IR higher than nine. In both datasets, there are two classes, and each sample is represented by nine features. The details of each experiment are as follows:

\begin{itemize}

\item \textbf{Glass2 dataset}: The value of $\tilde{C}$ was more than $10^6$ and the maximum distance between samples from the same class was 12.04 and the minimum distance was zero. Therefore, the range of $\sigma^2$ is $[0.001, 1000]$. Searching along the line which is defined in the second step of our searching algorithm obtained competitive results. The best accuracy was 92.06\% with sensitivity=0\% and specificity=100\% and this solution obtained when $\tilde{C}=10^6$, $C=1000$, and $\sigma^2=10^{-3}$.
\item \textbf{Glass4 dataset}: With linear SVM, the best $\tilde{C}$ was $\geq 10$. We used the same range of $\sigma^2$ that we used in the Glass2 dataset, and we found that the best accuracy was 97.2\% with sensitivity 53.3\% and specificity 97.2\% and this solution was found when $\tilde{C}=10$, $C=10$, and $\sigma^2=1$.  
\end{itemize}

From these findings, we can conclude that:
\begin{itemize}
\item Despite that the grid search algorithm scanned the whole parameter space, the proposed search algorithm obtained results in most cases better or at least equal to it (the grid search algorithm). This is because the proposed algorithm searches only in two one-dimensional spaces and this is much faster than searching within a two-dimensional space. As a consequence, we can use smaller grid spacing and this gives the chance for the proposed algorithm to find the optimal solution than the grid search.
\item Practically, linear SVM achieved reasonable performance with some data while adding some nonlinearities to the SVM model helps to improve the classification performance.
\item Analyzing the data by exploring the feature ranges of the training data gives the chance for the proposed algorithm to focus on a small set of values instead of searching in the whole space. For example, in \cite{lin2008particle,tharwat2017ba}, used the same range of features with all datasets which have different ranges of features, where $0.01\leq C\leq 3000$ and $0.01\leq \sigma \leq 100$. In \cite{tharwat2018chaotic}, the search space of $C$ increased to be $0.01\leq C\leq 35000$. By contrast, in our analysis, for example, with some datasets as the Sonar dataset, the range of $\sigma^2$ was $[0.04,10.89]$ and with the iris dataset the range of $\sigma^2$ is $[0.01,7]$. With this small search space, we can reduce the grid spacing. This increases the chance of finding the optimal solution or at least the proposed algorithm will be closer to the optimal solution than the grid search algorithm. Moreover, with a small search space, instead of searching along one line as reported in \cite{keerthi2003asymptotic}, we can try different values of $\tilde{C}$ and hence searching along many lines. Practically, for example, with $\tilde{C}\geq 0.01$, this means that $\tilde{C}$ have different values. However, after a certain limit, the obtained results decrease by increasing $\tilde{C}$ more than this limit and the results get worse.
\item The imbalanced data with high IR and also with a small number of minority samples obtain low sensitivity rate. This is because, in all our experiments, we used five-fold cross-validation and hence the training data have not enough samples from the minority class to train the model. For example, with the Glass4 dataset, there are 13 minority samples and if we divided the data into five folds, the best case that each fold has two or three minority samples while each fold has approximately $\frac{201}{5}\approx 40$ samples from the majority class. Hence, there is no enough minority data for training the SVM model and this will reduce the sensitivity results as mentioned in our results. Therefore, one of the algorithms could be used for generating new minority samples to give a better chance for the SVM model to explore the minority class.

\end{itemize}

\section{Conclusions}
\label{Sec:Conc}
Support vector machine (SVM) is one of the well-known learning algorithms and it has been used in many applications. However, the classification performance of SVM is highly affected by its parameter's values and this may lead the SVM model to severe underfitting or overfitting. In many studies, SVM parameters are used as a black box, without understanding the internal details. Moreover, many studies optimized SVM parameters without taking into account if the data is balanced or not. However, the problem of imbalanced data is a challenging problem for building robust classification models. In this paper, the behavior of SVM with Gaussian and linear kernel functions is analyzed with balanced and imbalanced data. These analyses including some numerical examples, visualization, and mathematical explanations to show (1) the training and testing error rates, (2) decision boundaries, (3), SVM margin, (4) complexity of the classification model, and (5) number of support vectors with different parameters' values. From this analysis, we proposed a new algorithm for optimizing SVM parameters. This algorithm has three main steps: (1) using linear SVM to search for the optimal $C$ and call it $\tilde{C}$, (2) analyze the data to find the range of the search space of $\sigma^2$, and (3) search for the optimal $C, \sigma^2$ along the line $\text{log} \sigma^2=\text{log} C-\text{log} \tilde{C}$.

This paper followed the approach of not only explaining the results of some experiments; but, also visualizing these results with figures to make it easy to understand. Additionally, some numerical examples are presented and graphically illustrated for explaining geometrically and mathematically the impact of the Gaussian and linear kernel functions on the performance of the SVM model.

Several directions for future studies are suggested. Firstly, datasets with a severe imbalance ratio should be used for evaluating the proposed algorithm. Secondly, in our experiments, all datasets have only two classes, but other datasets with multi-classes and high dimensions should be tested.

\section{Appendix}
\subsection{Quadratic Programming}
\label{Subsec:Quad}
Linearly constrained optimization problems with a quadratic objective function are called \emph{Quadratic program} or \emph{Quadratic Programming} (QP) optimization problems. In these problems, the objective function will be quadratic as follows:
\begin{align}
&\text{min } f(\textbf{x})= \frac{1}{2}\textbf{x}^T\textbf{Q}\textbf{x}+\textbf{cx} \nonumber \\
&\text{ s.t. } g(\textbf{x})= \textbf{Ax}\leq \textbf{b}
\label{EQN:StandardQuafraticfn}
\end{align}
where $\textbf{c}\in R^n$ is the coefficient of the linear terms in the objective function ($f$), $\textbf{Q}\in R^{n\times n}$ is symmetric matrix and it represents the coefficients of the quadratic term, $\textbf{Ax}\leq \textbf{b}$ represents the linear constraints and these constraints represent the boundary of the feasible region of the optimization problem, $\textbf{A}\in R^{m\times n}$ is the coefficients of the constraints, the vector $\textbf{x}$ has $n$ dimensions and it represents the decision variables, $m$ is the number of constraints, and $\textbf{b}$ is the right-hand-side coefficients. 

If the objective function is convex\footnote{The function $f({x})$ is convex if a line drawn between any two points on the function remains on or above the function in the interval between the two points \cite{kecman2001learning}.} for all feasible points or simply on the feasible region; hence, there is one local minimum which is the global minimum also. The objective function is convex if and only if the $\textbf{Q}$ matrix\footnote{If the function is twice differentiable, the Hessian matrix is the matrix of second order derivatives as follows, $[\textbf{H}(x)]_{ij}=\frac{\partial^2 f(x)}{\partial x_i \partial x_j}$. $\textbf{H}$ is symmetric (i.e. $\frac{\partial^2 f(x)}{\partial x_i \partial x_j}=\frac{\partial^2 f(x)}{\partial x_j \partial x_i}$). In quadratic programming optimization problems, $\textbf{H}(x)$ is the quadratic term in Equation (\ref{EQN:StandardQuafraticfn}). Thus, the Hessian matrix is used in some references to test the convexity of a function, and $\textbf{Q}$ and $\textbf{H}$ matrices are the same.} is positive semidefinite\footnote{$\textbf{X}$ is positive semi-definite if $v^T\textbf{X}v \geq 0$ for all $v\neq 0$. In other words, all eigenvalues of $\textbf{X}$ are $\geq0$.}. If $\textbf{Q}$ is positive definite and minimum or negative definite and maximum; so, the objective function is strictly convex\footnote{Let $X$ be a convex set. The function $f$ is strictly convex if: $\forall x_1\neq x_2 \in X, \forall t\in (0,1): f(tx_1+(1-t)x_2)<tf(x_1)+(t-1)f(x_2)$.} and the function will be in a round bowl shape; hence, it has only one global optimal solution. While the objective function will be convex but not strictly when $\textbf{Q}$ is positive semidefinite and minimum; hence, many points have the same objective value (i.e. we can find a flat region in the objective function). The objective function will have many local optimal solutions when $\textbf{Q}$ is indefinite.

\begin{figure*}[!t]
\centering
\subfloat[\label{fig:0}]{\includegraphics[width=0.33\textwidth]{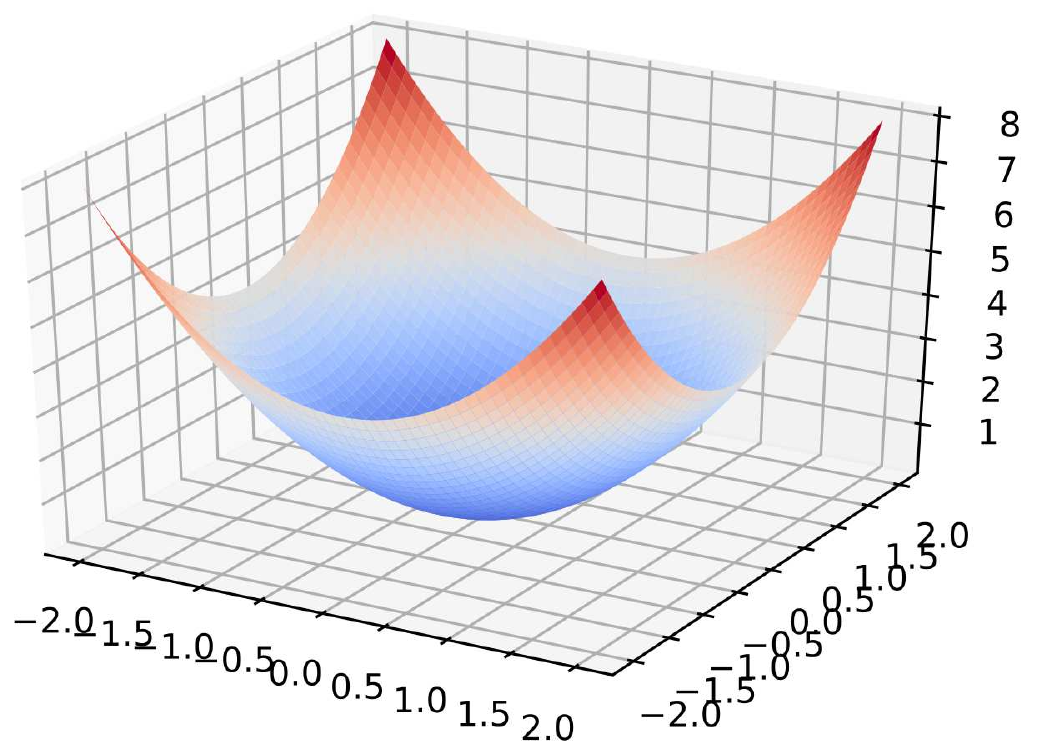}}\hfill
\subfloat[\label{fig:1}]{\includegraphics[width=0.33\textwidth]{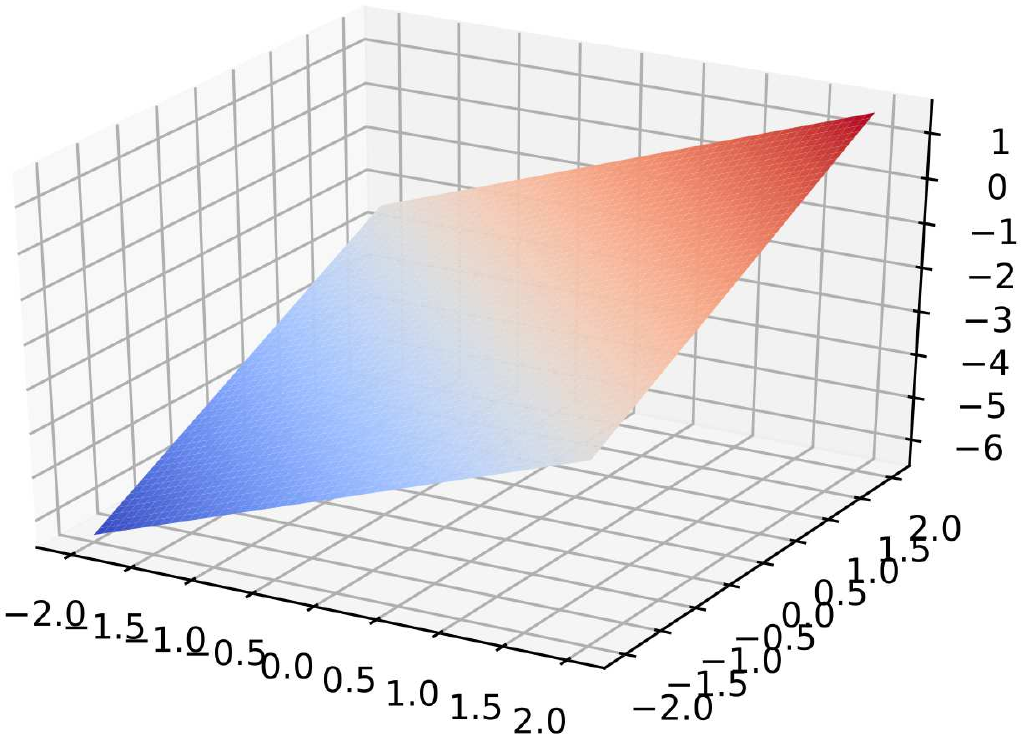}}\hfill
\subfloat[\label{fig:1}]{\includegraphics[width=0.33\textwidth]{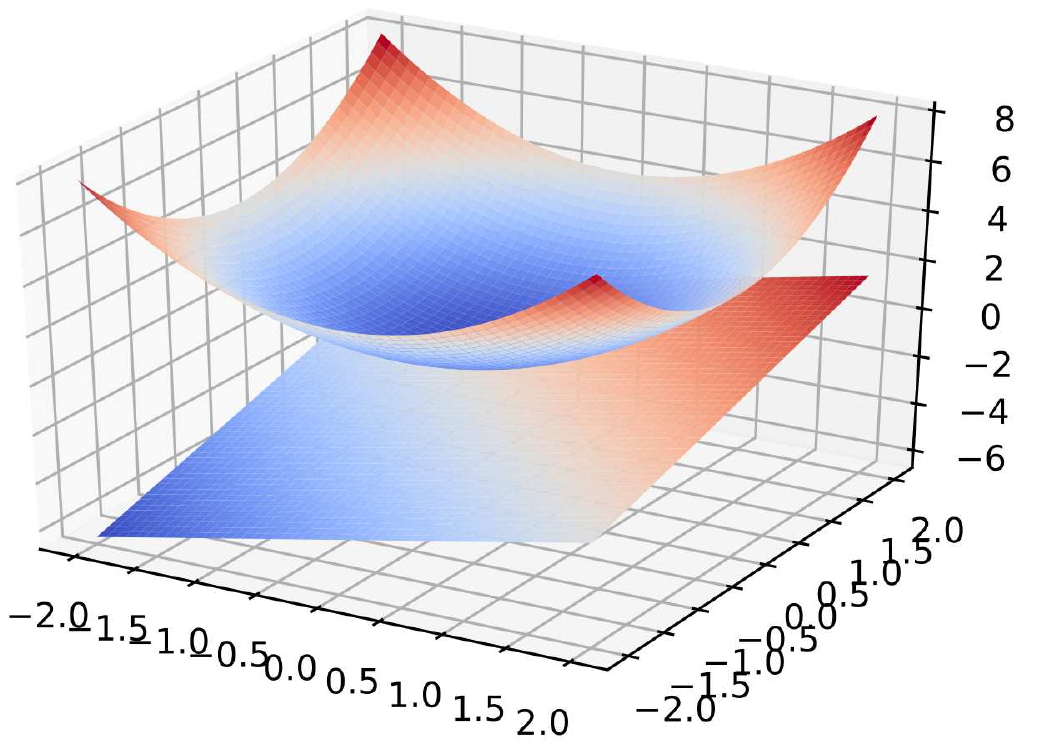}}\hfill
\subfloat[\label{fig:1}]{\includegraphics[width=0.33\textwidth]{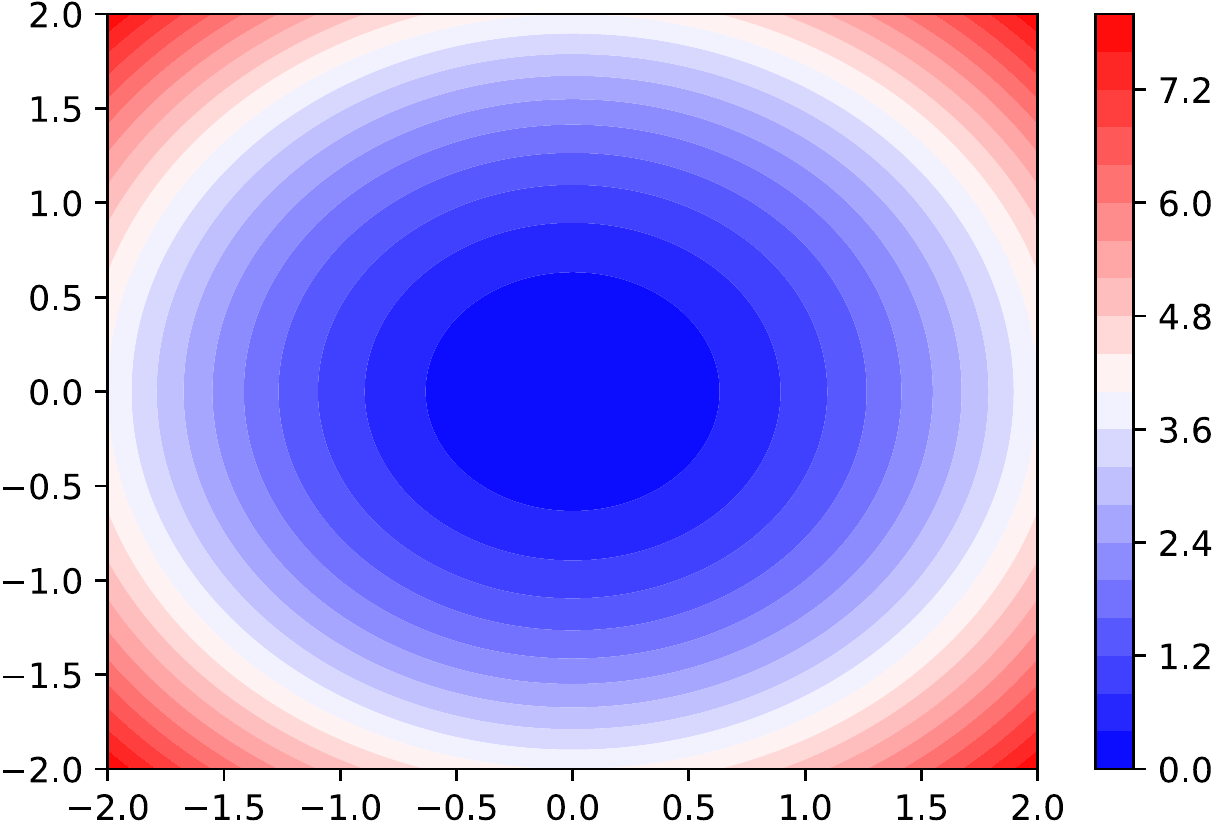}}\hfill
\subfloat[\label{fig:1}]{\includegraphics[width=0.33\textwidth]{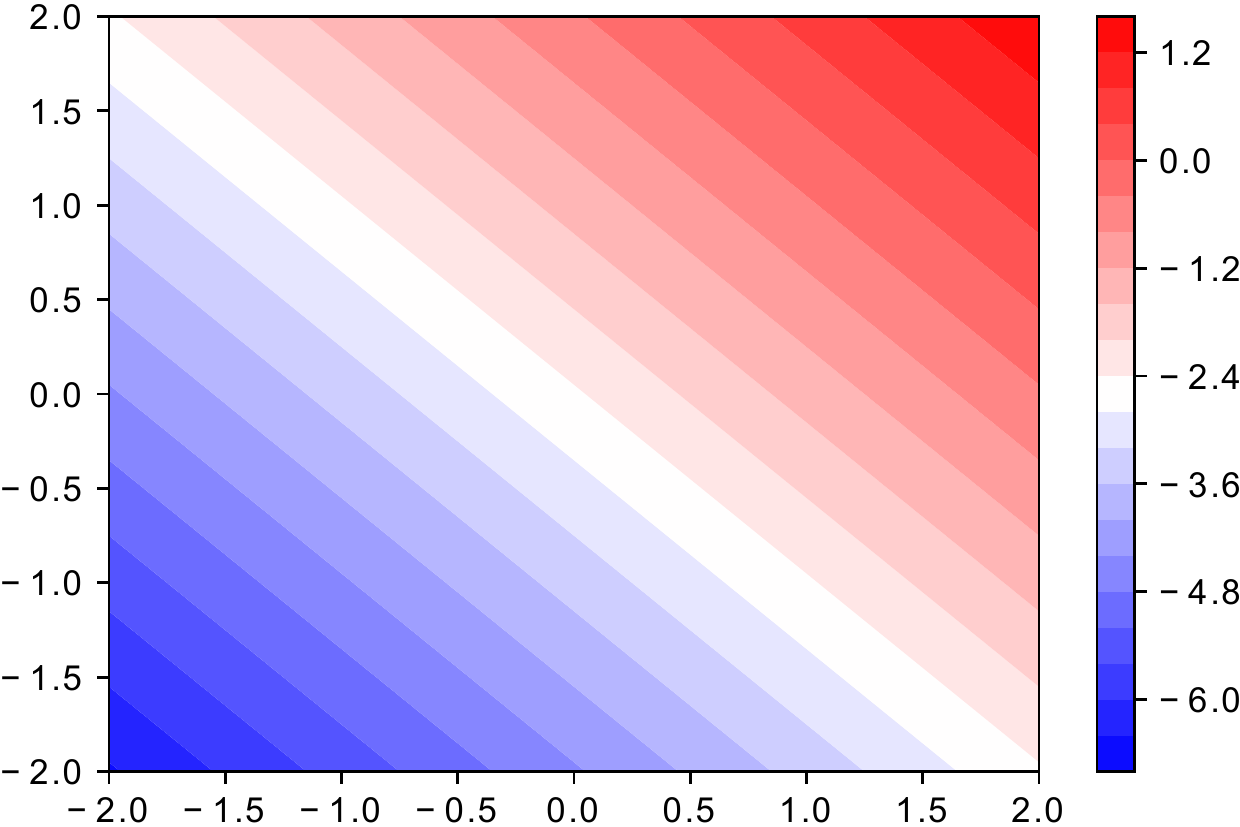}}\hfill
\subfloat[\label{fig:1}]{\includegraphics[width=0.33\textwidth]{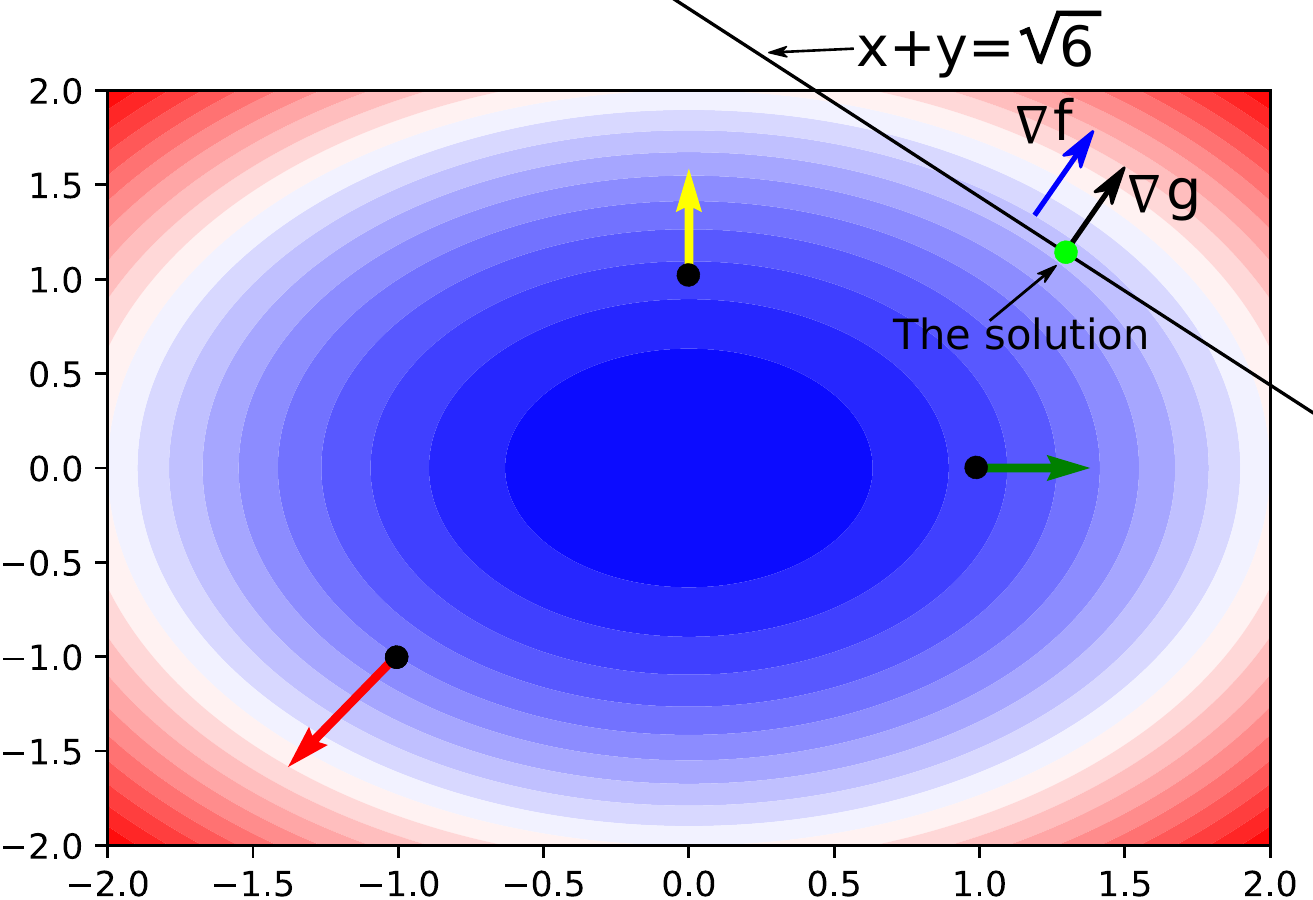}}\hfill
\caption{Visualization of the objective function and the constraint of our example in Equation (\ref{EQN:Quad:Example1}). (a and d) The 3D surface and the contour plot of the objective function. The objective function is a strictly convex optimization function and it has only one optimal solution at the origin. (b and e) The 3D surface and the contour plot of the linear constraint in our example in Equation (\ref{EQN:Quad:Example1}). (c and f) The 3D surface and the contour plot of the intersection between the objective function and the constraint of our example in Equation (\ref{EQN:Quad:Example1}). This intersection represents the solution for the optimization problem. In (f), the red, yellow, and green arrows represent different gradient vectors at three different points. The blue and black arrows are parallel (i.e. $\bigtriangledown f \parallel \bigtriangledown g$).}
\label{fig:ExampleObjectivefn}
\end{figure*}

Given an optimization problem as follows:
\begin{align}
&\text{min }  f(x,y) = x^2+y^2 \nonumber \\
&\text{ s.t. } g(x,y)= x+y=\sqrt{6}
\label{EQN:Quad:Example1}
\end{align}

From Equation (\ref{EQN:Quad:Example1}), the objective function ($f$) is quadratic and the constraint is linear. Figure \ref{fig:ExampleObjectivefn}(a and d) shows the surface and contour plots of the objective function ($f(x,y)$) and Fig. \ref{fig:ExampleObjectivefn}(b and e) displays the surface and contour plots of the constraint ($g(x,y)$). As shown, the objective function is strictly convex and the optimal solution is located at the origin. 

The gradient vector of a function $f(x,y)$ is denoted by $\bigtriangledown f$ and it is calculated as follows:
\begin{align*}
\bigtriangledown f& = \text{grad} f =\left \langle \frac{\partial f}{\partial x}, \frac{\partial f}{\partial y} \right \rangle
\end{align*}

The gradient of our example in Equation (\ref{EQN:Quad:Example1}) is as follows:
\begin{align}
\bigtriangledown f=\left \langle 2x, 2y \right \rangle
\label{EQN:GradientExample}
\end{align}

\begin{figure}[!ht]%2
\centering
\includegraphics[width=0.5\textwidth]{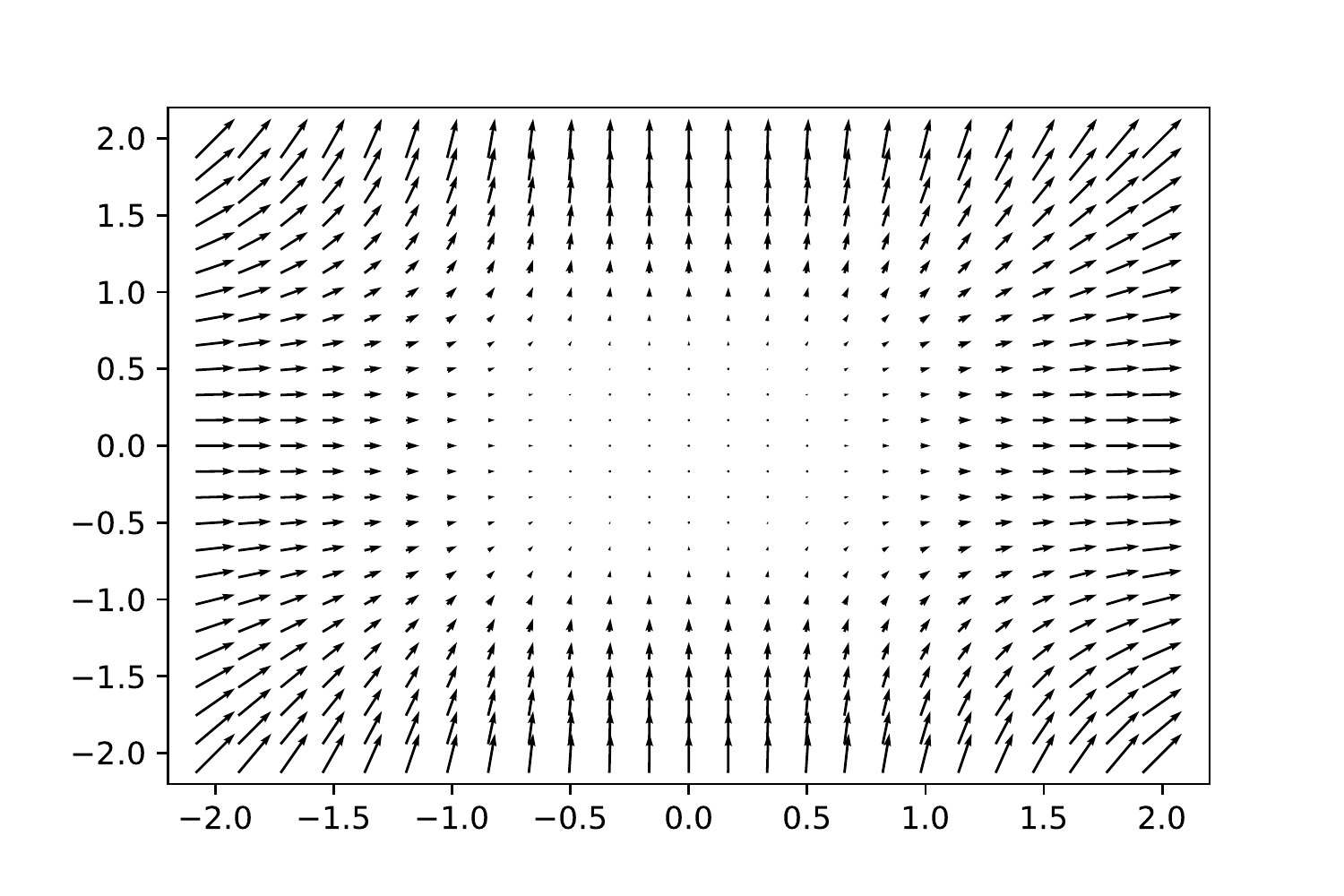} 
\caption{Quiver, vector field, or velocity vectors plot of the objective function in Equation (\ref{EQN:Quad:Example1}). The direction of arrows points to the direction of the greatest rate of change of the function. The length of each arrow represents the rate of change and it is clear that near the origin which is the optimal solution the arrows are short which reflects the small rate of change.}
\label{fig:Quiver}
\end{figure}

Figure \ref{fig:ExampleObjectivefn}(c and f) shows the gradient vectors of the objective function ($\bigtriangledown f$) and the gradient of the constraint ($\bigtriangledown g$). Figure \ref{fig:ExampleObjectivefn}(f) shows the gradient vectors at three different points $(-1,-1)$, $(1,0)$, and $(0,1)$. For example, with the first point ($-1,-1$), the gradient vector at this point can be calculated by substituting it in Equation (\ref{EQN:GradientExample}), and the gradient vector will be $\left \langle 2x, 2y \right \rangle=\left \langle -2, -2 \right \rangle$, and it is represented in the figure by a red arrow. Similarly, the other two gradient vectors at $(0,1)$ and $(1,0)$  are represented by yellow and green arrows, respectively. The gradient vectors at any point can be calculated and Fig. \ref{fig:Quiver} shows the gradient vectors of the objective function in Equation (\ref{EQN:Quad:Example1}) at many points. As shown, the gradient vector at any point is normal to the level curve, and it (the gradient vector) points in the direction of greatest rate of change of $f(x,y)$. 

From Fig. \ref{fig:ExampleObjectivefn}(f) and \ref{fig:Quiver}, it is clear also that $\bigtriangledown f(x,y) \perp f$ at any point and similarly also for the constraint function $\bigtriangledown g \perp g(x,y)$. In our example, the solution is a point(s) that satisfy the constraint and also minimizes the objective function. At this point, the two gradient vectors are parallel (i.e. $\bigtriangledown f \text{ and } \bigtriangledown g$ are in the same or opposite direction, so $\bigtriangledown f=\pm \alpha \bigtriangledown g$), and the constraint line ($g$) is tangent to the inner contour line of $f$, i.e. inner ellipse contour line of $f$ (see Fig. \ref{fig:ExampleObjectivefn}(c and f)). 

The optimization problem in Equations (\ref{EQN:StandardQuafraticfn}) can be solved by combining the objective function ($f$) and the constraint ($g$) into one Lagrangian function by introducing new slack variables $\alpha$ as follows:
\begin{align*}
L(\textbf{x},\alpha)=f(\textbf{x})-\alpha g(\textbf{x})
%\label{EQN:PrimalToDual1}
\end{align*}
where $L$ is called \emph{Lagrangian} and it can be solved as follows, $\bigtriangledown L(\textbf{x},\alpha)=0$ and this finds the points where the gradient of both $f$ and $g$ are parallel, and $\alpha$ is called \emph{Lagrange multiplier}. Hence, $L(\textbf{x},\alpha)$ will be calculated as follows:
\begin{align}
L(\textbf{x},\alpha)&=f(\textbf{x})- \sum_{i} \alpha_i g_i(\textbf{x}) \nonumber \\ 
&=\frac{1}{2}\textbf{x}^T\textbf{Qx}+ \textbf{cx} - \alpha (\textbf{Ax}-\textbf{b}) 
\label{EQN:Lagrangian}
\end{align}
and hence $L(x,y,\alpha)$ of our optimization problem in Equation (\ref{EQN:Quad:Example1}) will be
\begin{align}
L(x,y,\alpha)&=f(x,y) - \alpha g(x,y) \nonumber \\
&=x^2+y^2 - \alpha (x+y-\sqrt{6})
\label{EQN:QuadExample2}
\end{align}

Equation (\ref{EQN:Lagrangian}) is solved by differentiating it with respect to $x_i$ and $\alpha_i$ as follows\footnote{$x_i, i=1,2,\dots, n$ represent the decision variables and in our example in Equation (\ref{EQN:Quad:Example1}) we have two decision variables: $x$ and $y$.} 
\begin{align*}
&\frac{\partial L(\textbf{x},{\alpha})}{\partial x_i}=0 \Rightarrow \textbf{c}+\textbf{xQ}-\textbf{A}\alpha = 0\;,\; i=1,2,\dots,n \nonumber \\
&\frac{\partial L(\textbf{x},{\alpha})}{\partial \alpha_i}=0 \Rightarrow \textbf{Ax}-\textbf{b}= 0\;,\; i=1,2,\dots,m
\end{align*}

Similarly, in our example, we can find the solutions by differentiating Equation (\ref{EQN:QuadExample2}) with respect to $x,\;y$, and $\alpha$ as follows:
\begin{align}
\frac{\partial L(x,y,\alpha)}{\partial x} &=0 \Rightarrow 2x-\alpha=0 \nonumber \\
\frac{\partial L(x,y,\alpha)}{\partial y} &=0 \Rightarrow 2y-\alpha=0 \nonumber \\
\frac{\partial L(x,y,\alpha)}{\partial \alpha} &=0 \Rightarrow x+y-\sqrt{6}=0 
\label{EQN:Quad:Example3}
\end{align}

By solving Equation (\ref{EQN:Quad:Example3}), we found that $\alpha=2x=\sqrt{6}$ and $x=y=\frac{\sqrt{6}}{2}\approx 1.225$ (see Fig. \ref{fig:ExampleObjectivefn}(f)). 

\subsection{Dual vs. primal problems}
\label{Subsec:Dual}
The solutions of dual problems represent the bounds of the solutions of the primal problems. For example, given a minimization problem, we can formulate it as a maximization problem. Hence, we will find the maximum for the problem, and this maximum is the lower boundary to the solutions of the minimization problem. The dual solution ($D$) represents the lower bound of the solutions of the primal problem ($P$). The difference between $P$ and $D$ is called the \emph{duality gap}. If $P-D=0$; thus, there is no duality gap, and this is known as a \emph{strong duality}, this case achieves in case of convex optimization problems and satisfies constraints. If the gap is strictly positive, this is called the \emph{weak duality} (i.e. the optimal value of the primal minimization problem is greater than the dual problem) \cite{kojima1989primal}.  

However, sometimes solving dual problems is simpler than solving the primal problems, particularly when the number of decision variables is considerably less than the number of slack/surplus variables.

Given a primal problem as follows:
\begin{align*}
\text{ min } Z_P&=\textbf{CX } \nonumber \\
\text{s.t. } & \textbf{AX}\leq \textbf{B}, \;\; \textbf{X}\geq 0 
%\label{EQN:PrimaltoDual}
\end{align*}
and the corresponding dual problem of $Z_P$ in is
\begin{align*}
\text{ max } Z_D&=\textbf{B}^T\textbf{W}  \nonumber \\
\text{s.t. } & \textbf{A}^T \textbf{W} \geq  \textbf{C}^T,\;\;  \textbf{W}\geq 0 
%\label{EQN:PrimaltoDual2}
\end{align*}

The dual of dual problem is the primal problem, and if the primal problem has a solution then the dual problem also has a solution and vice versa.

The Lagrangian of the dual function of our example in Equation (\ref{EQN:Quad:Example1}) is as follows:
\begin{align*}
\underset{\alpha\geq 0}{\text{max }} L_D(\alpha)&=(\frac{\alpha}{2})^2+(\frac{\alpha}{2})^2-\alpha (\frac{\alpha}{2}+\frac{\alpha}{2}-\sqrt{6})\nonumber \\
&=\frac{\alpha^2}{2}-\alpha (\alpha-\sqrt{6})=-\frac{\alpha^2}{2}+ \sqrt{6}\alpha
\end{align*}
where $x=y=\frac{\alpha}{2}$ (as in Equation (\ref{EQN:Quad:Example3})). Now, $L_D$ has only one parameter and it can be solved easily as follows:
\begin{align*}
\frac{\partial L_D}{\partial \alpha} &=0 \Rightarrow -\alpha+\sqrt{6}=0
\end{align*}
as a result, $\alpha=\sqrt{6}$ and $x=y=\frac{\alpha}{2}=\frac{\sqrt{6}}{2}$ and these results agree with the results of the primal problem. In other words, the optimal solutions of both primal and dual form of the same problem are the same and there is no duality gap.

\bibliographystyle{splncs}
\bibliography{bibliography}
\end{document}